\documentclass{article}

\usepackage[preprint]{neurips_2024}
\usepackage{tcolorbox}
\usepackage{tocbibind} % This package automatically includes the ToC in the table of contents
\usepackage[toc,page,header]{appendix}
\usepackage{minitoc}

\usepackage{graphicx} 
\usepackage{amsmath}
\usepackage{amsfonts}
\usepackage{adjustbox}
\usepackage{xkeyval}      
\usepackage[utf8]{inputenc} % allow utf-8 input
\usepackage[T1]{fontenc}    % use 8-bit T1 fonts
\usepackage{hyperref}       % hyperlinks
\usepackage{url}            % simple URL typesetting
\usepackage{booktabs}       % professional-quality tables
\usepackage{amsfonts}       % blackboard math symbols
\usepackage{nicefrac}       % compact symbols for 1/2, etc.
\usepackage{microtype}      % microtypography
\usepackage{xcolor}         % colors
\usepackage{changepage}

\usepackage{graphicx}
\usepackage{subcaption}
\usepackage{wrapfig}
\tcbuselibrary{breakable}

\title{ \raisebox{-0.35\height}{\includegraphics[width=1.33 cm]{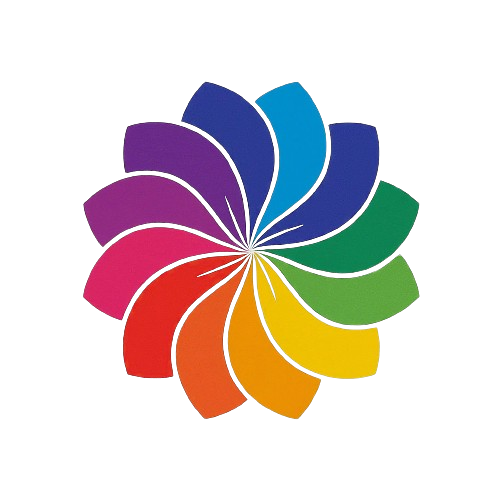}} Spectra: Surprising Effectiveness of Pretraining Ternary Language Models at Scale}

\author{
    Ayush Kaushal\textsuperscript{1,2}\thanks{Equal contribution, listed in alphabetical order.} \\
    \texttt{ayush@nolano.ai} \\
    \And
    Tejas Vaidhya\textsuperscript{1,2,4}\footnotemark[1]  \\
    \texttt{tejas@nolano.ai} \\
    \And
    Arnab Kumar Mondal\textsuperscript{4}\\
    \texttt{arnab.mondal@mila.quebec} \\
    \And
    Tejas Pandey\textsuperscript{1,3}  \\
    \texttt{tejaspandey2003@kgpian.iitkgp.ac.in} \\
    \And
    Aaryan Bhagat\textsuperscript{5} \\
    \texttt{abhag017@ucr.edu} \\
    \And
    Irina Rish\textsuperscript{1, 2, 4} \\
    \texttt{irina.rish@gmail.com} \\
    \AND
    \textsuperscript{1}\textnormal{Nolano AI}, \textsuperscript{2}\textnormal{University of Montreal}, \textsuperscript{3}\textnormal{IIT Kharagpur}, 
    \\
    \textsuperscript{4}\textnormal{Mila - Quebec AI Institute}, \textsuperscript{5}\textnormal{UC Riverside}
}

\newcommand{\commentall}[1]{}

\begin{document}
\maketitle
\vspace{-0.6 cm}
\begin{abstract}

Rapid advancements in GPU computational power has outpaced memory capacity and bandwidth growth, creating bottlenecks in Large Language Model (LLM) inference. Post-training quantization is the leading method for addressing memory-related bottlenecks in LLM inference, but it suffers from significant performance degradation below 4-bit precision. This paper addresses these challenges by investigating the pretraining of low-bitwidth models specifically {\em Ternary Language Models (TriLMs)} as an alternative to traditional floating-point models (FloatLMs) and their post-training quantized versions (QuantLMs). We present {\em Spectra LLM suite}, the first open suite of LLMs spanning multiple bit-widths, including FloatLMs, QuantLMs, and TriLMs, ranging from 99M to 3.9B parameters trained on 300B tokens. Our comprehensive evaluation demonstrates that TriLMs offer superior scaling behavior in terms of model size (in bits). Surprisingly, at scales exceeding one billion parameters, TriLMs consistently outperform their QuantLM and FloatLM counterparts for a given bit size across various benchmarks. Notably, the 3.9B parameter TriLM matches the performance of the FloatLM 3.9B across all benchmarks, despite having fewer bits than FloatLM 830M. Overall, this research provides valuable insights into the feasibility and scalability of low-bitwidth language models, paving the way for the development of more efficient LLMs.
    
To enhance understanding of low-bitwidth models, we are releasing 500+ intermediate checkpoints of the Spectra suite at
\href{https://github.com/NolanoOrg/SpectraSuite}{https://github.com/NolanoOrg/SpectraSuite}.

\end{abstract}
\vspace{-0.5 cm}

\section{Introduction}
   \begin{figure}[h]
        \centering
        \begin{adjustbox}{width=0.8\textwidth}
        \begin{subfigure}{0.45\textwidth}
            \centering            \includegraphics[width=0.91\linewidth, height=0.21\textheight ]{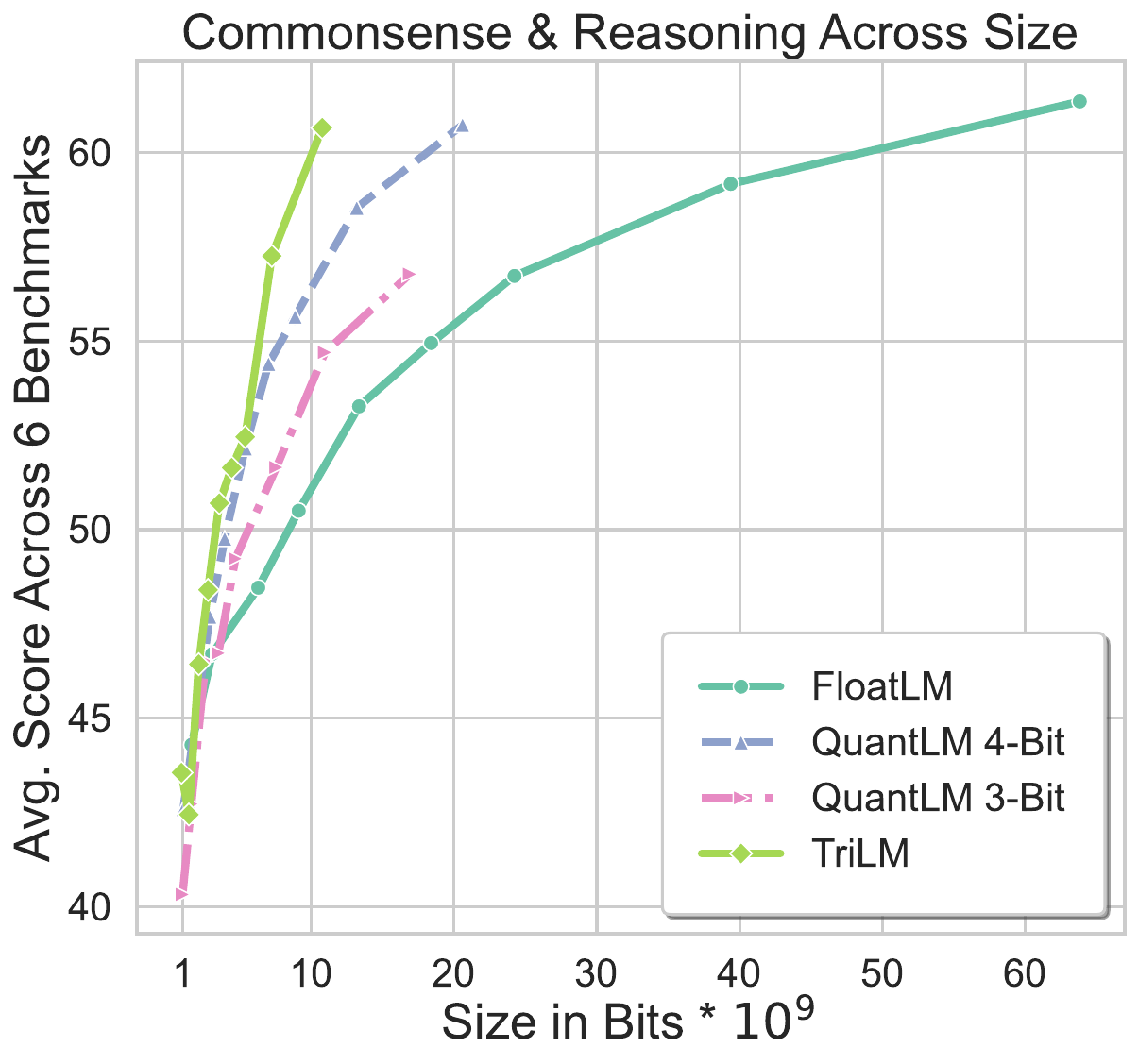}
            \caption{C\&R: acc vs size}
            \label{fig:commonsense_and_reasoning_across_size}
        \end{subfigure}
        \begin{subfigure}{0.45\textwidth}
            \centering
            \includegraphics[width=0.91\linewidth,  height=0.21\textheight ]{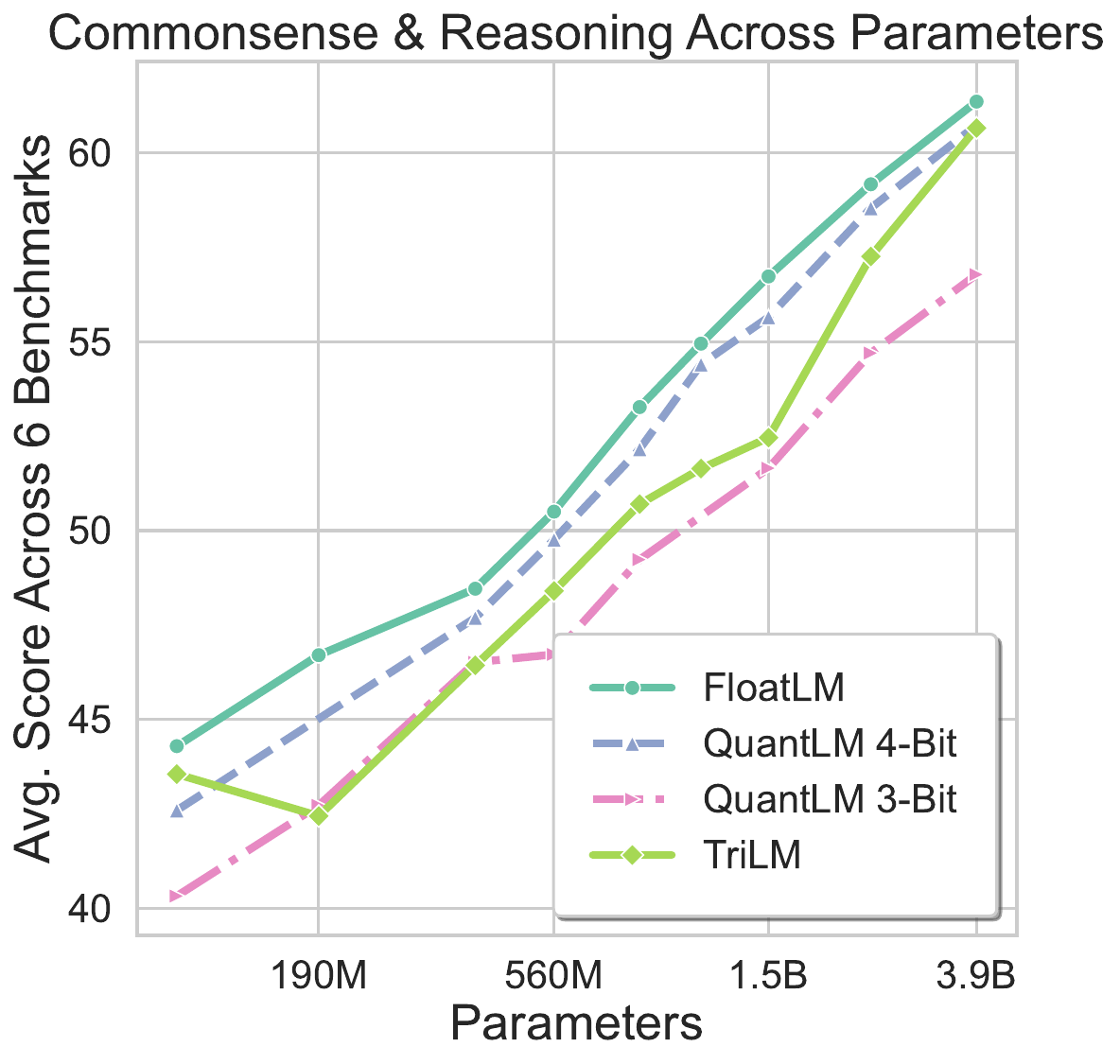}
            \caption{C\&R: acc vs parameters}
            \label{fig:commonsense_and_reasoning_across_params}
        \end{subfigure}
        \end{adjustbox}
        \vskip\baselineskip
        \begin{adjustbox}{width=0.8\textwidth}
        \begin{subfigure}{0.45\textwidth}
            \centering
            \includegraphics[width=0.91\linewidth, height=0.21\textheight ]{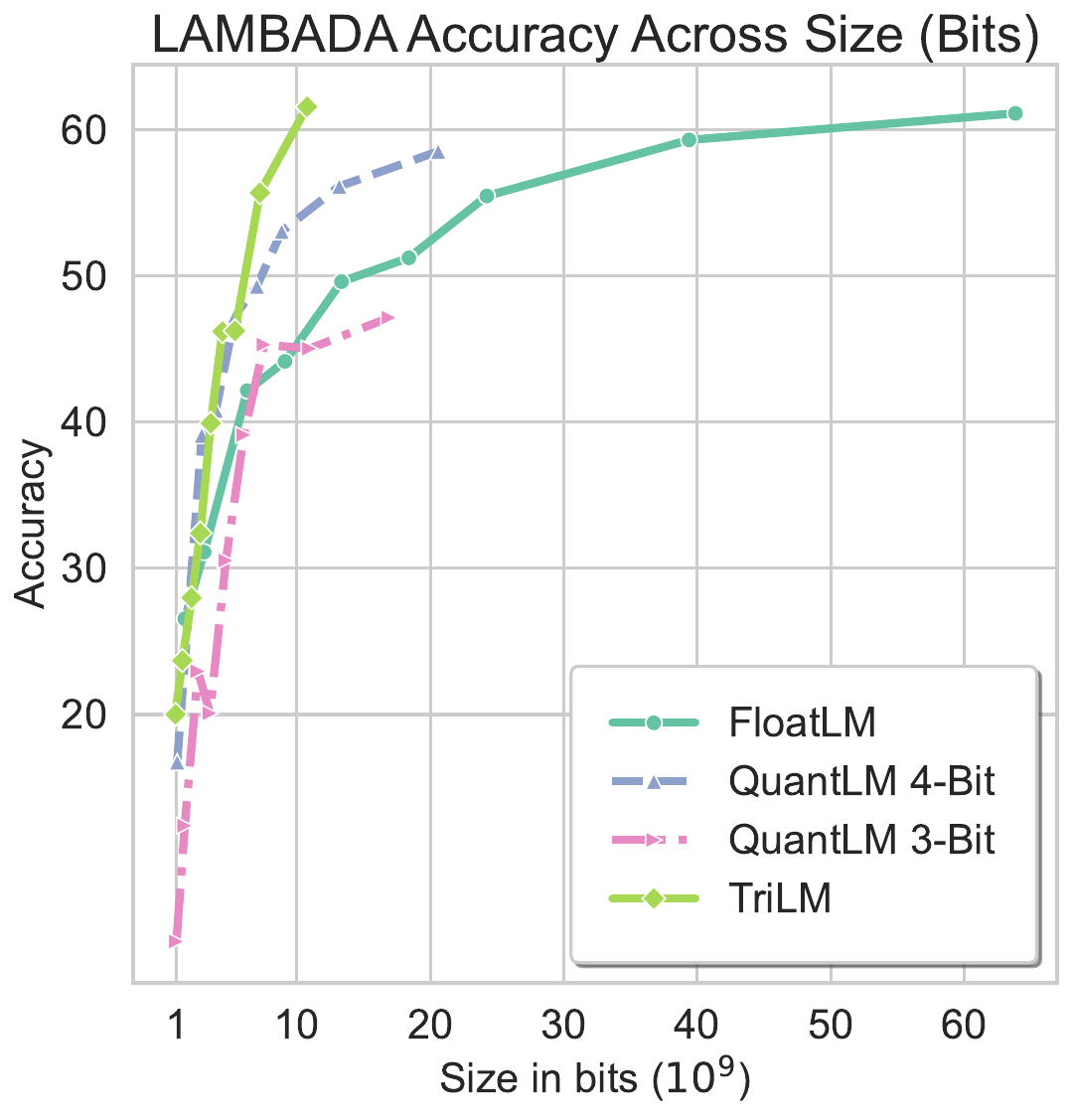}
            \caption{LAMBADA: acc vs size}
            \label{fig:LAMBADA_acc_across_size}
        \end{subfigure}
        \begin{subfigure}{0.45\textwidth}
            \centering
            \includegraphics[width=0.91\linewidth, height=0.21\textheight]{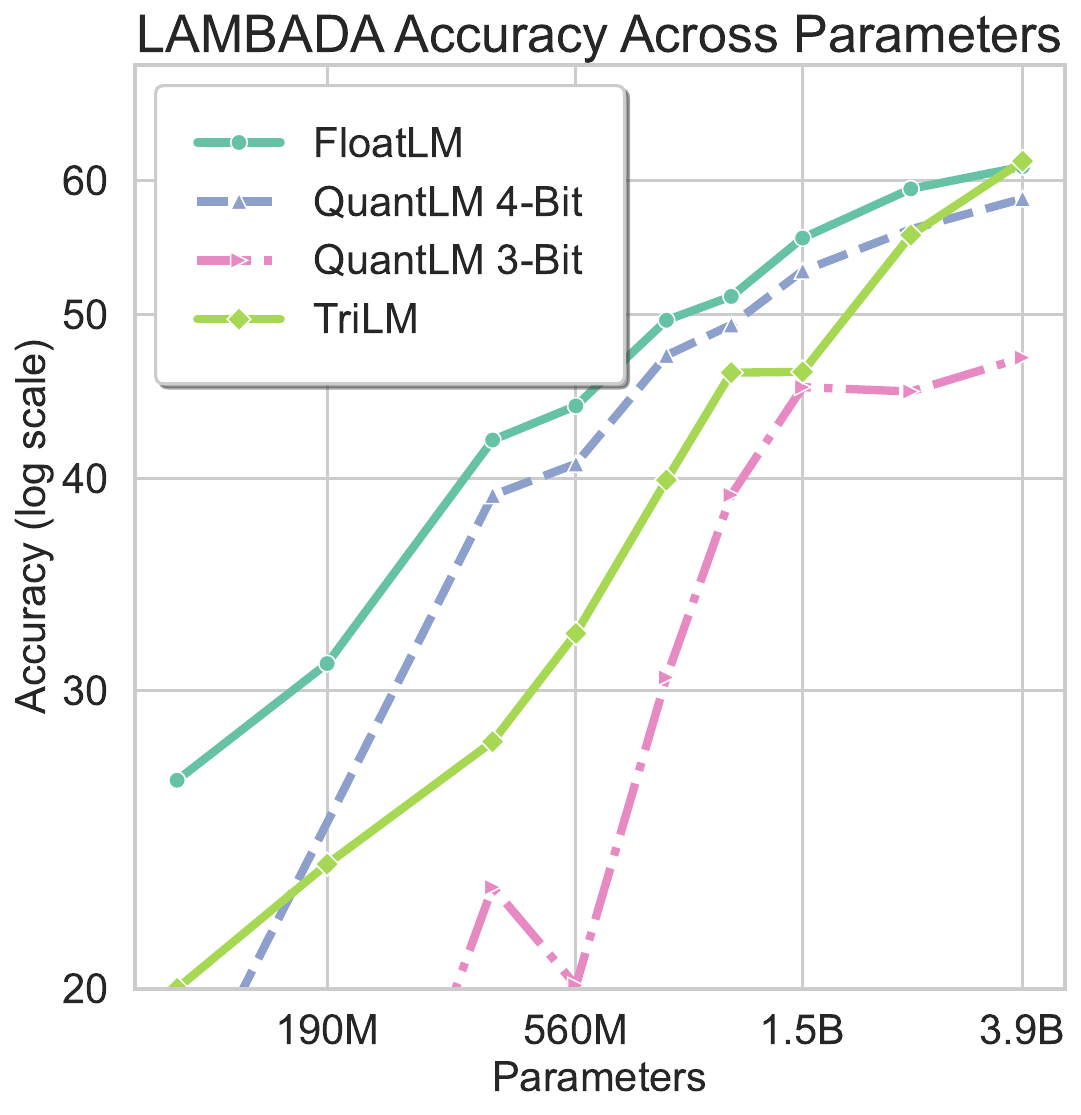}
            \caption{LAMBADA: acc vs parameters}
            \label{fig:LAMBADA_acc_across_params}
        \end{subfigure}
        \end{adjustbox}
        \caption{\footnotesize{Common Sense and Reasoning (C\&R) \& LAMBADA Accuracy for ternary TriLM, FP16 FloatLM and quantized QuantLM models across different model sizes, in bits and number of parameters. C\&R scores are averaged across 6 benchmarks. At 3B+ scales, TriLMs demonstrate better performance for their size than QuantLM and competitive performance to FloatLM of the same parameters. See Tables \ref{tab:evaluation_spectra_suite_part1.a}, \ref{tab:evaluation_spectra_suite_part1.b} and \ref{tab:evaluation_spectra_suite_part2} for details.}}  
        \vspace{-0.5 cm}
        \label{fig:commonsense_reasoning_comparisons}
    \end{figure}

    The computational power of GPUs, measured in FLOPs, is increasing exponentially, doubling approximately every 1.26 years. In contrast, memory capacity and bandwidth are growing at a slower pace, doubling every 2 and 2.9 years, respectively \citep{gholami2024aimemorywall}. This disparity highlights that computing capabilities are outpacing advances in memory technology. In Large Language Models (LLMs) inference, the primary bottlenecks are caused by model size (bits), which affects memory usage (memory capacity) and data transfer to processors (memory bandwidth). These issues are becoming more critical than the growing number of model parameters which affect the computational limits (FLOPs). For instance, state-of-the-art LLMs such as 340B  Nemotron  4 \citep{nvidia2024nemotron4340btechnicalreport} have sizes (in bits) exceeding the memory capacity of data center GPUs, such as 8xH100s. Token generation speed, or latency,  is now limited by memory bandwidth \citep{kim2024squeezellm}. Addressing these bottlenecks requires more expensive training, exceeding Chinchilla's compute-optimal regime \citep{hoffmann2022trainingcomputeoptimallargelanguage}, with less than 10B parameter models being trained on up to 15 trillion tokens \citep{llama2_paper, llama3modelcard, gemmateam2024gemmaopenmodelsbased}. Another widely used technique is post-training quantization during deployment \citep{zhu2023surveymodelcompressionlarge}; however, we demonstrate in Section \S\ref{results} that this approach is sub-optimal.

In post-training quantization, LLMs initially trained using 16-bit floating point precision (referred to as FloatLMs) undergo parameter quantization, and the resulting models are termed QuantLMs.
    % In post-training quantization, LLMs initially trained in 16-bit floating point (FP16/BF16) format (referred to as FloatLM) have their parameters quantized, i.e. converted to a smaller bitwidth after training; we refer to the resulting models as  QuantLMs.
 These models use optimized kernels for deployment, offering speedups nearly proportional to the compression factor \citep{frantar2024marlin}. However, quantizing to very low bitwidths creates a significant mismatch between the representations of pretrained FloatLM and the deployable QuantLM, resulting in undesired behavior and quality degradation \citep{li2024evaluatingquantizedlargelanguage, huang2024goodlowbitquantizedllama3}. Some state-of-the-art methods  \citep{gptq,egiazarian2024extremecompressionlargelanguage} mitigate this issue by using calibration and re-training data from target domains; however, this approach increases sensitivity to the calibration data. For instance, seemingly simple choices, like whether to length-normalize the calibration data, can significantly impact QuantLM's performance \citep{malinovskii2024pvtuningstraightthroughestimationextreme}. Other works have observed that QuantLM at 4 bits (4-bit QuantLMs) have about 65\% lower knowledge capacity per parameter compared to trained and aligned FloatLMs \citep{allenzhu2024physicslanguagemodels33}.
    
    An alternate approach to reducing model bitsize while maintaining parameter count involves training neural networks with low effective bitwidths \citep{zhou2018dorefanettraininglowbitwidth}. This approach offers compression benefits beyond post-training quantization without its associated drawbacks.  While low-bitwidth approaches typically employ binary (1-bit) or ternary quantization (1.58-bit), binary quantization generally underperforms compared to regular FP16 models \citep{liu2023binaryternarynaturallanguage} (see Appendix \S\ref{appendix:binary_vs_ternary_LLMs}).
    In contrast, ternary modeling can match performance while significantly reducing memory requirements (as we demonstrate in section \S\ref{results}). 
    % Consequently, this paper focuses on ternary networks.
    % For instance, BitNet b1.58 \citep{bitnet_b1.58} demonstrates that LLMs trained from scratch with low effective bitwidths (1.58 bits) follow similar scaling laws as FloatLMs \citep{kaplan2020scalinglawsneurallanguage} and perform competitively at the 3B+ parameter scale.
    Nevertheless, \emph{the relative performance of pretrained low-bitwidth language models compared to QuantLMs across similar sizes (in bits) and similar parameter counts remains unclear}. This is a crucial unanswered question, given the high inference costs during the deployment of very large-scale LLMs. Moreover, \emph{the training dynamics and scaling law of these low-bitwidth models are also poorly understood, partly due to the lack of publicly available systematic suites and associated comparative studies}. 
    
    % The aforementioned challenges were the primary motivation for our work, which resulted in the following set of contributions presented in this paper.

    Motivated by these challenges, we make the following contributions in this paper:

    % \paragraph{Demonstrating advantages of TriLM architecture and training dynamics for Ternary Language Modeling.} We empirically demonstrate the superiority of  TriLM's approach over BitNet b1.58, despite being simpler and more stable. We also highlight the critical role of its optimization schedule and discuss TriLM's training dynamics, including a sudden loss drop at the halfway point and accelerated convergence in the final third. All TriLM models were trained on the same data in the same order. We release over 500 intermediate checkpoints from TriLMs and FloatLMs in the Spectra suite.
    \paragraph{Feasibility and Scalability of Training Ternary Language Models (TriLMs)} We discuss the deployment advantages (in section \ref{subsec:memory_bottleneck}) and theoretical feasibility (in section \ref{subsec:lowbit_entropy}) of training low-bitwidth models at scale. We then introduce ternary language models (TriLMs) and systematically study their scaling laws compared to FloatLMs. Our analysis reveals that TriLMs offer better scaling behavior in terms of model size, measured in bits (refer to Section \ref{subsec:scaling_law}). Moreover, as the number of parameters increases, the difference in validation loss between TriLMs and FloatLMs becomes insignificant, indicating TriLM's competitive performance at scale.

    % and, match the validation losses of FloatLMs at a scale of 103 billion parameters for given dataset of 300B tokens. 
       % (in bits)  (see Section \ref{subsec:scaling_law}).

    \paragraph{Spectra LLM suite.} We present \textbf{Spectra}, the first open suite of LLMs spanning many bit-widths. This suite includes FloatLMs, the corresponding  QuantLMs at 3, 4, 6, and 8 bits, and ternary LLMs (TriLMs). The latter uses ternary weights \{-1, 0, +1\}.  The suite features 9 models ranging from 99M to 3.9B parameters, all trained on the same 300B token dataset, totalling 54 models. We believe that the Spectra LLM suite makes a valuable contribution to the LLM research community by facilitating comparative studies, exploring the scalability and efficiency of ternary modeling, and improving interpretability from neuronal to connection levels.
    
    %, as it enables comparative studies, examines ternary modeling's scalability and efficiency, and enhances interpretability from neuronal to connection levels.

   % we can talk about showing advantages of teranry over binary and then superiority of our ternery method.

    \paragraph{Evaluation and comparative analysis of TriLMs, FloatLMs, and QuantLMs at different scales.} We evaluate TriLMs, FloatLMs, and QuantLMs across multiple benchmarks, spanning commonsense, reasoning, knowledge capacity and toxicity. 
    At the billion parameter scale, TriLMs consistently outperform their QuantLM and FloatLM counterparts of the same bit-size across all the aforementioned benchmarks (see Figure \ref{fig:commonsense_reasoning_comparisons}).
    % Across commonsense and reasoning as well as knowledge capacity, TriLMs at the billion-parameter scale consistently outperform their QuantLMs and FloatLMs counterparts for a given bit size (see Figure \ref{fig:commonsense_reasoning_comparisons}). 
    Surprisingly, TriLM 3.9B matches the performance of FloatLM 3.9B across all benchmarks, despite getting a higher perplexity and being $5.9$ times smaller in bitsize.

    However, we also note that certain challenges remain in TriLMs. For instance, TriLM 3.9B exhibits the same level of toxicity and stereotyping as FloatLM 3.9B, significantly higher than a similarly sized FloatLM 830M when measured in bits.  
    While TriLM 3.9B and FloatLM 3.9B show similar validation perplexity on some datasets, such as  Penn Tree Bank and Lambada, a gap persists at this scale on web corpora, both in-domain (i.e., on a test subset of SlimPajama, the same domain used to train the models) and out-of-domain (e.g., on Dolma, C4 and RefinedWeb datasets). We provide detailed perplexity results in the section  \S\ref{appendix:Benchmark_Details::subsection:perplexity_on_other_datasets}.

\section{Motivation for Low-Bitwidt Models}

  \subsection{Memory Bottlenecks and Language Model Deployment} \label{subsec:memory_bottleneck} 
 % Recently observations \citep{gholami2024aimemorywall} suggest that, given the slower pace of improvements in memory and communication as compared to compute (FLOPs), the bottleneck continues to shift away from computation towards memory-related characteristics of hardware for deploying large language models. This shift underscores the importance of exploring solutions that directly address memory constraints. 

\textbf{Experimental Setup:} First, we, explore the impact of training low-bitwidth language models on deployment focussing on addressing memory bottlenecks.
 Our analysis includes transformer configurations from the LLaMa family \citep{llama1_paper,llama2_paper}. As larger vocabularies in LLMs are becoming increasingly common for efficient multilingual modeling, we use a vocabulary size of 128k from LLaMa 3 \citep{llama3modelcard} for our analysis. We assume the Embedding and LM Head weights are retained in Half-Precision across all bitwidths for these analyses.

\begin{figure}[htbp]
    \centering
    \begin{adjustbox}{width=0.8\textwidth}
    \begin{subfigure}[b]{0.5\textwidth}
        \includegraphics[width=1.05\textwidth]{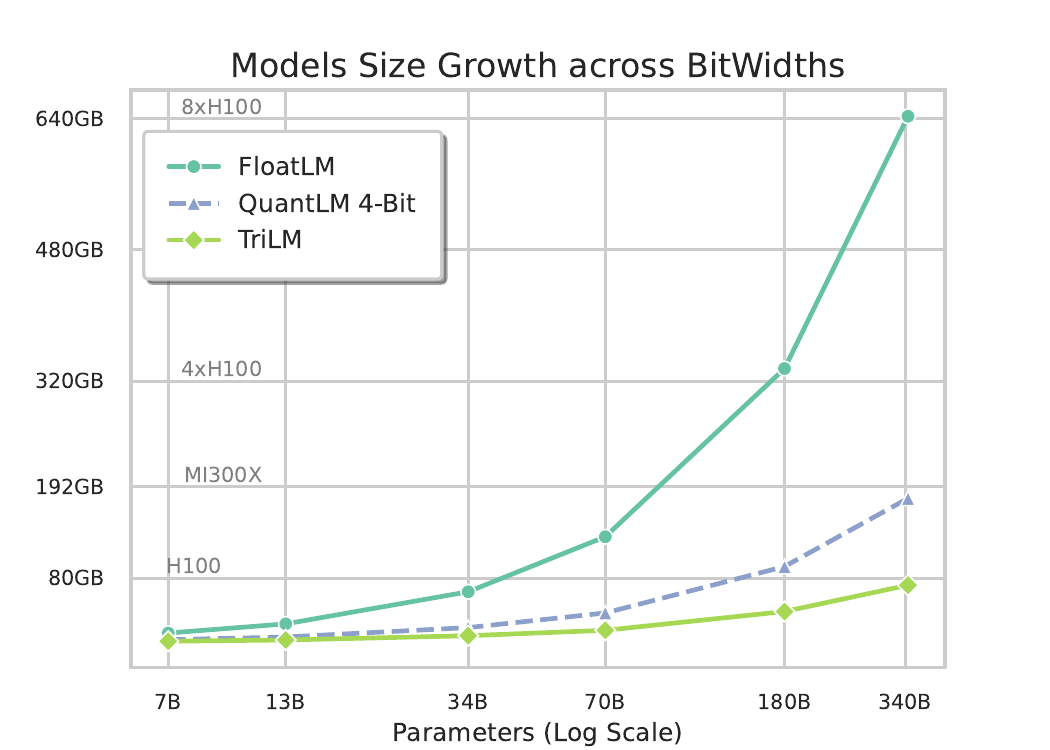}
        \caption{Model Size growth at various BitWidths}
        \label{fig:model_size_growth_at_scale_across_bitwidth}
    \end{subfigure}%
    \begin{subfigure}[b]{0.49\textwidth}
        \includegraphics[width=0.94\textwidth]{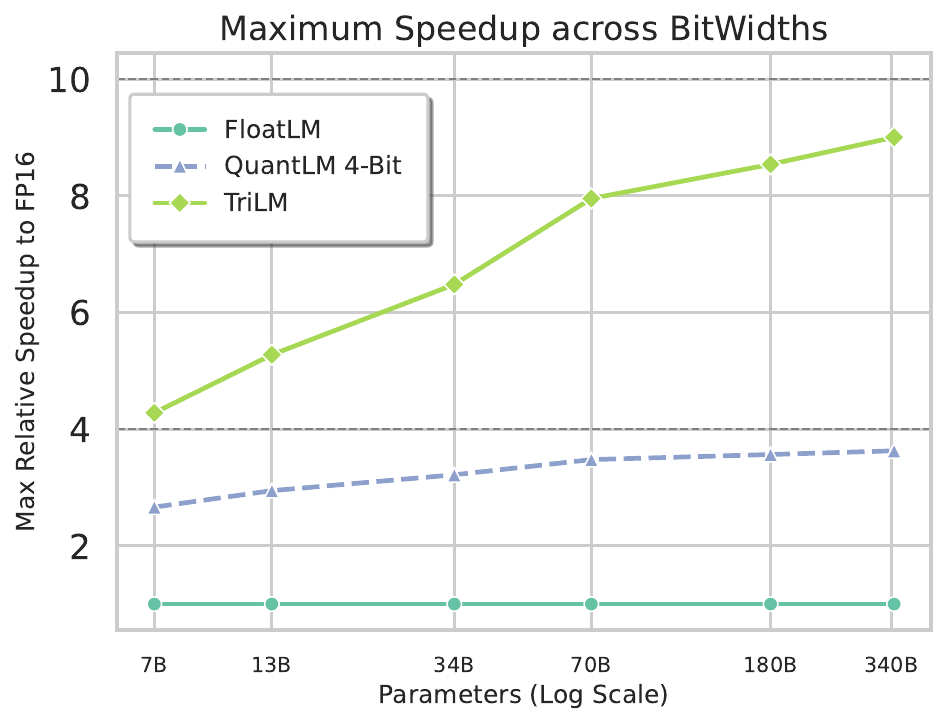}
        \caption{Maximum Possible Speedup at different BitWidths}
        \label{fig:maximum_speedup_at_scale_across_bitwidth}
    \end{subfigure}
    \end{adjustbox}
    \caption{\footnotesize{Expected gains from low bitwidth modeling. TriLMs can fit over 300B parameters on a single H100 and achieve up to a theoretical maximum of 10x faster autoregressive decoding compared to FloatLM.}}
    \label{fig:model_growth_and_speedup_trends}
\end{figure}

    % \begin{figure}[htbp]
    %     \centering
    %     \begin{subfigure}[b]{0.5\textwidth}
    %         \includegraphics[width=0.94\textwidth]{figures/trends_decreasing_memory_to_flop_ratio.pdf}
    %         \caption{Memory Capacity vs Peak FLOPS \label{fig:trends_decreasing_memory_to_flop_ratio}}
    %     \end{subfigure}
    %     \begin{subfigure}[b]{0.49\textwidth}
    %         \includegraphics[width=1.05\textwidth]{figures/model_size_growth_across_bitwidths.pdf}
    %         \caption{Model Size growth at various BitWidths \label{fig:model_size_growth_at_scale_across_bitwidth}}
    %     \end{subfigure}
    %     \begin{subfigure}[b]{0.5\textwidth}
    %         \includegraphics[width= 0.94\textwidth]{figures/trends_decreasing_bytes_to_flop_ratio.pdf}
    %         \caption{Peak Memory Bandwidth vs Peak FLOPS \label{fig:trends_decreasing_bytes_to_flop_ratio}}
    %     \end{subfigure}
    %     \begin{subfigure}[b]{0.49\textwidth}
    %         \includegraphics[width=0.94\textwidth]{figures/maximum_speedup_at_scale_across_bitwidth.pdf}
    %         \caption{Maximum Possible Speedup at different BitWidths \label{fig:maximum_speedup_at_scale_across_bitwidth}}
    %     \end{subfigure}
    %     \caption{Trends of Memory/FLOP and Bandwidth/FLOP across different (datacenter) GPGPUs and expected gains from low bitwidth modeling. TriLMs can fit over 300B parameters on a single H100 and achieve up to a theoretical maximum of 10x faster autoregressive decoding compared to FloatLM.}
    %     \label{fig:hardware_trends_and_efficiency_gains}
    % \end{figure}

\paragraph{Memory Capacity of GPGPUs and Model Size (in Bits):}

Figure \ref{fig:trends_decreasing_memory_to_flop_ratio} in Appendix \S\ref{app:memory_analysis_for_LLM} reveal that memory capacity has consistently lagged behind computational power across various accelerators, including recent hardware like Blackwell \citep{hardware_datasheet_nvidia_blackwell}, MI325X \citep{hardware_datasheet_amd_mi325x}, and Gaudi3 \citep{hardware_datasheet_intel_gaudi2_gaudi3}. This gap is further exacerbated by advanced computational techniques like Ampere sparse or FP8. As shown in Figure \ref{fig:model_size_growth_at_scale_across_bitwidth}, the model sizes (in GB) for TriLM, QuantLM 4-Bit, and FloatLM scale differently with parameter counts. For simplicity, the figure excludes overhead from KV Cache, activations, and compilation during model deployment. FloatLM reaches the memory capacity of a single H100 at 34B parameters, with larger models exceeding the capacity of multiple GPUs. In contrast, QuantLM 4-Bit supports up to 300B parameters on a single MI300X. TriLMs, with over 300B parameters and appropriate packing, can fit on a single H100, making them not only efficient for GPU deployment but also ideal for edge devices.

\paragraph{Memory bandwidth of GPGPUs and model inference speedup:}

Figure \ref{fig:trends_decreasing_bytes_to_flop_ratio} and Appendix \S\ref{app:memory_analysis_for_LLM} demonstrate the trends of Memory Bandwidth over FLOPs for accelerators over the years, highlighting the slower growth of memory bandwidth compared to computation. This trend, indicating a downward slope, aligns with \cite{kim2024squeezellm}'s establishment of the memory wall in autoregressive LLM computation. They found that the speed of token generation is bottlenecked by the rate at which data is fed from memory to processors, rather than the processing speed of the hardware. Consequently, the autoregressive decoding of LLM inference can theoretically achieve speedup proportional to its compression factor. Figure \ref{fig:maximum_speedup_at_scale_across_bitwidth} illustrates the maximum possible speedup for QuantLM 4-Bit and TriLM compared to FP16 at different parameter counts. Even at 7 billion parameters, TriLMs can be more than 4 times faster at autoregressive decoding than FloatLM and 2 times faster than QuantLM 4-bit. While QuantLM 4-Bit plateaus at a maximum possible speedup factor of 4x, TriLMs plateau much higher at 10x for FloatLM.

\vspace{-0.2cm}

\subsection{Low bits can capture weight variance effectively at scale}\label{subsec:lowbit_entropy}

In this section, we use information theory to support our hypothesis: \emph{as the number of parameters increases, training language models with low-bitwidth can effectively capture the necessary weight variance without significant information loss}. Assuming a fixed training dataset, we base this hypothesis on analyzing weight distributions in FloatLMs ranging from 99M to 3.9B parameters.

\begin{figure}[h]
    \centering
    \begin{minipage}{0.48\linewidth}
        \includegraphics[width=0.98\linewidth]{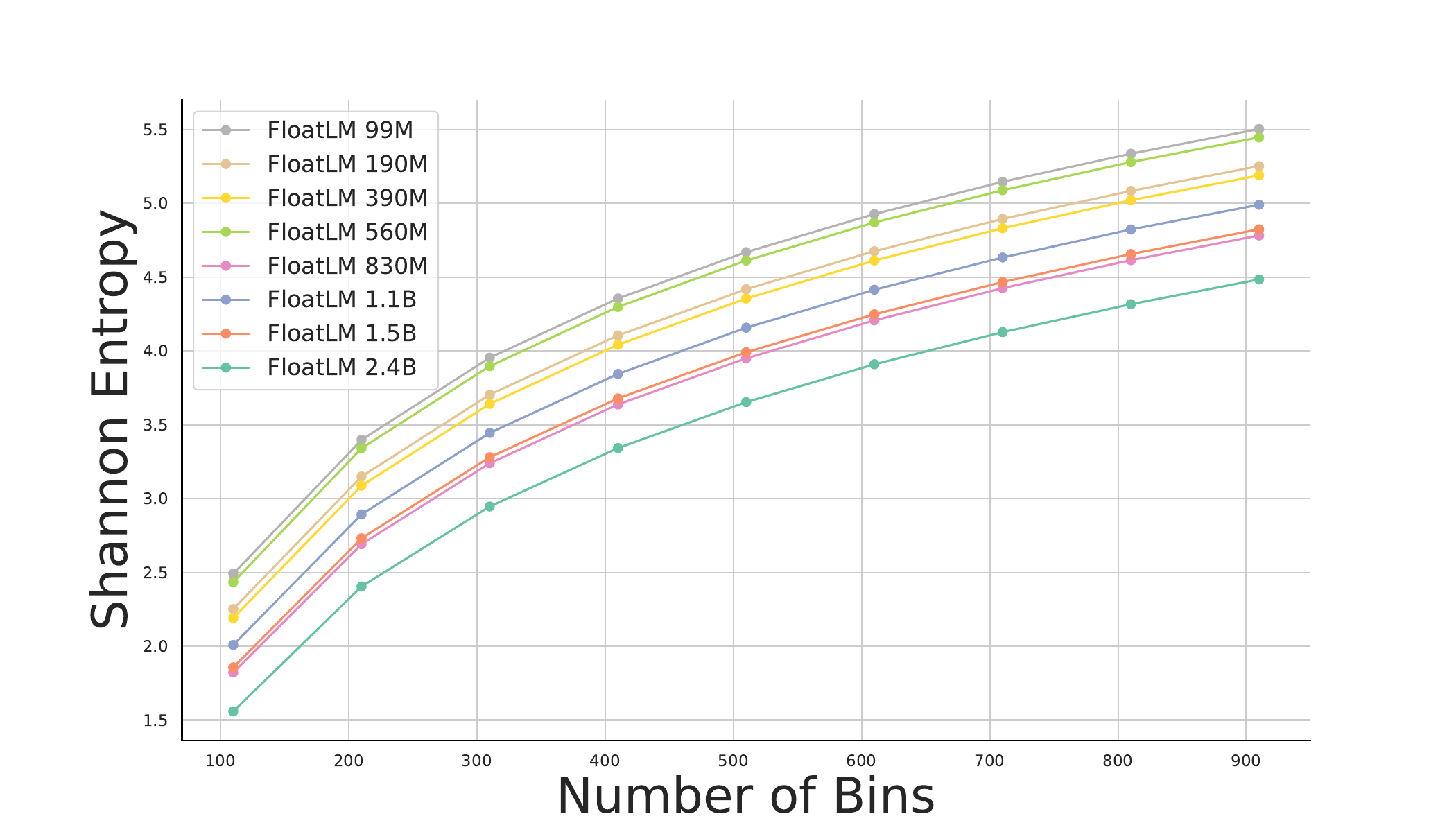}
        \caption{\footnotesize{Shannon entropy (in bits) of discretized weight distribution with increasing number of bins.}}
        \label{fig:shannon}
    \end{minipage} \hspace{0.02\textwidth}
    \begin{minipage}{0.48\linewidth}
        \includegraphics[width=0.98\linewidth]{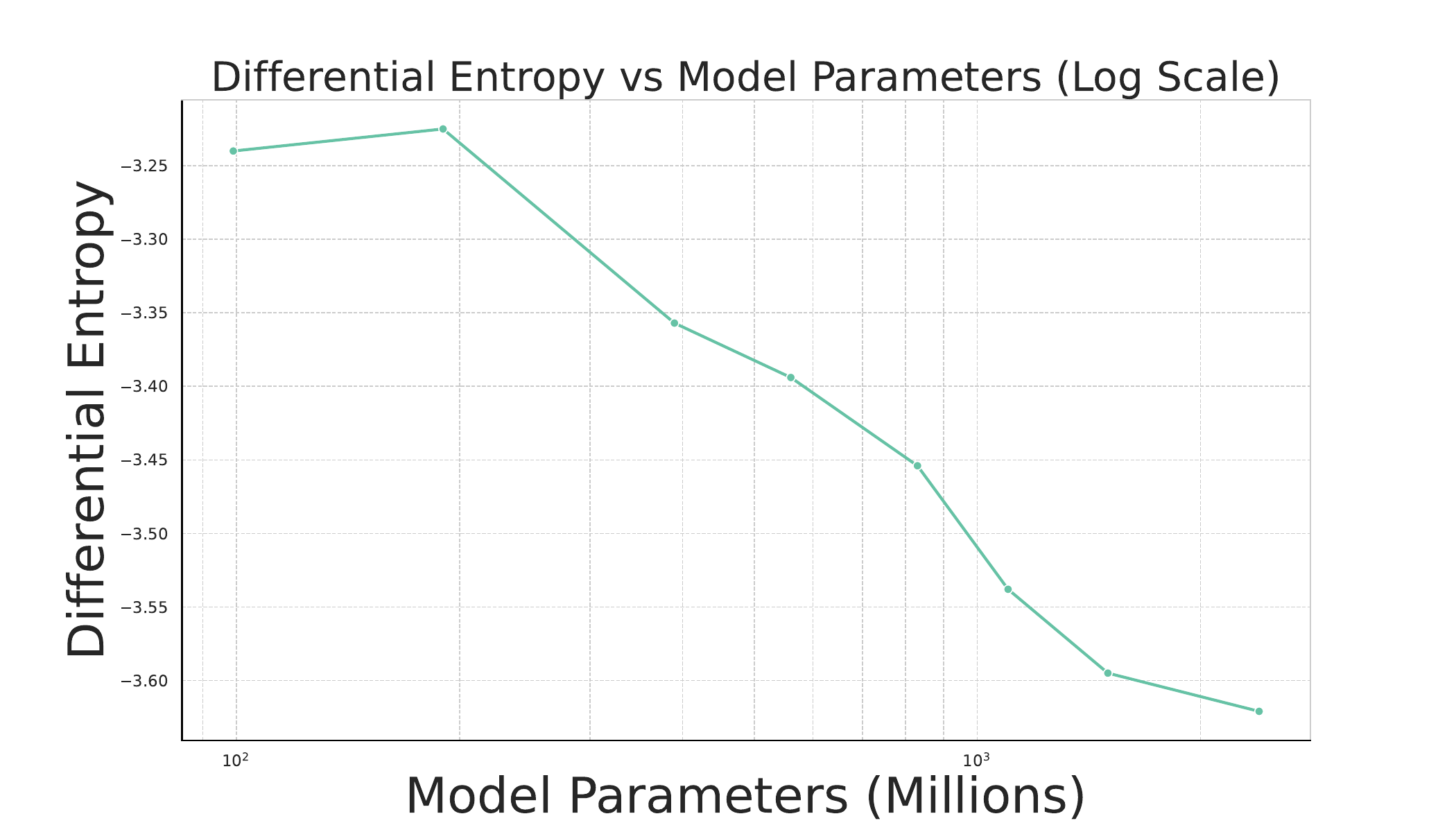}
        \caption{\footnotesize{Differential entropy of Gaussian fits on weight distributions across different scales.}}
        \label{fig:differential_entropy}
    \end{minipage}
\end{figure}

Assuming that the weights of a trained model follow a Gaussian distribution (see Appendix \S\ref{appdx:weight_distribution}), we fit a Gaussian to these weights to understand their statistical behavior across model scales. The differential entropy is calculated using the expression: $H(W) = \frac{1}{2} \log_2(2\pi e \sigma_{W}^2)$
where \(\sigma_{W}\) is the standard deviation of the weights \citep{papoulis2002probability}. As shown in Figure \ref{fig:differential_entropy}, differential entropy decreases with an increase in the number of parameters. This decrease indicates that the weights become more concentrated around the mean as the model size increases, suggesting higher predictability and less variability \citep{mackay2003information}. This reduced variability is due to overparameterization, which leads to redundancy in the weights \cite{zhang2017understanding, neyshabur2018role}. 

Additionally, we also use Shannon entropy, calculated by discretizing the weight distribution into (\(N\)) bins and computing $H_{Shannon} = -\sum_{i=1}^{N} p_i \log_2 p_i$, where \(p_i\) is the probability of the weights falling into the \(i\)-th bin. The probability is given by the normalized frequency of weights within each bin. Discretization allows us to apply discrete entropy measures to continuous data, with the number of bins determining the representation granularity. Shannon entropy measures the average minimum number of bits needed to encode the weight values based on their distribution, providing a quantifiable measure of their "information content" \citep{shannon1948mathematical}. Lower Shannon entropy indicates a more predictable, less diverse distribution that can be effectively encoded using fewer bits. Examining Shannon entropy across scales and bin sizes, we validate our finding that larger models require fewer bits for effective representation. As plotted in Figure \ref{fig:shannon}, there is a general trend of decreasing Shannon entropy with increasing parameter count for a given number of bins.

These analyses support the potential of low-bitwidth models to match full-precision performance, especially as model sizes grow. In Section \ref{subsec:scaling_law}, we further substantiate this by analyzing the scaling behavior of our low-bitwidth TriLMs compared to FloatLMs.

\subsection{Selecting the Appropriate Low-Bitwidth Model}
Selecting the appropriate quantization level is crucial when training low-bitwidth models, as it significantly impacts both computational efficiency and model performance. Binary models quantize weights to \{-1, 1\}, simplifying multiplication operations in neural networks to XOR operations. Such operations are highly efficient on digital hardware due to their basic logic gate implementation \citep{courbariaux2016binaryconnect}. XOR operations, which can be executed rapidly and with lower power consumption than traditional multiplications, make binary models especially appropriate for resource-constrained environments \citep{hubara2017binarized}. Ternary models extend this concept by incorporating a zero state, allowing weights to be \{-1, 0, +1\}. This modification not only maintains computational simplicity by replacing multiplications with additions and subtractions \citep{li2016ternary} but also leverages the sparsity of neural networks to enhance computational efficiency by eliminating calculations where weights are zero \citep{zhu2017trained}.

In contrast, models quantized to 2, 3, or 4 bits involve more complex arithmetic operations, including full multiplications, which are computationally more demanding than those required for binary and ternary models \citep{zhou2016dorefanet}. While these higher bitwidth models offer more detailed weight representations, they do so at the expense of greater computational resource usage. Therefore, this work only considers binary and ternary language models for further analysis.

While binary quantization minimizes computational complexity, it often results in significant performance degradation, failing to meet the standards of practical applications \citep{rastegari2016xnor}. In contrast, ternary quantization introduces a zero state that more accurately approximates weight distributions without substantially increasing computational demands \citep{li2016ternary}. This capability makes ternary models particularly advantageous, as they maintain performance levels closer to those of full-precision models while still achieving significant efficiency gains \citep{zhu2017trained}. Our observations align with these, as the scaling trends for given data size for our TriLMs consistently outperform those of BiLMs (Binary LLMs) across all parameter counts and bit sizes considered in this work, as shown in Appendix \S\ref{appendix:binary_vs_ternary_LLMs} and Figure \ref{fig:scaling_law_curves_binary_ternary}. As a result, this work primarily focuses on the TriLM model, which we describe in the following section.

\vspace{-0.2cm}

\section{TriLM: Ternary Language Model}

In this section, we present the architectural and optimization details of the TriLM (Ternary Language Model). We discuss the training and inference processes, as well as the optimization schedule.

    % \begin{figure}[h]
    %         \centering
    %         \vspace{-2 mm}
    %         \includegraphics[width=0.87\linewidth]{figures/TriLM_Architecture.pdf}
    %         \caption{ The computational flow of forward, backward, and inference processes in TriLM's linear layer with N-Way model parallelism.}

    %         \vspace{-2 mm}
    %         \label{fig:trilm_vs_floatlm_vs_bitnet_architecture}
    % \end{figure}

    \subsection{Architecture} \label{section:TriLMs_and_FloatLMs::subsection:Architecture}

% In this section, we present the architectural and optimization details of the TriLM (Ternary Language Model). The subsections provide an in-depth analysis of the architectural choices distinguishing TriLM from BitNet, as well as the optimization strategies employed during training.

    \begin{figure}[h]
            \centering
            \vspace{-2 mm}
            \includegraphics[width=\linewidth]{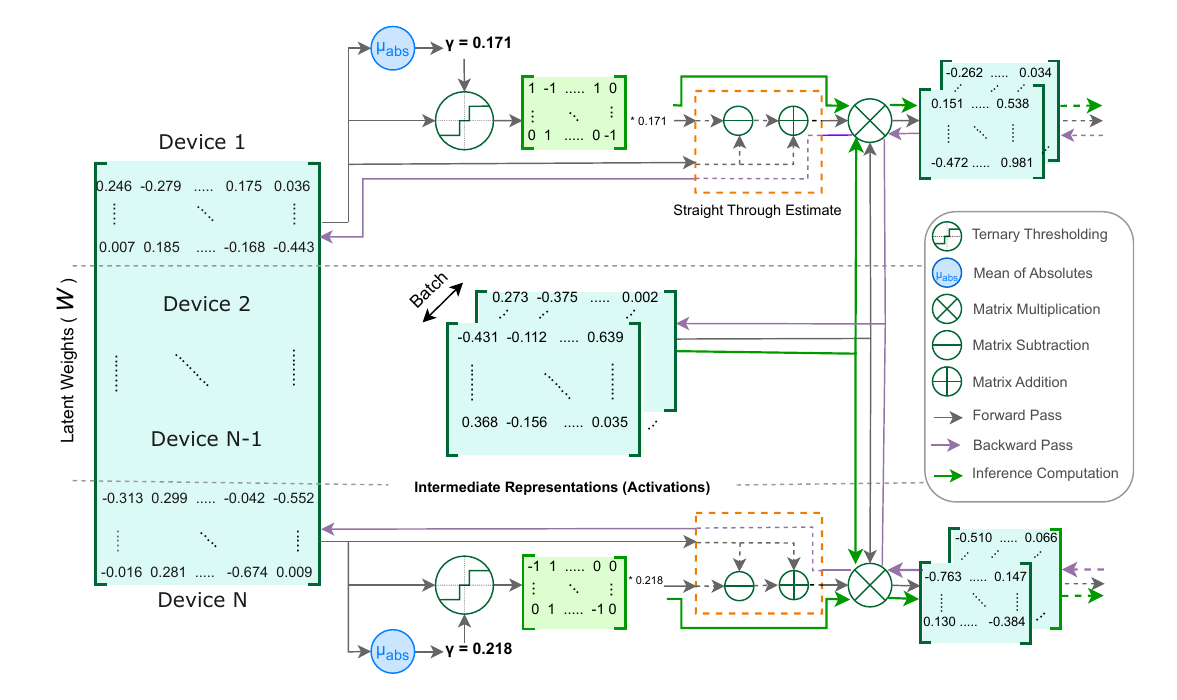}
            \caption{ \footnotesize{The computational flow of forward, backward, and inference processes in TriLM's linear layer with N-Way model parallelism.}}
            \vspace{-0.6 cm}
            \label{fig:compute_flow_diagram}
    \end{figure}

    \vspace{-0.2 cm}
    % \subsection{Architecture} \label{section:TriLMs_and_FloatLMs::subsection:Architecture}
    
    TriLM is a LLaMa-style \citep{llama1_paper} autoregressive transformers \citep{attention_is_all_you_need} model with RMSNorm \citep{rmsnorm_paper}, SwiGLU Gated MLP \citep{glu_variants_improve_transformer}, Rotary Position Embedding (RoPE) \citep{rope_rotary_position_embedding_paper}, Multi-Headed Attention and no bias terms. In TriLMs, we represent the weights of all the linear layers in one of three possible ternary states $\{-1, 0, 1\}$, along with an additional floating-point number called `scale value' shared across the matrix. During training, we maintain the \textit{latent} (or master) weights in floating point precision, allowing for the accumulation of small updates over iterations that eventually contribute to a transition in the estimated ternary state of a parameter. As shown in Figure \ref{fig:compute_flow_diagram}, during the forward pass, we ternarize the floating point latent weights on-the-fly. This process involves calculating the scale value to the absolute mean of the latent weights. After scaling, we estimate the ternary state of each parameter by rounding off to the nearest ternary state. In the backward pass, we use a straight-through estimator to compute gradients of the floating point latent weights \citep{bengio2013estimating}. Since we only need the scale values and the ternarized states during inference, we achieve a significant reduction in both model size and inference time at larger scales compared to FloatLMs. We provide a formal description of these forward pass, backward pass, and inference time equations in the Section (\S\ref{appendix:architecture_and_training_details::subsection:equations}). We represent the embedding and language model head in half-precision floating point across all our experiments.

    Since, training of TriLMs requires on-the-fly computation of scale values, synchronizing for a single scalar across devices in model parallel training \citep{megatron_paper} can cause significant communication overhead. To address this, we allow each device to independently compute these scale values over its own matrix shard. This approach introduces additional artifacts, where the number of scalar values for each matrix is the same as the degree of model parallelism used during training (see section \ref{section:TriLMs_and_FloatLMs::subsection:Implementation_Artifacts}). However, the impact on modelsize is negligible, for matrices with millions of parameters, we only add $6$ scalar values each. 

    Concurrent research on low-bit LLMs, like BitNet 1.58 \citep{bitnet_b1.58}, also demonstrates the feasibility of training LLMs with ternary weights. Our experiments demonstrate that TriLM’s architecture not only outperforms BitNet b1.58 but is also simpler and more stable. Moreover, both the larger BitNet 1.3B model presented in their paper and our replication of the BitNet 1.1B model underperform compared to our TriLM 1.1B on commonsense and reasoning benchmarks. We provide further details on these comparisons in Appendix \ref{section:differences_from_bitNet_architecture}.

        % Following are the key differences in TriLM's architecture. We follow GPT3's Pre-Normalization \citep{brown2020language} approach to normalize before each linear layer - this was observed to be crucial for stable training in FP16. Thus, normalization is done twice in each transformer layer, at the input representations to the two sub-layers - attention and Gated MLP. This is in contrast to BitNet, where before each linear layer (i.e. 4-7 times per transformer layer depending on the implementation), the activation (or intermediate representations) are normalized, scaled and quantized to 8 bits. We use RMSNorm with a scale parameter over the parameterless RMSNorm.

        \subsection{Forward Pass, Backward Pass and Inference Equations}  \label{appendix:architecture_and_training_details::subsection:equations}

        Table \ref{table:trilm_vs_floatlm_equations} shows the equations for TriLM, BiLM (Binary Langauge Model) and FloatLM for forward pass, backward pass and inference.
            
            \begin{table}[h]
                \centering
                \begin{adjustbox}{width=\textwidth}
                \begin{tabular}{cccc}
                    \toprule
                    Type & Forward Pass & Backward Pass & Inference \\
                    \midrule
                    FloatLM & $Y=XW^T$ & 
                        \begin{tabular}{@{}c@{}}
                            $\frac{\partial L}{\partial X} = \frac{\partial L}{\partial Y} W$ \\
                            $\frac{\partial L}{\partial W} = \frac{\partial L}{\partial Y}^T X$
                        \end{tabular} 
                     & $Y = X W^T$ \\
                    \midrule
                    TriLM & 
                        \begin{tabular}{@{}c@{}}
                            $\gamma = \epsilon + \frac{1}{nm} \sum_{i=1}^{n}\sum_{j=1}^{m}|W_{ij}|$ \\
     $\widehat{W_{ij}} = \text{round}\left(\min\left(\max\left(\frac{W_{ij}}{\gamma}, -1\right), 1\right)\right)$ \\
                            $\widetilde{W_{ij}} = \gamma \widehat{W_{ij}}$ \\
                            $Y = X \widetilde{W}^T$
                        \end{tabular} &
                        \begin{tabular}{@{}c@{}}
                            $\frac{\partial L}{\partial X} = \frac{\partial L}{\partial Y} \widetilde{W}$ \\
                            $\frac{\partial L}{\partial W} = \frac{\partial L}{\partial Y}^T X$
                        \end{tabular} &
                        \begin{tabular}{@{}c@{}}
                            Compute $\widehat{W}$ and $\gamma$ once and cache  \\
                            $\widetilde{W_{ij}} = \gamma \widehat{W_{ij}}$ \\
                            $Y = X \widetilde{W}^T$
                        \end{tabular} \\
                    \midrule
                    BiLM & 
                    \begin{tabular}{@{}c@{}}
                        $\alpha = \frac{1}{nm} \sum_{i=1}^{n}\sum_{j=1}^{m}|W_{ij}|$ \\
                        $\widehat{W_{ij}} = \text{sign}(W_{ij} - \frac{1}{nm} \sum_{i=1}^{n}\sum_{j=1}^{m}W_{ij})$ \\
                        $\widetilde{W_{ij}} = \alpha \widehat{W_{ij}}$ \\
                        $Y = X \widetilde{W}^T$
                    \end{tabular} &
                    \begin{tabular}{@{}c@{}}
                        $\frac{\partial L}{\partial X} = \frac{\partial L}{\partial Y} \widetilde{W}$ \\
                        $\frac{\partial L}{\partial W} = \frac{\partial L}{\partial Y}^T X$
                    \end{tabular} &
                    \begin{tabular}{@{}c@{}}
                        Compute $\widehat{W}$ and $\alpha$ once and cache \\
                        $\widetilde{W_{ij}} = \alpha \widehat{W_{ij}}$ \\
                        $Y = X \widetilde{W}^T$
                    \end{tabular} \\
                \bottomrule
            \end{tabular}
                \end{adjustbox}
                \vspace{0.2 cm}
                \caption{\footnotesize{Equations in the Linear Layer of TriLMs and FloatLMs.}}
                \label{table:trilm_vs_floatlm_equations}
            \end{table}

        \paragraph{Reason for restricting the quantization approach to linear weights in TriLMs.} 
        In developing extremely large language models like TriLMs, a key architectural strategy is to quantize only the linear layer weights while keeping the embedding layers and language model head in higher precision. This is driven by the need to reduce model size while maintaining performance. Linear layers (dense layers) hold the bulk of the parameters in transformer models \citep{attention_is_all_you_need}. Quantizing these weights to ternary states significantly reduces the model size, facilitating deployment on memory-constrained hardware. However, the embedding layers and language model head remain in higher precision (e.g., half-precision floating point) to preserve critical functions in language understanding and generation. Embedding layers encode important semantic and syntactic information, and quantizing them would degrade performance \citep{mikolov2013efficientestimationwordrepresentations}. Similarly, the language model head, which maps internal representations to the vocabulary space, requires high precision to maintain prediction quality \citep{press2017usingoutputembeddingimprove}

    \subsection{Optimization Schedule}\label{section:TriLMs_and_FloatLMs::subsection:Optimization_Schedule}
      
\begin{wrapfigure}{l}{0.5\textwidth}
                \centering
                    \includegraphics[width=0.85\linewidth]{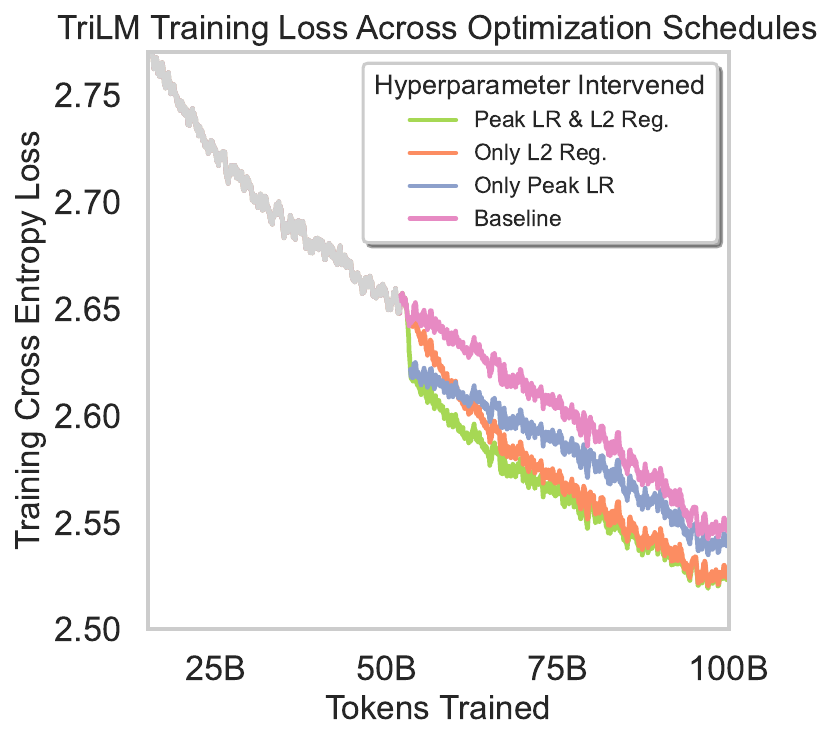}
                \caption{\footnotesize{Training loss for a 1.1B parameter TriLM, across different optimization schedules.}}
                \vspace{-4 mm}
                \label{fig:learning_schedule_ablation_training_loss_across_steps}
        \end{wrapfigure}

             Optimization of low bitwidth neural networks (such as in Quantization Aware Training) \citep{llm_qat, pb_llm, binary_nn_from_scratch, binary_vit} requires a set of considerations like higher initial learning rate and reduced weight decay. Our optimization schedule for TriLM closely follows that of BitNet \citep{bitnet_b1.58} consisting of two interventions in a vanilla linear decay learning rate scheduling with warmup and weight decay (L2 Regularization). (1) \textit{Peak LR} - at roughly the halfway point, we reduce the peak learning rate. (2) \textit{L2 Reg.} - at roughly two-thirds of the training, we remove the weight decay regularization as ternarization provides sufficient regularization \citep{courbariaux2016binaryconnecttrainingdeepneural}. Figure \ref{fig:learning_schedule_ablation_training_loss_across_steps} demonstrates the ablation run performed for a 1.1B parameter model on 100B tokens with both, only one and neither of these interventions. 

 % Similar phenomena have been documented in training schedules characterized by brief periods of rapid learning rate decay, such as those described in MiniCPM (Hu et al., 2024). Conversely, the removal of L2 regularization, or weight decay, led to an accelerated convergence rate, which can produce effects analogous to reducing the peak learning rate, also resulting in a swift decrease in loss.

Among the four experimental runs, we observed that the lowest final training loss occurred when both L2 regularization and peak learning rate (LR) adjustments were implemented. This was closely followed by the scenario in which only L2 regularization was adjusted, and subsequently by the condition where only peak LR was modified.  Notably, discontinuing the peak learning rate at the midpoint of training resulted in a significant and rapid decline in training loss.  Similar phenomena have been documented in training schedules characterized by brief periods of rapid learning rate decay, such as those described in MiniCPM\citep{hu2024minicpmunveilingpotentialsmall}.  Conversely, the removal of L2 regularization, or weight decay, led to an accelerated convergence rate, which can produce effects analogous to reducing the peak learning rate, also resulting in a swift decrease in loss. These relative training loss observation at 100B tokens also go hand in hand with relative downstream performance across commonsense and reasoning tasks, which are listed in Table \ref{tab:commonsense1_scores_evaluation_ablation_part1} and \ref{tab:commonsense1_scores_evaluation_ablation_part2}. Consequently, we have established a fixed optimization schedule for the TriLM, which entails dropping the peak learning rate at the halfway point while removing weight decay at the two-thirds mark.
            
            % Thus, we fix the TriLM optimization schedule. We drop in the peak learning rate at halfway and weight decay is removed at the two-thirds mark.
        
% \subsection{Ternary vs Binary Large Language Models}

% In addition to our work on Ternary Language Models (TriLMs), we also explore Binary Language Models (BiLMs), where the weights in the linear layers are constrained to binary values, \{-1, 1\}, with an accompanying floating-point scaling factor. 

% In the presented TriLM architecture, we adapt the quantization function to a binarization formulation as proposed by \citet{courbariaux2016binaryconnecttrainingdeepneural}. We compare the training curves of both TriLM and BiLM at three different parameter counts. The figure shows that BiLM always underperforms compared to TriLM at all sizes (measured in bits) and scales (parameter count). 

% Based on above results, we decided to scale TriLM further to analyze trends at scale and compare it with other post-training quantization methods and Half-precision models.

% Similar to TriLM, we also train binary language models (BiLMs) where each weight in the linear layers are constraint the weights to \{-1, 1\}, along with a floating-point scale. In the TriLM architecture, we adapt the quantization function to a binarization formulation as proposed by \citep{ courbariaux2016binaryconnecttrainingdeepneural}. The formal definition is provided in Table \ref{table:trilm_vs_floatlm_equations}.

\section{Spectra suite: Spanning Parameters and Bitwidths}
    The Spectra suite includes comprehensive families of Large language models designed to span different parameter counts and bit-widths. This suite includes three main model families: \textit{TriLMs}, \textit{FloatLMs}, and \textit{QuantLMs} (3, 4, 6, and 8 bits). Drawing inspiration from established model suites such as those by \citep{biderman2023pythia, liu2023llm360, groeneveld2024olmoacceleratingsciencelanguage}, Spectra aims to facilitate scientific research on low-bitwidth LLMs.

    \subsection{Overview of Spectra Suite}
                
The Spectra suite stands out with the following key properties:

        \begin{enumerate}
            \item \textbf{Scale:} The suite spans a broad spectrum of scales across parameter count (99M to 3.9B), sizes ($9*10^8$ to $6.4*10^{10}$ bits) and bitwidths (1.58 bits to 16 bits). 
            
            \item \textbf{Uniform Training:} All the TriLMs and FloatLMs are trained on a identical data sequences, specifically a 300B subset of Slim Pajama \citep{slimpajama_dataset} dataset, ensuring training consistency. Data ordering and batch sizes are also kept consistent within each model family to support reproducibility and comparability in research.   QuantLMs undergo quantization using the same calibration data, maintaining uniformity in model quantization procedures.
            
            \item \textbf{Public Accessibility:} The training data and intermediate checkpoints are publicly available for study.
            \item \textbf{Consistent Model Size Mapping:} All models across the families maintain a consistent one-to-one mapping for parameter count.
        \end{enumerate}
        
\begin{wrapfigure}{r}
        {0.45\textwidth}
        \centering
        \includegraphics[width=0.8\linewidth]{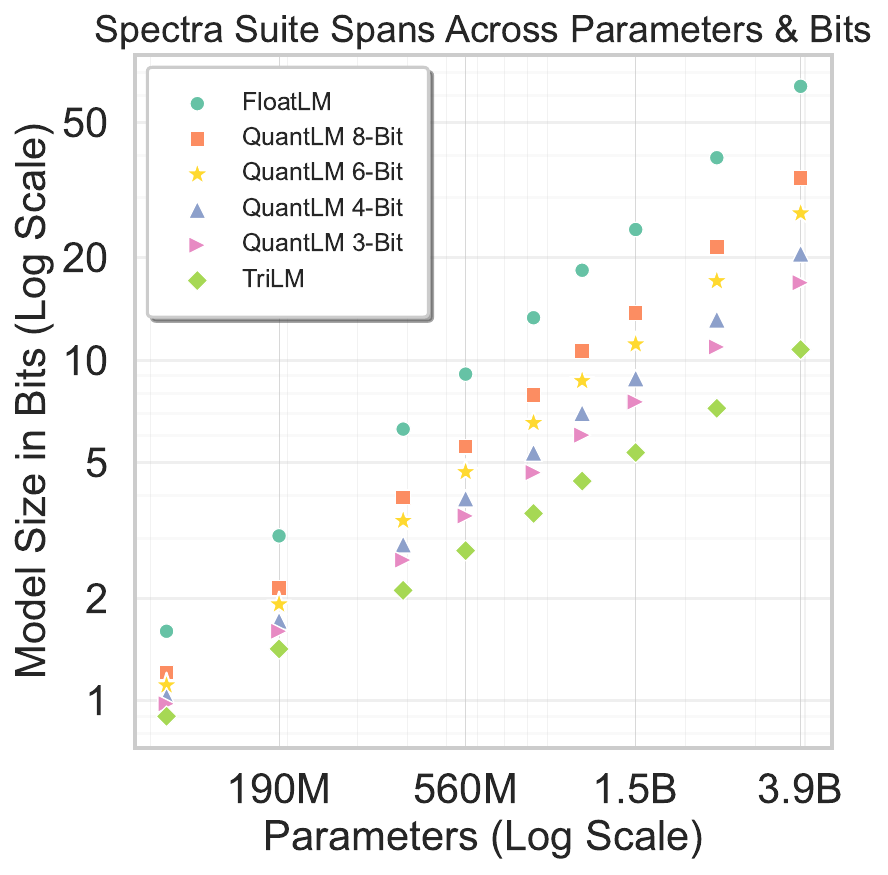}
        \caption{\footnotesize{The Spectra Suite spans across two dimensions of parameters and scale. Each point corresponds to a LLM in the spectra suite.}} 
\label{fig:spectra_lm_suite_spans_across_parameters_and_bits}
\vspace{- 0.5 cm}
\end{wrapfigure}

        Figure \ref{fig:spectra_lm_suite_spans_across_parameters_and_bits} demonstrates the Spectra LM suite spanning across two dimensions - size (bits) and parameters. For each parameter count, we have 6 models across different bitwidths. Due to availability of FloatLM, Spectra can easily be extended with new QuantLMs by using different Post Training Quantization methods. 

        The details on dataset and tokenizer, pretraining setup, and hyperparameters across all the models are detailed in the Appendix \S\ref{appendix:architecture_and_training_details}, while information on other families is covered in the next section.

        \subsection{FloatLMs and QuantLMs}
        \textbf{Family of FloatLMs.} We utilize  LLaMa-style \citep{llama1_paper} architecture akin to TriLM. In FloatLMs, parameters in the weight matrices of linear layers are represented as floating-point numbers (FP16/BF16). The optimization schedule for FloatLM follows a cosine decay scheduling with weight decay and includes a learning rate warmup. This methodology is consistent with the practices established in models such as Pythia, OLMo, LLM360. For more details, refer to the Appendix (\ref{appendix:TriLMs_and_FloatLMs::subsection:Hyperparameters}).

        \textbf{Family of QuantLMs.}
        Recently, data-aware quantisation techniques like GPTQ \citep{gptq} have emerged as efficient solutions for near-lossless weight quantization down to 4-bit precision \citep{dettmers2023case_for_4_bit_precision}.  In our work, we implemented GPTQ post-training quantization to FloatLM, creating the QuantLM family of models across 3, 4, 6, and 8 bits. We quantized all transformer layer weights. For 3-bit and 4-bit quantization, we employ a group size of 128, which results in effective bit rates of 3.25 and 4.25 bits per parameter, respectively. We've refined our approach by incorporating best practices from recent research \citep{malinovskii2024pvtuningstraightthroughestimationextreme}, particularly in terms of calibration data and scaling it to a million tokens for improved reconstruction. To ensure a fair comparison with TriLM, we maintain certain components in their original precision. Specifically, we do not quantize the embedding, language model head, or activations. Additionally, we use symmetric quantization (without zero offset) as it is simpler, is supported by fast inference kernels \citep{frantar2024marlin} and offers similar performance to assymmetric quantization (with separate zero offsets in addition to scale for each group). It also offers consistency and a fairer comparison with TriLMs. It's worth noting that our Spectra suite is designed with flexibility in mind, allowing for easy extension to other quantization methods as needed.

    \subsection{Training Dynamics and Scaling Laws}
    
      \label{subsec:scaling_law}
    
        \begin{figure}[h]
            \centering
            \begin{subfigure}{0.4\textwidth}
                \centering
                \includegraphics[width=\linewidth, height=0.73\textwidth]{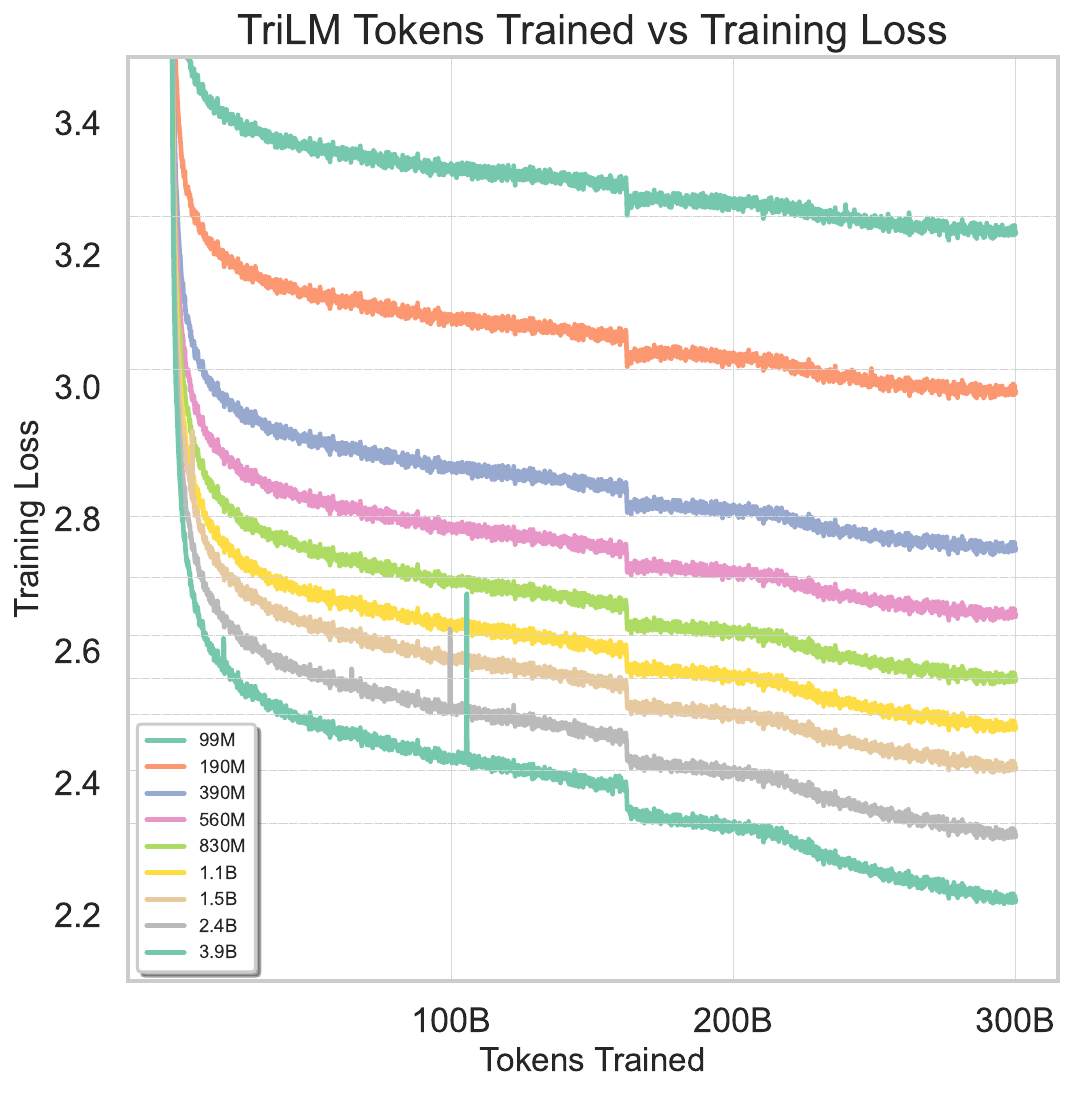}
                \caption{\footnotesize{Training loss over time for TriLMs.}}
                \label{fig:trilm_training_loss}
            \end{subfigure} 
            \hspace{0.03\textwidth}
            \begin{subfigure}{0.4\textwidth}
                \centering
                \includegraphics[width=\linewidth, height=0.73\textwidth]{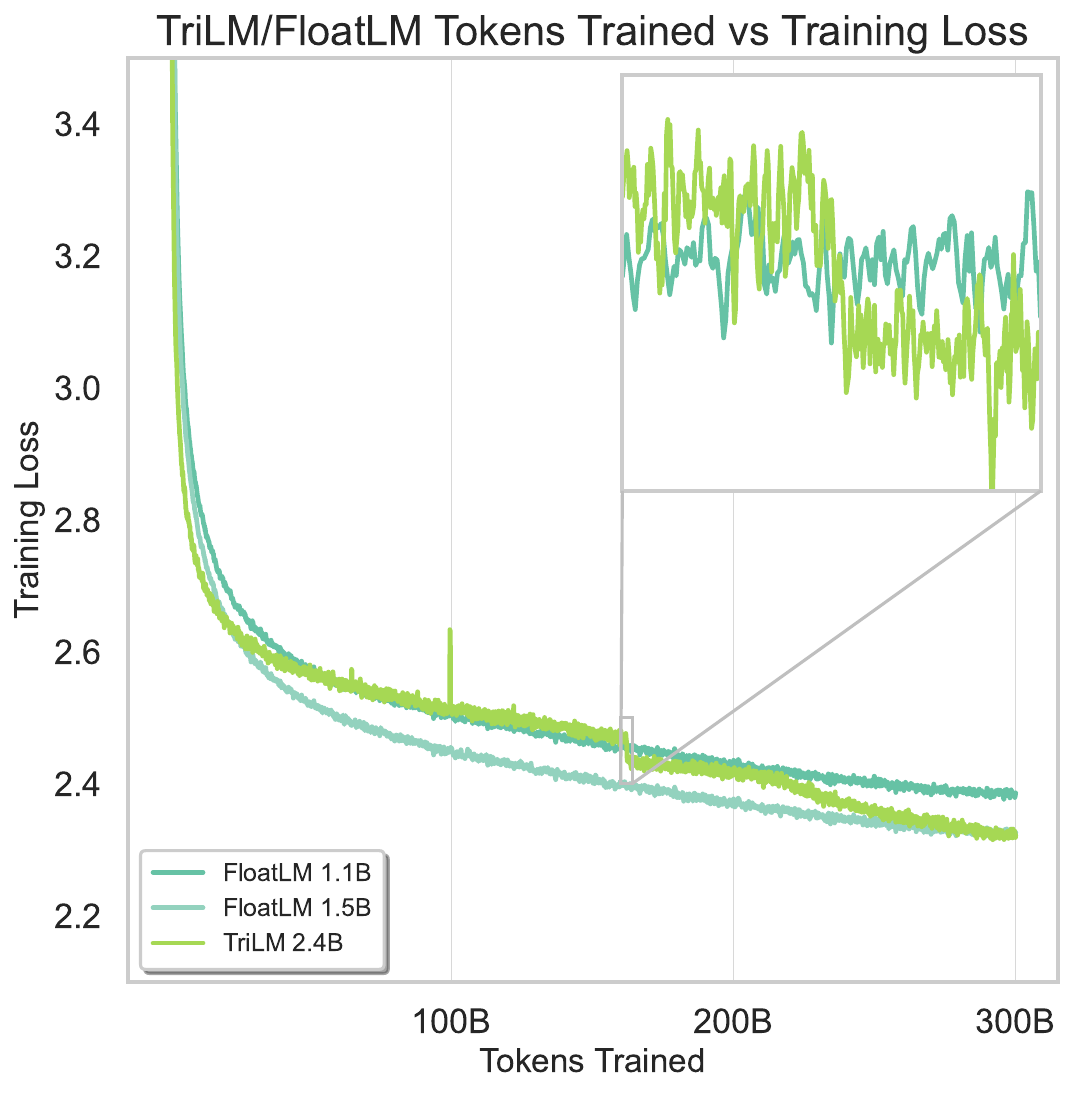}
                \caption{\footnotesize{Training loss of TriLM vs FloatLM.}}
    \label{fig:trilm_vs_floatlm_zoomed_comparison_training_loss}
            \end{subfigure}
            \caption{\footnotesize{Training Cross Entropy Loss across steps for the TriLM family of models. At the halfway point (150B tokens) when we lower the peak learning rate, we observe a sudden drop in training loss. In the two-third way, removing weight decay leads to faster convergence.}}
        \end{figure}
            
        \paragraph{Training Dynamics.} Figure \ref{fig:trilm_training_loss} illustrates the training loss curves for all the TriLMs trained and Figure \ref{fig:trilm_vs_floatlm_zoomed_comparison_training_loss} shows the relative training loss of a TriLM in comparison to two smaller FloatLMs. The loss curves demonstrate a continuous and consistent improvement in TriLMs with an increase in parameter count. During training, we make several key observations. First, minor spikes and drops in training loss occurred consistently across different TriLM scales at the same token counts, as all models were trained on identical data with the same ordering. Notably, the two largest models—TriLM 2.4B and TriLM 3.9B—each experienced a large loss spike in the first half of training. Second, adjusting the learning rate at the midpoint led to a sharp decline in training loss over a few hundred million tokens, though its impact varied by model size: for the larger models (2.4B and 3.9B), the rate of loss reduction returned to the prior pace after the initial sharp drop, while for smaller models (1.1B and 1.5B), the loss reduction plateaued, and models below 1B parameters saw an increase in training loss. Lastly, the removal of weight decay at the two-thirds mark accelerated convergence for all models, with the effect being most pronounced in the largest TriLM models. 
        % For more details and ablation of the optimization schedule, refer to Appendix \ref{section:TriLMs_and_FloatLMs::subsection:Optimization_Schedule}

      \begin{figure}[h]
            \centering
            \begin{subfigure}{0.48\textwidth}
                \centering
                \includegraphics[width=\linewidth, height=0.62\textwidth]{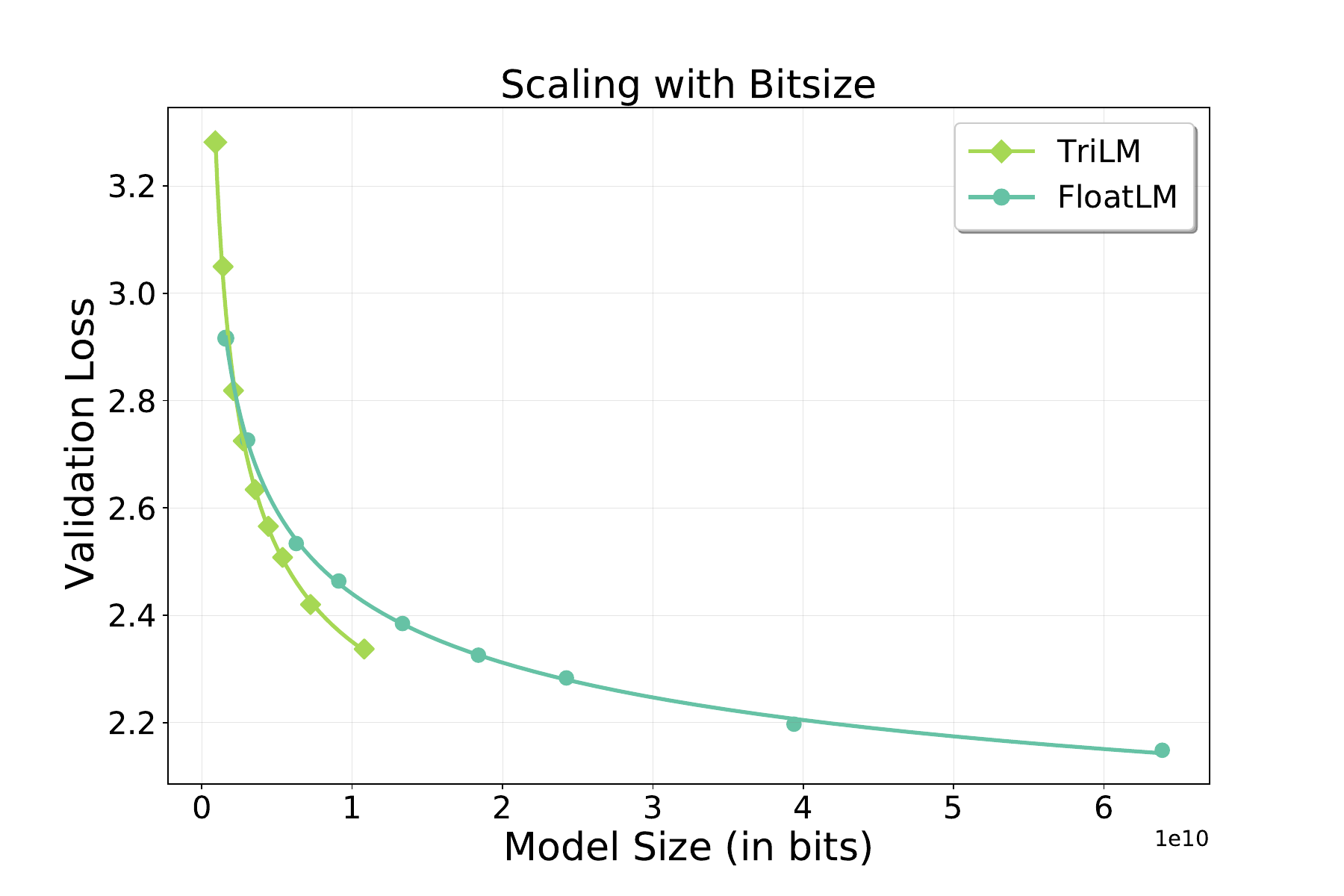}
                \caption{\footnotesize{Scaling laws - perplexity across size (bits).}}
                \label{fig:validation_perplexity_across_size}
            \end{subfigure}
             % \hspace{0.0\textwidth}
            \hspace{0.5mm}
            \begin{subfigure}{0.48\textwidth}
                \centering
                \includegraphics[width=\linewidth, height=0.62\textwidth]{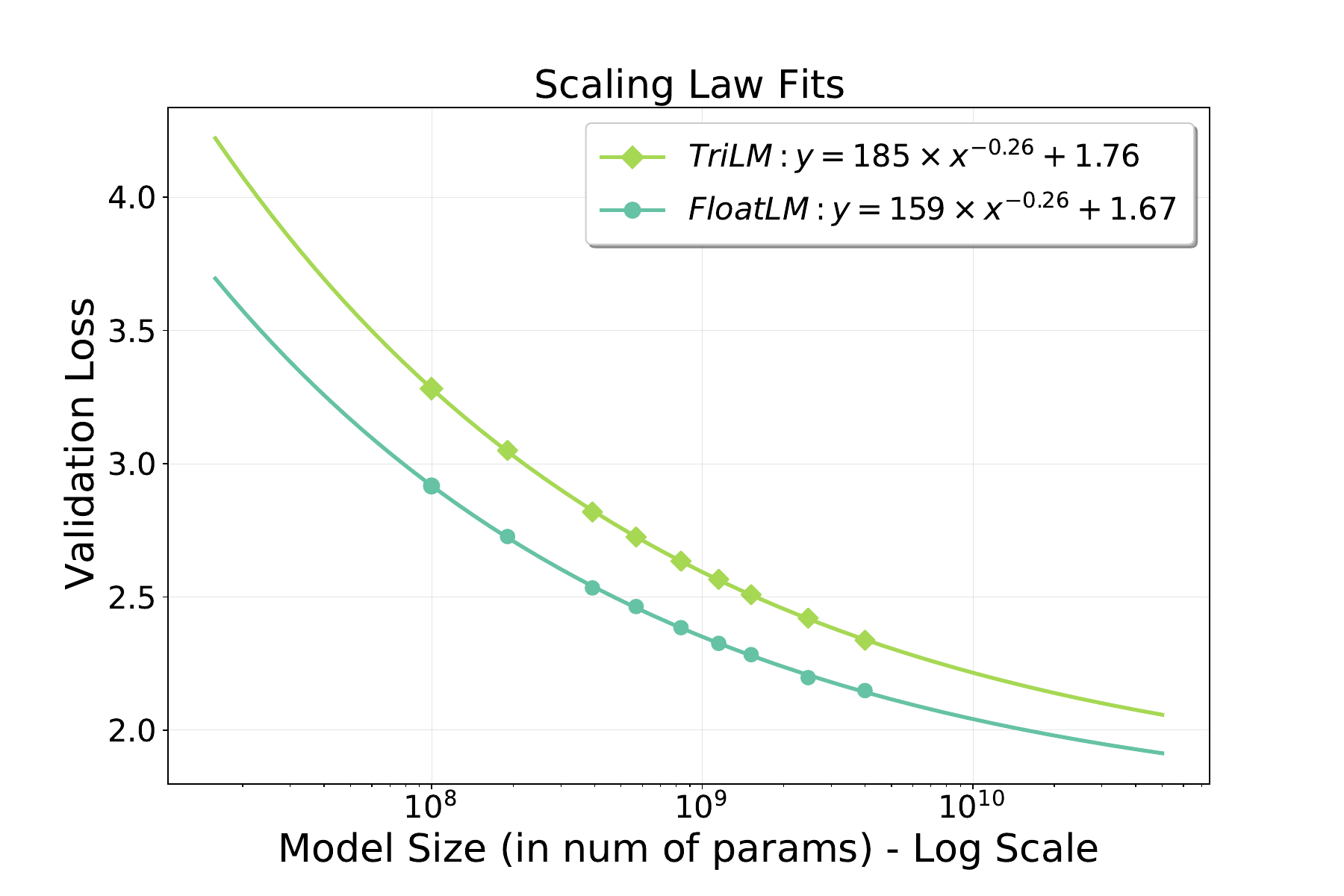}
                \caption{\footnotesize{Scaling laws - perplexity across parameters.}}
                \label{fig:validation_perplexity_vs_parameters}
            \end{subfigure}
            \caption{\footnotesize{Final validation loss across sizes (in bits) and parameters. TriLMs with increasing size offer better performance than FloatLMs of the same number of bits; and the gap in validation perplexity closes at scale.}}
        \end{figure}

        \paragraph{Scaling Laws.}Figures \ref{fig:validation_perplexity_across_size} and \ref{fig:validation_perplexity_vs_parameters} illustrate the final validation loss across different model size in terms of bits and number of parameters ($N$) respectively. In terms of effective model size in bits (\ref{fig:validation_perplexity_across_size}), which is crucial during inference, TriLMs exhibit significantly better scaling than FloatLMs. Notably, TriLM 3.9B validation error matches with FloatLMs 1.1B, which is nearly 1.7 times larger in terms of effective bit size.
To investigate the scaling behaviour in terms of $N$, we fit the validation loss to a power-law with offset\footnote{derived using a fixed data regime of 300B tokens} \cite{hoffmann2022trainingcomputeoptimallargelanguage} (see Figure \ref{fig:validation_perplexity_vs_parameters} and Appendix \S\ref{appdx:scaling_law}):
\begin{equation}
\mathcal{L}_{\text{type}}(N) = \frac{A_{\text{type}} }{ N^{ {\alpha}_{\text{type}}}} + \epsilon_{\text{type}}, \text{ where }
\begin{cases} 
A_{\text{TriLM}} = 185, &\alpha_{\text{TriLM}} = 0.26, \hspace{1.7em}\epsilon_{\text{TriLM}} = 1.76 \\
A_{\text{FloatLM}} = 159, &\alpha_{\text{FloatLM}} = 0.26, \hspace{1em}\epsilon_{\text{FloatLM}} = 1.67
\end{cases}
\end{equation}

We employ the Levenberg-Marquardt algorithm (\cite{Levenberg1944, Marquardt1963}) for efficient non-linear least squares fitting. Both FloatLM and TriLM share the scaling exponent \(\alpha = 0.26\), indicating similar scaling behavior with the number of parameters \(N\). However, the offset terms \(\epsilon_{\text{TriLM}} = 1.76\) and \(\epsilon_{\text{FloatLM}} = 1.67\) reveal that their validation losses converge as \(N\) increases. Although TriLM starts with a higher coefficient \(A = 185\), suggesting greater initial validation loss than FloatLM (\(A = 159\)), this difference becomes insignificant with larger $N$, aligning their performance at asymptotic scales as shown in Figure \S\ref{fig:validation_perplexity_vs_parameters}.

\begin{wrapfigure}{r}{0.46\linewidth}
    \centering
    \vspace{-8 mm}
    \includegraphics[width=\linewidth]{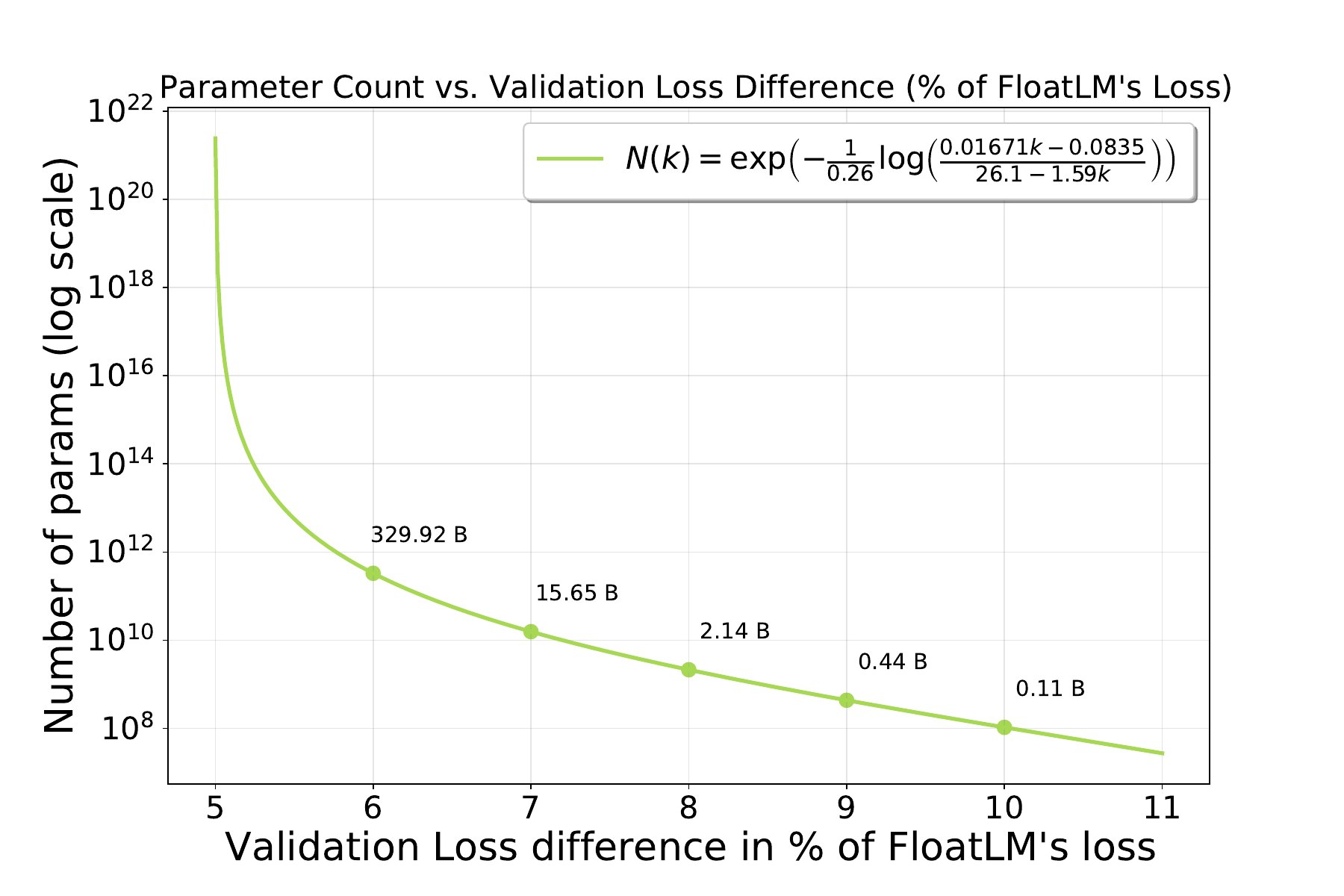}
    \caption{\footnotesize{Comparison of Power Law and Power Law-with-offset Fits for TriLM and FloatLM.}}
    \vspace{-4 mm}
    \label{fig:params_percent_validation}
\end{wrapfigure}
As shown in Figure \ref{fig:params_percent_validation}, using the scaling equations for TriLMs and FloatLMs, we derive the relationship between parameter count and the percentage difference in validation loss relative to FloatLMs. We observe that at 330B and 15.6B parameters, the validation losses for TriLMs are within $6\%$ and $7\%$ of FloatLMs' validation losses, respectively. This indicates that TriLMs are likely to closely match the performance of FloatLMs at larger scales. 

Despite the observed differences in validation loss at the scale of models considered in this work, we demonstrate in Section \ref{results} that at 3.9B parameters, TriLM offers competitive downstream performance compared to a FloatLM of the same parameter count across a variety of benchmarks in commonsense reasoning and knowledge-based tasks. Moreover, as discussed in Section \S\ref{appendix:Benchmark_Details::subsection:perplexity_on_other_datasets}, both models show similar perplexity on clean datasets such as Penn Tree Bank and OpenAI's Lambda. However, the gap in perplexity is observed in overlapping web-based datasets like Dolma and RefinedWeb.

\section{Results}
\label{results}
    We evaluate the families of LLMs on three aspects - commonsense \& reasoning tasks, knowledge-based tasks, and toxicity, all of which are crucial measures of their downstream performance. Readers may refer to the appendix for more details regarding the benchmarks Appendix (\S\ref{appendix:Benchmark_Details}).

    \paragraph{Commonsense and Reasoning.}
        We evaluate Spectra Suite models using eight distinct commonsense and reasoning benchmarks consisting of tasks from logical and reasoning questions to grounded and physical commonsense tasks: Arc Easy, Arc Challenge \citep{ARC}, BoolQ \citep{boolq}, HellaSWAG \citep{hellaswag}, WinoGrande \citep{winogrande}, PIQA \citep{piqa}, LAMBADA \citep{lambada}, LogiQA \citep{logiqa}, all under zero-shot settings. Figures \ref{fig:commonsense_and_reasoning_across_size} and \ref{fig:commonsense_and_reasoning_across_params} display the average performance of the LLMs on the first six benchmarks across size in bits and number of params. Figures \ref{fig:LAMBADA_acc_across_size} and \ref{fig:LAMBADA_acc_across_params} present the performance for the LAMBADA dataset. The results show that TriLMs consistently demonstrate superior performance for their size across all benchmarks at the $2.4$B and $3.9$B parameter scales. At the largest scale of $3.9$B, TriLM surpasses FloatLM on LAMBADA and achieves competitive average scores across six benchmarks. Additionally, TriLMs at the largest scales consistently outperform 4-bit QuantLMs of equivalent parameter count. However, across the considered scales, all LLMs show poor performance on LogiQA, making it difficult to identify a clear performance trend. For detailed benchmarking across all datasets \textendash see Tables \ref{tab:evaluation_spectra_suite_part1.a}, \ref{tab:evaluation_spectra_suite_part1.b} and \ref{tab:evaluation_spectra_suite_part2}.

    \paragraph{Knowledge.}\label{section:Evaluation::subsection:Commonsense_and_Reasoning}
        \begin{figure}[h]
            \centering
            \begin{subfigure}{0.25\textwidth}
                \centering
                \includegraphics[width=\linewidth]{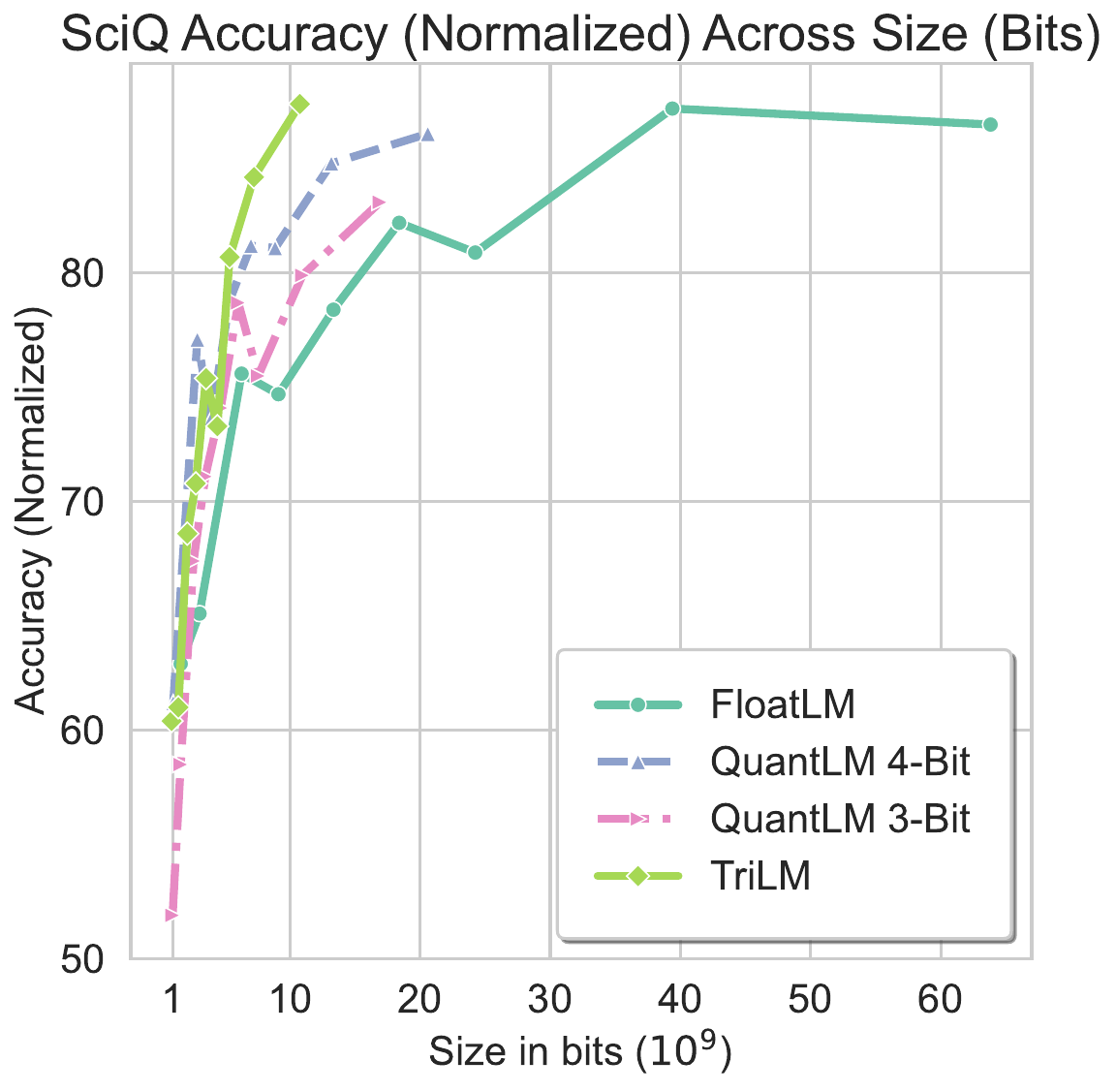}
                \caption{\tiny{vs. Size in SciQ}}
                \label{fig:SciQ_Acc_Norm_across_size}
            \end{subfigure}
            \hfill
            \begin{subfigure}{0.26\textwidth}
                \centering
                \includegraphics[width=\linewidth]{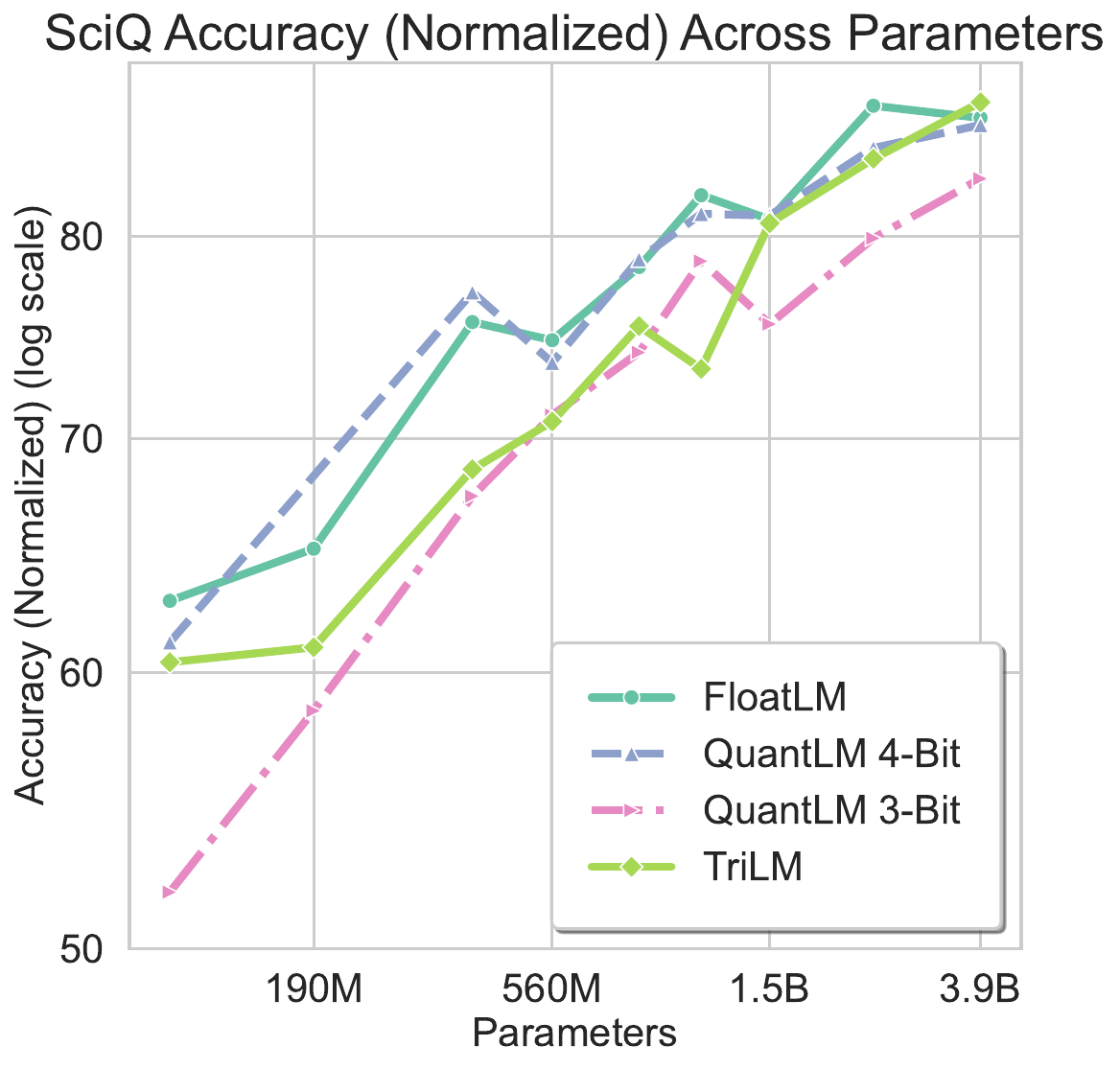}
                \caption{\tiny{vs. Params in SciQ}}
                \label{fig:SciQ_Acc_Norm_across_params}
            \end{subfigure}
            \hfill
            \begin{subfigure}{0.23\textwidth}
                \centering
                \includegraphics[width=\linewidth]{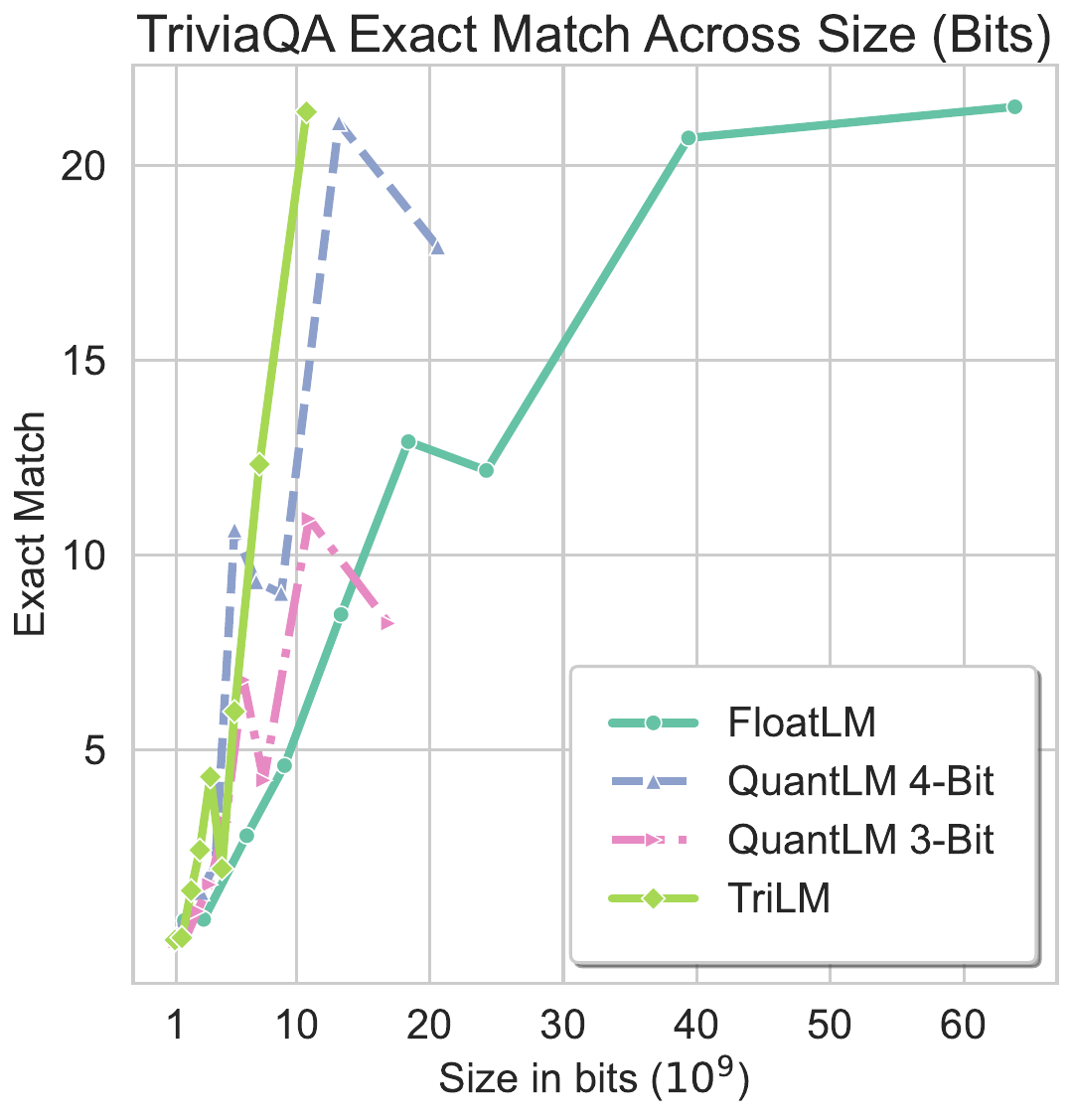}
                \caption{\tiny{vs. Size in TriviaQA}}
                \label{fig:TriviaQA_exact_match_across_size}
            \end{subfigure}
            \hfill
            \begin{subfigure}{0.24\textwidth}
                \centering
                \includegraphics[width=\linewidth]{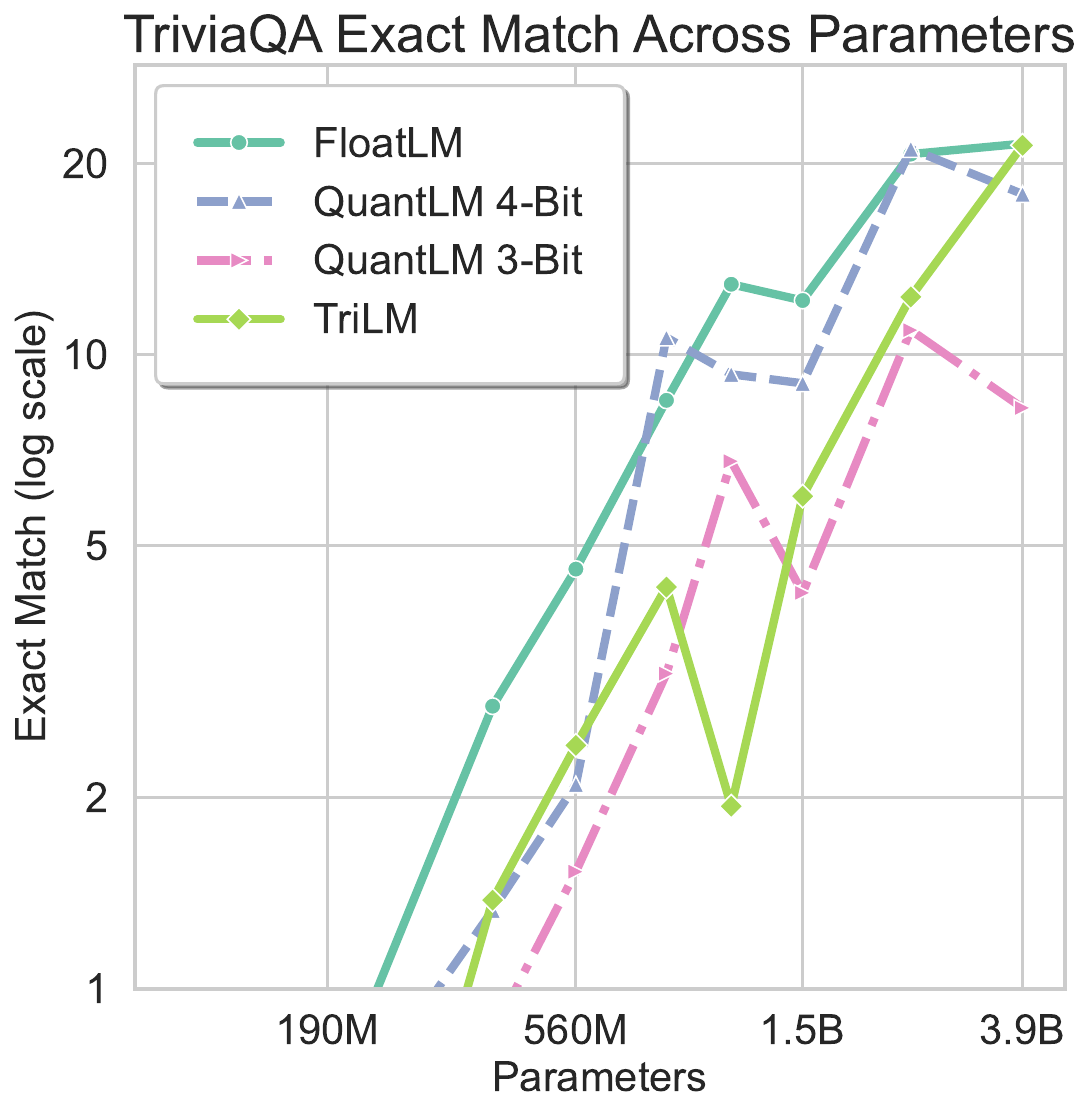}
                \caption{\tiny{vs. Params in TriviaQA}}
                \label{fig:TriviaQA_exact_match_across_params}
            \end{subfigure}
            \caption{\footnotesize{Performance of ternary TriLM, FloatLM and quantized QuantLM (3-bit \& 4-bit) models on SciQ and TriviaQA tasks across Size (Bits) and Parameters. Refer to Tables \ref{tab:evaluation_spectra_suite_part2} and \ref{tab:evaluation_spectra_suite_part3} for details.}}
            \label{fig:knowledge_performance}
        \end{figure}

\begin{figure}[h]
        \centering
        \begin{subfigure}{0.24\textwidth}
            \centering
            \includegraphics[width=\linewidth]{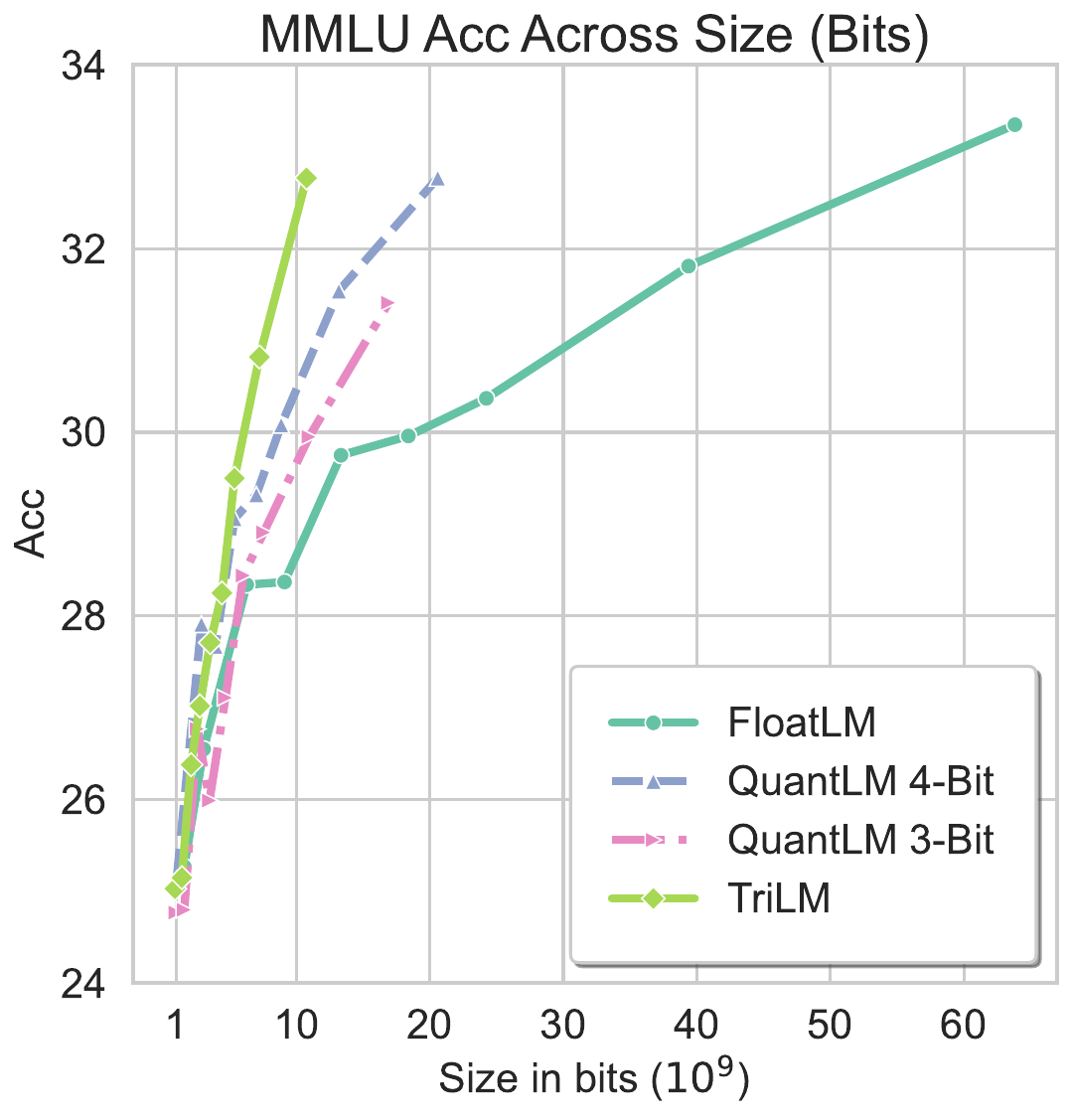}
            \caption{\tiny{Vs. Size (Bits)- Avg}}
            \label{fig:MMLU_across_size}
        \end{subfigure}
        \hfill
        \begin{subfigure}{0.24\textwidth}
            \centering
            \includegraphics[width=\linewidth]{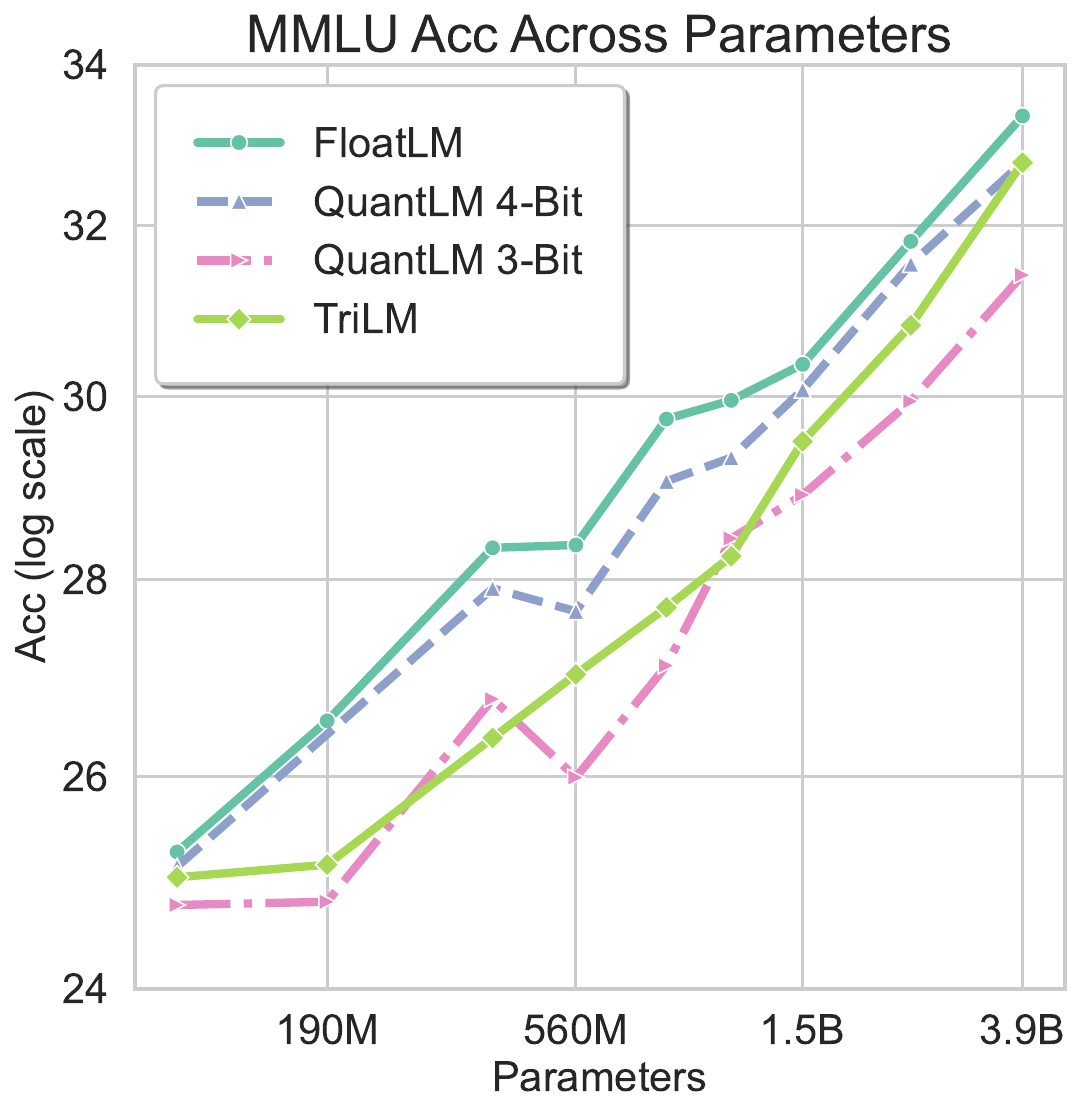}
            \caption{\tiny{Vs. Parameters- Avg}}
            \label{fig:MMLU_across_params}
        \end{subfigure}
        \hfill
        \begin{subfigure}{0.24\textwidth}
            \centering
            \includegraphics[width=\linewidth]{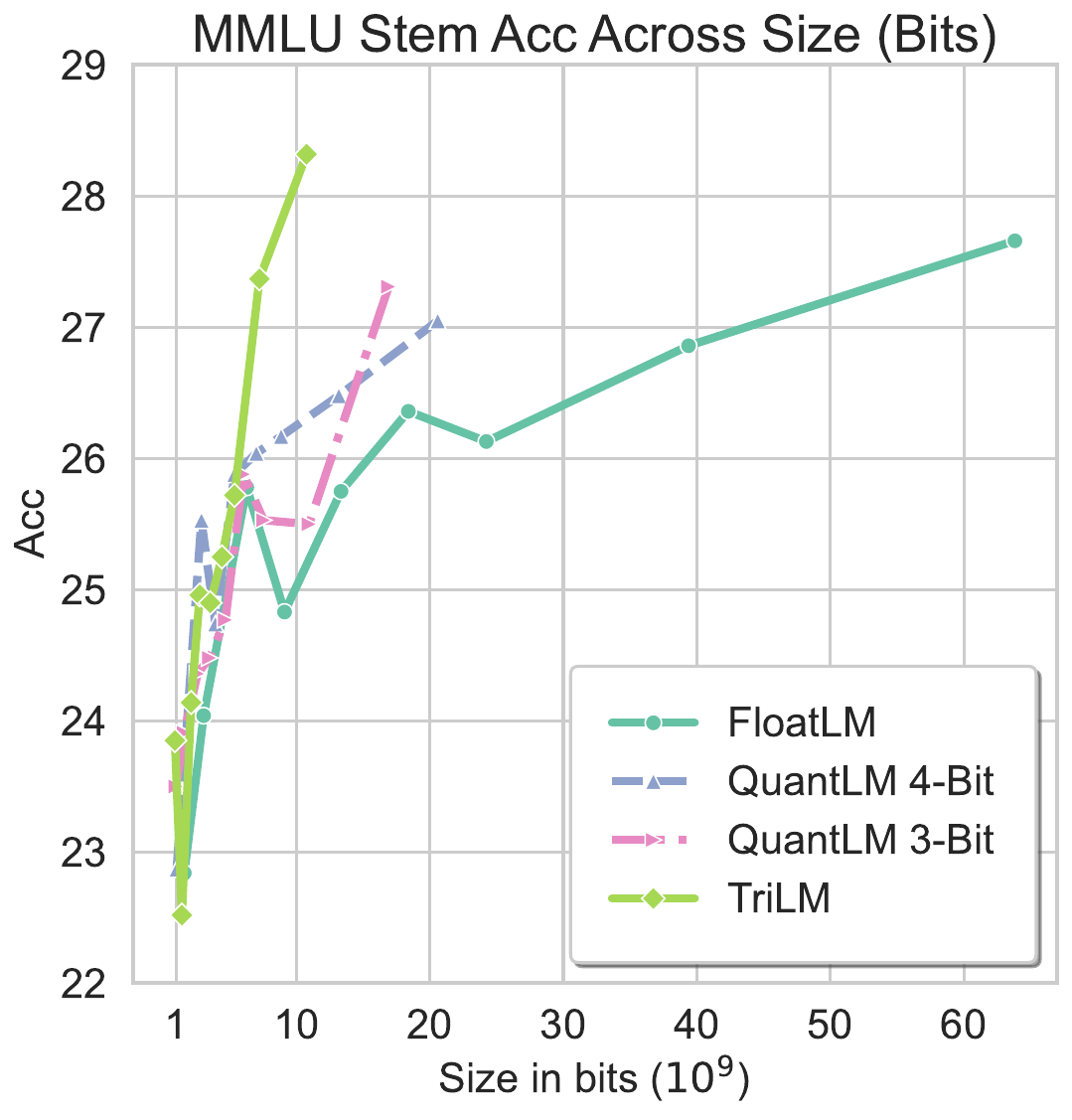}
            \caption{\tiny{Vs. Size- STEM}}
            \label{fig:MMLU_Stem_across_size}
        \end{subfigure}
        \hfill
        \begin{subfigure}{0.24\textwidth}
            \centering
            \includegraphics[width=\linewidth]{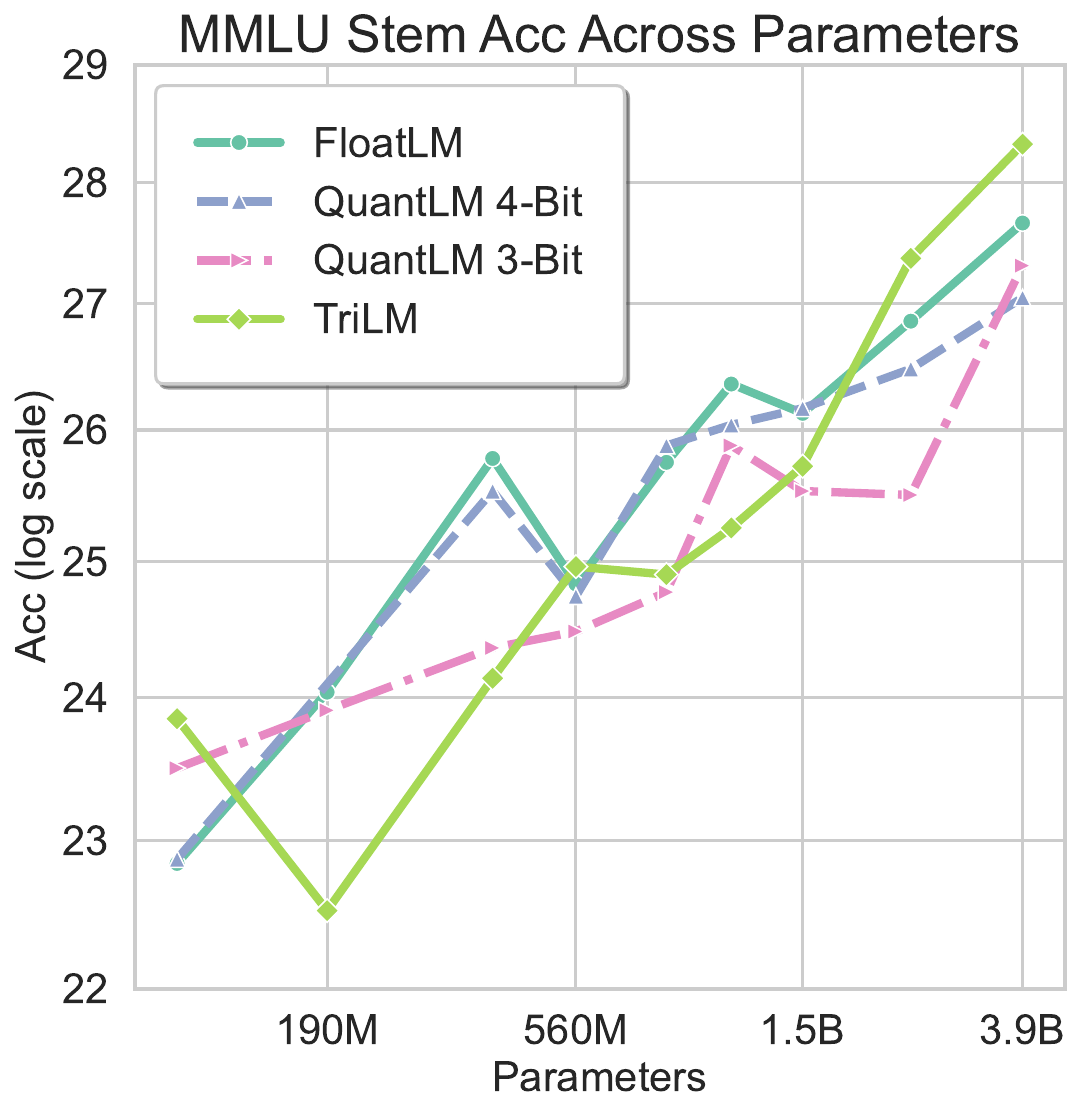}
            \caption{\tiny{Vs. Parameters- STEM}}
            \label{fig:MMLU_Stem_across_params}
        \end{subfigure}
        \caption{\footnotesize{MMLU Accuracy for ternary TriLM, FloatLM and quantized QuantLM (3-bit \& 4-bit) models across Size and Parameters. Please refer to Table \ref{tab:evaluation_spectra_mmlu} for details.}}
        \vspace{-0.4 cm}
        \label{fig:MMLU_comparisons}
    \end{figure}

        Several downstream practical uses of LLMs require them to have knowledge about common subjects like science or politics. To evaluate the performance of LLMs on these subjects, we use SciQ \citep{sciq}, TriviaQA \citep{triviaqa} and MMLU \citep{mmlu} benchmarks in zero-shot settings. Figures \ref{fig:SciQ_Acc_Norm_across_size} and \ref{fig:SciQ_Acc_Norm_across_params} show the accuracy of the Spectra suite LLMs on SciQ across size in bits and parameter counts. Figures \ref{fig:TriviaQA_exact_match_across_size} and \ref{fig:TriviaQA_exact_match_across_params} depict the accuracy for TriviaQA, while \ref{fig:MMLU_across_size} and \ref{fig:MMLU_across_params} do the same for MMLU. Across both benchmarks, at large 2.4B+ scales, TriLMs offer the best performance at a given size (bits). Surprisingly, despite having fewer bits, the knowledge capacity of TriLM does not have any significant degradation as observed in the case of QuantLMs \citep{allenzhu2024physicslanguagemodels33}. This indicates that low-bitwidth LLMs like TriLMs have similar knowledge capacity to FloatLMs, indicating that knowledge capacity is parameterized via the presence and nature of a connection (+1 or -1), rather than its strength. Tables \ref{tab:evaluation_spectra_suite_part2} and \ref{tab:evaluation_spectra_suite_part3} expand on these results.

    \paragraph{Toxicity.}
        We evaluate the Spectra suite across various safety and toxicity benchmarks of TruthfulQA \citep{truthfulqa}, Big Bench BBQ Lite \citep{bbq} and CrowsPairs \citep{crows_pairs}. These scores are listed in the Appendix in Table \ref{tab:evaluation_spectra_suite_part3}. We observe that none of the LLMs, even those with up to 3.9 billion parameters and trained on 300 billion tokens, perform significantly better than random guessing on the TruthfulQA benchmark. For the other two datasets, there is a noticeable correlation between the occurrence of toxicity and stereotypes and the LLMs' performance on various tasks. In particular, TriLMs with fewer than one billion parameters exhibit less stereotyping than FloatLMs with a similar parameter count. However, this difference diminishes as the scale increases, with TriLM 2.4B and TriLM 3.9B exhibiting biases comparable to those of FloatLM 2.4B and FloatLM 3.9B, respectively, on these benchmarks. This suggests that, although TriLMs initially show reduced bias compared to similarly sized FloatLMs, their performance aligns with FloatLMs of equivalent parameter counts at larger scales. This also highlights that TriLMs exhibit considerably more stereotyping than FloatLMs of comparable size (measured in bits), yet perform comparably to FloatLMs with similar parameter counts. 

    \paragraph{Perplexity on other datasets.}\label{appendix:Benchmark_Details::subsection:perplexity_on_other_datasets}

\begin{figure}[h]
    \centering
        \includegraphics[width=0.8\linewidth]{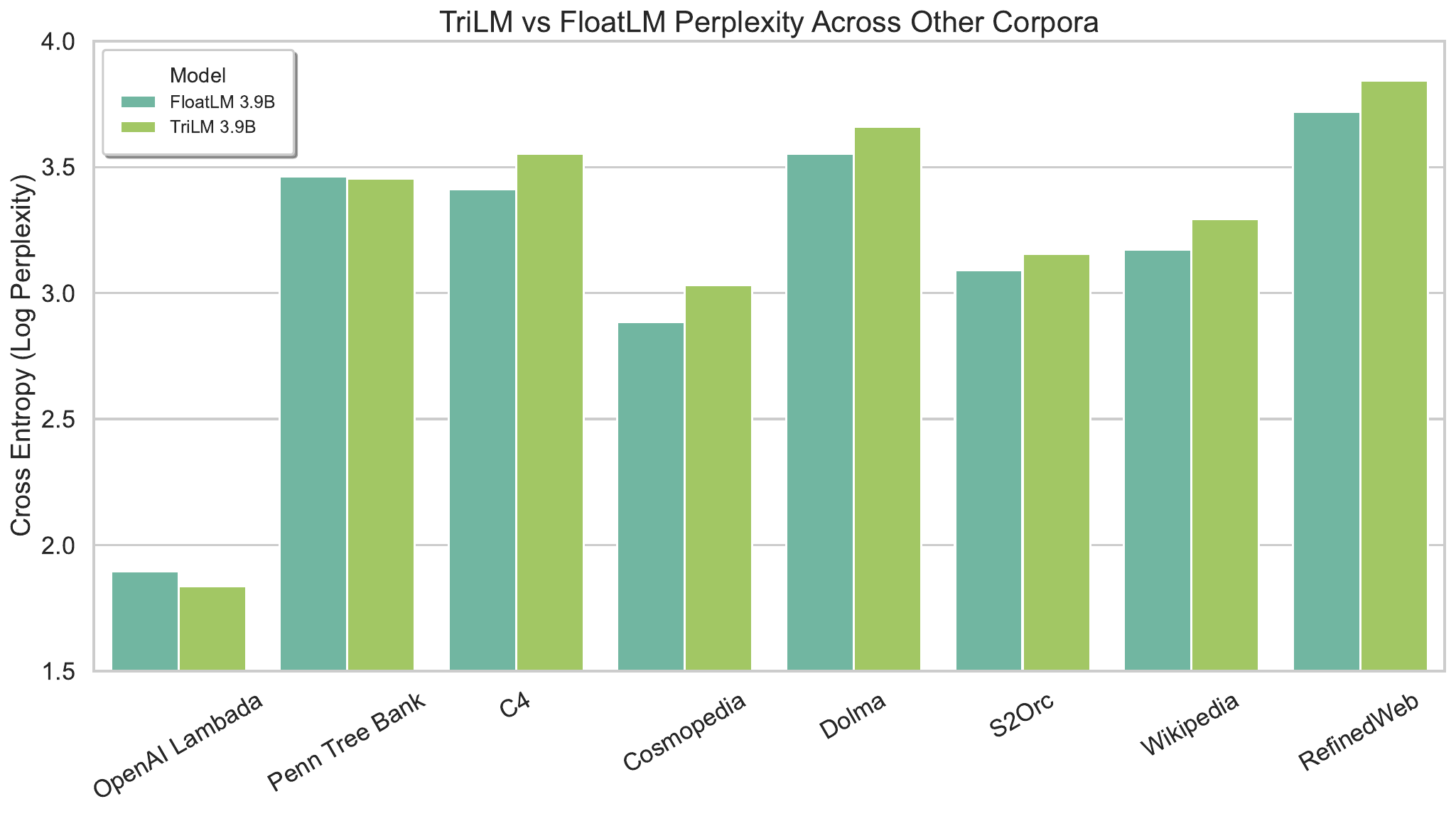}
        \caption{\footnotesize{Cross-entropy (log perplexity) comparison between TriLM and FloatLM (both 3.9B parameters) across various datasets apart from SlimPajama.}}
    \label{fig:perplexity_across_non_training_datasets}
\end{figure}

     We measure perplexity using TriLM 3.9B and FloatLM 3.9B across various other corpora than SlimPajama, which was used for training. These corpora include OpenAI Lambada, Penn Tree Bank, C4, Cosmopedia, Dolma, S2Orc, Wikipedia, and RefinedWeb. A portion of Wikipedia, C4 is included in Slim Pajama. Some other corpora like Dolma and RefinedWeb, may also have overlaps from C4, Wikipedia as well as Common Crawl. Figure \ref{fig:perplexity_across_non_training_datasets} demonstrates that while TriLM 3.9B is similar or better than FloatLM 3.9B on PTB and Lambada, across the other datasets, with potential overlaps with SlimPajama, it's performance is consistently worse - indicating lower capability to memorize training data as well as worse in-distribution performance, despite competitive out of distribution performance.
     
\paragraph{Illustrative examples.} We also generated a few examples of poem and essay writing, as well as reading comprehension (see Appendix \S\ref{app:readingcomprehension}). Our results show that the TriLM 3.9B model is able to generate cohesive and correct responses with randomly sampled examples.

\section{Related Work}

\paragraph{Quantization of Large Language Models after Training.}

 Post-training quantization (PTQ) algorithms convert a pretrained high-precision model (FP32 / FP16 / BF16) into a lower precision format without requiring the original training process \citep{cai2020zeroqnovelzeroshot, hubara2020improvingposttrainingneural, choukroun2019lowbitquantizationneuralnetworks}. These methods can be either data-independent or need a small calibration dataset. Additionally, PTQ for LLMs presents unique challenges due to numerical outliers in both weights and activations \citep{bondarenko2021understandingovercomingchallengesefficient}. GPTQ \citep{gptq} is a state-of-the-art one-shot weight quantization method aimed at finding a matrix of quantized weights (say $\hat{W}$) that minimizes the squared error relative to the full precision layer output. By leveraging second-order information, GPTQ derives a closed-form solution to this optimization problem, making it scalable to large LLMs. Other methods \citep{dettmers2023spqrsparsequantizedrepresentationnearlossless, lin2024awqactivationawareweightquantization, lee2024owqoutlierawareweightquantization} emphasize the importance of outlier weights that correspond to high-magnitude activations. Some recent methods also quantized activation along with the weights \citep{xiao2024smoothquantaccurateefficientposttraining,yao2022zeroquantefficientaffordableposttraining, yao2023zeroquantv2exploringposttrainingquantization}. \cite{ahmadian2023intriguingpropertiesquantizationscale} demonstrate that large activation outliers can be effectively mitigated at scale by making appropriate optimization decisions during the pretraining phase.
 
\paragraph{Training Language Models At Lower Precision.}
    Several prominent language models such as GPT \citep{brown2020languagemodelsfewshotlearners}, NeoX \citep{black2022gptneox20bopensourceautoregressivelanguage}, Llama and Pythia families have been traditionally trained using mixed precision (FP32/FP16 or FP32/BF16) \citep{micikevicius2018mixedprecisiontraining} or half-precision (FP16/BF16) \citep{kalamkar2019studybfloat16deeplearning}. Recently, \cite{tao-etal-2022-compression} introduced QuantGPT, a model that incorporates contrastive and logit distillation from a full-precision teacher to a quantized student model during pretraining. Further developments, such as BitNet \citep{bitnet_b1} and BitNet b1.58 \citep{bitnet_b1.58},  have specifically focused on quantization-aware training for extremely low-bitwidth networks in transformer-based models. In their work, models are trained at low ``effective'' precision of binary and ternary respectively - where the latent (or master) weights during training are maintained in higher precision like FP16. The model weights are binarized or ternarized on the fly during the forward pass and gradients are backpropagated for the latent weights using the straight-through estimator \citep{courbariaux2016binaryconnecttrainingdeepneural}. Prior works emphasize the importance of maintaining latent (or master) weights at high precision to allow accumulation of small updates during training - for example, \cite{peng2023fp8lmtrainingfp8large} observed a significant performance drop on the language model when the latent (or master) model weights were switch from 16-bits (FP16/BF16) to 8-bits (FP8) during training. Concurrent architectural improvements such as Flash Attention \citep{dao2022flashattentionfastmemoryefficientexact, dao2023flashattention2fasterattentionbetter}, the mixture of experts \citep{zoph2022stmoedesigningstabletransferable},  xLSTM \cite{beck2024xlstm} and state space models \citep{gu2024mambalineartimesequencemodeling, dao2024transformersssmsgeneralizedmodels, gu2022efficiently} complement these advancements in lower precision modeling.

\section{Conclusion and Future Work}

In this work, we address memory limitations in large language model (LLM) deployment by exploring both post-training quantization and direct low-bitwidth training. We introduce the Spectra LLM suite, featuring 54 models ranging from 99 million to 3.9 billion parameters, trained on 300 billion tokens. This suite includes Float16 LLMs (FloatLMs), quantized QuantLMs (3–8 bits), and our proposed ternary LLMs (TriLMs). Our findings reveal that TriLMs scale better than their half-precision Float16 counterparts in terms of effective model bit size, and they can achieve comparable validation loss when scaled to a large number of parameters. Additionally, our results demonstrate that TriLMs surpass other models in bit-size efficiency and achieve performance comparable to FloatLMs at 3 billion+ parameters across multiple benchmarks.

Future work should address the remaining challenges of toxicity, stereotyping, and performance gaps on web corpora associated with low-bitwidth models. Investigating scaling laws across different data regimes for various low-bitwidth architectures can provide deeper insights into their behavior and limitations. Additionally, combining these models with state-space architectures like Mamba \cite{gu2024mambalineartimesequencemodeling} could further enhance efficiency and performance without sacrificing accuracy. These research directions hold promise in advancing efficient language modeling further.

\section{Broader Impact}

    \paragraph{Interpretability Beyond Neuron Level:} While several efforts have been made to understand how language models work and means to steer them without training, these methods have mostly focussed on intervening at neuron level. TriLMs opens a new degree of interpretability - at the connection level. Here, the connections between any two neurons in a layer are in one of the three states - 0 (no connection), -1 (negative connection) and +1 (positive connection), each with equal strength. This is in sharp contrast to FloatLMs, where these connections can be of varying strengths, making it harder to study interpretability beyond neuron level. By releasing the checkpoints across our training runs, we facilitate research along these directions.

    \paragraph{Environmental Benefits and Resource Efficiency:} The open release of our models mitigates future emissions by allowing others to bypass the need for pretraining models from scratch. Moreover, TriLMs much lesser resource to deploy, and can perform the autoregressive generation as a faster pace - making them critical to scenarios demanding strict latency. Additionally, TriLMs represent a substantial advancement in enhancing performance on resource-constrained edge devices, including smartphones, laptops, and automobiles.

    \paragraph{Impact on Specialised Hardware:} While TriLMs offers significant memory reduction and latency improvements on General Purpose GPUs like H100 and RTX4090, certain specialized hardware benefits more from ternary modeling. Hardware (like Cerabras\footnote{https://www.cerebras.net/product-chip/}) that support high byte-to-flop ratio computations, can leverage the sparsity stemming from ternarization for speedup in both training as well as inference. On the other hand, hardware with limited Memory/SRAM (like Groq\footnote{https://groq.com/}), benefit from reduction in the number of chips needed to deploy an LLMs.

    \paragraph{Reduced Training Costs:} The Chinchilla scaling laws established that for training compute optimality, it may be recommended to train larger LLMs for lesser tokens than smaller LLMs for more tokens for achieving the desired model performance. However, memory requirements and latency associated with deployment of larger models, has motivated costlier training runs that go far beyond Chinchilla optimality. For example a LLaMa 3 model with only 8B parameter was trained for 15T tokens. Since, TriLM and ternary models in general can reduce the memory requirements and latency, this can motivate a shift inparameter-token tradeoff for efficient training runs towards Chinchilla's compute-optimal regime.

    % \subsection{Advancing Research through Open Access:}    

    %     The open suite of TriLM, FloatLM, and QuantLM families aims to empowers researchers to explore the nuanced impacts of precision levels on model performance and efficiency, thereby catalyzing ongoing advancements in the development and deployment of language models, as well as enhancing their interpretability and safety. By providing a range of publicly accessible models trained on openly available data, the suite offers unprecedented transparency in the training process. Intermediate checkpoints are available for all models, accompanied by detailed documentation of training procedures and hyperparameters. 
        
    %     % This comprehensive suite enables researchers to investigate the capacities and limitations of TriLMs at various scales, thus facilitating advancements in model development, and safety.

\section*{Acknowledgement}

    We acknowledge the support from the Mozilla Responsible AI Grant, the Canada CIFAR AI Chair Program and the Canada Excellence Research Chairs Program. This research was enabled by the computational resources provided by the Summit supercomputer, awarded through the Frontier DD allocation and INCITE 2023 program for the project "Scalable Foundation Models for Transferable Generalist AI" and SummitPlus allocation in 2024. These resources were supplied by the Oak Ridge Leadership Computing Facility at the Oak Ridge National Laboratory, with support from the Office of Science of the U.S. Department of Energy. We extend special thanks to Jens Glaser for his assistance with the Summit and Frontier supercomputers, and to Darshil Doshi for insightful discussions around scaling laws

\bibliography{neurips_2024}
\bibliographystyle{abbrvnat}

\newpage
\appendix

    \renewcommand{\arraystretch}{1.3}

\section{Architecture and PreTraining Details} \label{appendix:architecture_and_training_details}

        This section provides a comprehensive overview of the  pretraining for TriLM (Ternary Language Model) and FloatLM (Floating Point Language Model). We outline the forward and backward pass equations specific to their linear layers, highlighting the contrast between the FP16 matrices in FloatLM and the ternary matrices with scalar scaling in TriLM. Additionally, it covers dataset selection, tokenizer usage, and preprocessing methods employed for training data preparation. These discussions provide information on pretraining setups, implementation nuances, and key hyperparameters critical to the models' development.

        \subsection{Data and Tokenizer} \label{appendix:architecture_and_training_details::subsection:data_and_tokenizer}
            \paragraph{Dataset Selection:}

            Let input be $X \in R_{b \times n}$ for a linear layer with FP16 weight matrix $W \in R_{m \times n}$ and $Y \in R_{b \times m}$ be the output. The same matrix $W$ is also used to denote latent weights in TriLMs during training.

            For ternarized layers in TriLMs, we also have a scalar scale $\gamma \in R$, matrix with ternarized states $\widehat{W} \in \{-1, 0, 1\}_{n \times m}$ and ternarized matrix $\widetilde{W} \in R_{n \times m}$. We set $\epsilon=1e-5$.

            Due to lack of availability of Pile 300B \citep{pile_dataset_paper} used in Pythia, we opted to use a 300B token sample of deduplicated Slim Pajama dataset\footnote{We also make this subset public}. We sample from each subset with the probability proportional to its size.

            \textbf{Training Data Preparation:}
            \begin{itemize}
                \item \textbf{Main experiments (Spectra suite)}: We used the full 300B token sample.
                \item \textbf{Ablation studies}: Training runs with 100B tokens, we sample from these 300B tokens with equal probability weight to each data-point.
                \item \textbf{Fine-Web Edu experiments}: We tokenized one-third of a 350B token sample, from which we then sampled 100B tokens for our experiments.
            \end{itemize}

\begin{wraptable}{r}{0.4\textwidth}
                \centering
                \small
                \begin{tabular}{lr}
                    \toprule
                    \textbf{Dataset}       & \textbf{Size (Tokens)} \\
                    \midrule
                    Arxiv                  & 13B                 \\
                    Book                   & 13B                 \\
                    C4                     & 80B                 \\
                    Common Crawl           & 156B                \\
                    GitHub                 & 16B                 \\
                    Stack Exchange         & 10B                 \\
                    Wikipedia              & 12B                 \\
                    \midrule
                    \textbf{Total}         & 300B                \\
                    \bottomrule
                \end{tabular}
                \caption{\footnotesize{300B Subset of Slim Pajama}}
                \label{tab:dataset_info}
                \vspace{-1.4 cm}
            \end{wraptable}
            \paragraph{QuantLM:}
            For the creation of QuantLM, we utilized a subset of the Slimpajama-627B dataset, consisting of 512 samples with a sequence length of 2048. These samples were normalized for length. Our approach closely follows the methodology outlined in \citep{malinovskii2024pvtuningstraightthroughestimationextreme}.

            \paragraph{Tokenizer and Optimization Techniques:}
            We use the GPT-NeoX 20B tokenizer following Pythia. For speeding up training, we round embedding rounding of to the nearest multiple of 128 times the model parallel size.

        \subsection{PreTraining Setup} \label{appendix:architecture_and_training_details::subsection:PreTraining_Setup}

            We scale using 2D-parallelism with Megatron-style sharding \citep{megatron_paper} and use ZeRO stage 2 Deepspeed \citep{deepspeed_github_codebase} for ZeRO \citep{zero_optimizer_paper}. Our implementation was based on GPT NeoX Codebase \citep{gpt_neox_github_codebase}. We use AdamW \citep{kingma2017adam} for optimization. We train on nodes with IBM Power9 PC CPUs and 6x16GB V100. Due to the lack of BFloat16 support in V100, we train both TriLM and FloatLM in FP16 using Mixed Precision Training and Dynamic Loss Scaling. Please refer to \S\ref{section:TriLMs_and_FloatLMs::subsection:Implementation_Artifacts} for more implementation specific details. We extensively use Huggingface \citep{wolf-etal-2020-transformers} and Wandb \citep{wandb} for handling the checkpoints and experiment tracking.

        \subsection{Hyperparameters} \label{appendix:TriLMs_and_FloatLMs::subsection:Hyperparameters}

            Table \ref{tab:hyperparameters} shows the hyperparameters for TriLM and FloatLM's transformer architecture and their learning rate. We set Adam $\beta$ are set to (0.9, 0.95) for both families of models and all the reported runs are trained to 2048 sequence length. FloatLM and TriLM are respectively trained with batch sizes of $2M$ and $1M$ tokens respectively.

            \begin{table*}[h!]
                \centering
                \scriptsize
                \begin{tabular}{cccccccccc|}
                    \toprule
                    \textbf{Params} & \textbf{Hidden} & \textbf{GLU} & \textbf{Heads} & \textbf{Layers} & \textbf{MP} & \textbf{FloatLM LR} & \textbf{TriLM LR}                       \\ 
                    \midrule
                    99.74M (99M)    & 512             & 1280         & 8              & 16              & 1 & $4.0 * 10^{-4}$  & $2.4 * 10^{-3} \rightarrow 1.5 * 10^{-3} $ \\ 
                    190.0M (190M)   & 768             & 2048         & 12             & 16              & 1 & $4.0 * 10^{-4}$  & $2.4 * 10^{-3} \rightarrow 1.5 * 10^{-3} $ \\ 
                    392.4M (390M)   & 1024            & 2560         & 16             & 24              & 1 & $3.0 * 10^{-4}$  & $1.8 * 10^{-3} \rightarrow 1.2 * 10^{-3} $ \\ 
                    569.2M (560M)   & 1280            & 3072         & 20             & 24              & 1 & $2.8 * 10^{-4}$  & $1.6 * 10^{-3} \rightarrow 1.1 * 10^{-3} $ \\ 
                    834.0M (830M)   & 1536            & 4096         & 24             & 24              & 1 & $2.5 * 10^{-4}$  & $1.5 * 10^{-3} \rightarrow 1.0 * 10^{-3} $ \\ 
                    1.149B (1.1B)   & 1792            & 5120         & 28             & 24              & 2 & $2.2 * 10^{-4}$  & $1.3 * 10^{-3} \rightarrow 9.0 * 10^{-4} $ \\ 
                    1.515B (1.5B)   & 2048            & 6144         & 32             & 24              & 2 & $2.0 * 10^{-4}$  & $1.2 * 10^{-3} \rightarrow 8.0 * 10^{-4} $ \\ 
                    2.461B (2.4B)   & 2304            & 7680         & 36             & 30              & 3 & $2.0 * 10^{-4}$  & $1.2 * 10^{-3} \rightarrow 8.0 * 10^{-4} $ \\ 
                    3.989B (3.9B)   & 3072            & 9216         & 24             & 30              & 6 & $1.5 * 10^{-4}$  & $1.2 * 10^{-3} \rightarrow 8.0 * 10^{-4} $ \\ 
                    \bottomrule
                \end{tabular}
                \caption{Hyperparameters across model sizes for TriLM and FloatLM.}
                \label{tab:hyperparameters}
            \end{table*}

            \begin{table*}[h!]
                \centering
                \scriptsize
                \begin{tabular}{cccccccccc}
                    \toprule
                    \textbf{Params} & \textbf{99M} & \textbf{190M} & \textbf{390M} & \textbf{560M} & \textbf{830M} & \textbf{1.1B} & \textbf{1.5B} & \textbf{2.4B} & \textbf{3.9B}   \\ 
                    \midrule
                    FloatLM         & 1.60         & 3.05          & 6.28          & 9.11          & 13.34         & 18.39         & 24.23         & 39.38         & 63.83 \\
                    QuantLM 8-Bit   & 1.21         & 2.14          & 3.96          & 5.58          & 7.91          & 10.64         & 13.77         & 21.55         & 34.39 \\
                    QuantLM 6-Bit   & 1.11         & 1.92          & 3.38          & 4.70          & 6.55          & 8.70          & 11.15         & 17.09         & 27.03 \\
                    QuantLM 4-Bit   & 1.03         & 1.72          & 2.88          & 3.93          & 5.36          & 7.00          & 8.86          & 13.18         & 20.59 \\
                    QuantLM 3-Bit   & 0.98         & 1.60          & 2.59          & 3.49          & 4.68          & 6.03          & 7.55          & 10.95         & 16.91 \\
                    TriLM           & 0.90         & 1.42          & 2.11          & 2.76          & 3.55          & 4.42          & 5.36          & 7.23          & 10.76 \\
                    \bottomrule
                \end{tabular}
                \caption{Sizes in bits (*$10^9$) for Spectra suite of LLMs across varying parameter counts.}
                \label{tab:spectra_size_vs_parameters}
            \end{table*}

\subsection{Known Implementation Artifacts} \label{section:TriLMs_and_FloatLMs::subsection:Implementation_Artifacts}

\begin{table}[h]
\centering
    \begin{adjustbox}{width=0.7\textwidth}
    
    \begin{tabular}{cccc}
        \toprule
        Model & Min. Loss-Scale & \# Skipped Batches & \# Skipped Tokens \\
        \midrule
        FloatLM 99M  & 256.0  & 181 & 0.37B \\
        TriLM 99M    & 1024.0 & 303 & 0.33B \\
        \midrule
        FloatLM 190M & 512.0  & 168 & 0.35B \\
        TriLM 190M   & 512.0  & 305 & 0.33B \\
        \midrule
        FloatLM 390M & 1024.0 & 170 & 0.35B \\
        TriLM 390M   & 512.0  & 312 & 0.34B \\
        \midrule
        FloatLM 560M & 256.0  & 164 & 0.33B \\
        TriLM 560M   & 512.0  & 294 & 0.32B \\
        \midrule
        FloatLM 830M & 2048.0 & 175 & 0.36B \\
        TriLM 830M   & 128.0  & 307 & 0.33B \\
        \midrule
        FloatLM 1.1B & 2048.0 & 158 & 0.32B \\
        TriLM 1.1B   & 512.0  & 306 & 0.33B \\
        \midrule
        FloatLM 1.5B & 256.0  & 170 & 0.35B \\
        TriLM 1.5B   & 512.0  & 318 & 0.34B \\
        \midrule
        FloatLM 2.4B & 1024.0 & 165 & 0.34B \\
        TriLM 2.4B   & 256.0  & 294 & 0.32B \\
        \midrule
        FloatLM 3.9B & 256.0  & 164 & 0.34B \\
        TriLM 3.9B   & 128.0  & 309 & 0.33B \\
        \bottomrule
    \end{tabular}
    \vspace{0.4 cm}
    \end{adjustbox}        
    \caption{\footnotesize{Final loss-scale and number of batches skipped across TriLM and FloatLM training runs - We are able to maintain above the recommended loss scales of 128 for mixed precision training 
    \citep{micikevicius2018mixedprecisiontraining}.}}
    \label{table:dummy}
    
\end{table}

  % Similar to BitNet \citep{bitnet_b1}, our models have artifacts from model parallelism. Specifically, computing the scale $\gamma$ across the entire weight matrix, which has been sharded across multiple devices, incurs a significant communication overhead due to all-reduce operations. In our implementation, we compute these scales over the portion of weight matrix local to each device. Thus, for inference over TriLM models, scales should be independently computed over each model parallel group.  It is important to note that this change has a negligible effect on bits per parameter, amounting to less than $<10^{-5}$, even at the highest model parallelism level of $6$ for our largest model.

  %   Since we train in FP16, we expect some artifacts resulting from this training method. However, we do not expect significant performance differences when comparing mixed precision training with BF16 or even FP32. This is based on the observation that the lowest loss scales recorded during any of our runs were at least as high as the recommended value of 128 \citep{micikevicius2018mixedprecisiontraining} (see Table \ref{table:dummy}).

Similar to BitNet \citep{bitnet_b1}, our models exhibit artifacts resulting from model parallelism. A key issue arises when computing the scale, $\gamma$, across the entire weight matrix, which is sharded across multiple devices. This process introduces a significant communication overhead due to the all-reduce operations. In our implementation, we address this by computing the scales over the portion of the weight matrix local to each device. Consequently, during inference with TriLM models, scales are computed independently within each model parallel group. Importantly, this modification has a negligible impact on the bits per parameter, amounting to less than $10^{-5}$, even at the highest model parallelism level of $6$ for our largest model.

Given that we train in FP16, some artifacts are expected as a result of this training method. However, we do not anticipate significant performance differences when comparing mixed precision training with BF16 or even FP32. This expectation is based on the observation that the lowest loss scales recorded during our runs were consistently at or above the recommended value of 128 \citep{micikevicius2018mixedprecisiontraining} (refer to Table \ref{table:dummy}).

%Additionally, in BitNet b1.58 (Section 3), the authors compared their models to their reproduced FP16 LLaMA LLM, which suggests that our settings closely resemble theirs.

\subsection{Differences from BitNet Architecture}
\label{section:differences_from_bitNet_architecture}

    % \begin{figure}
    %         \centering
    %         \includegraphics[width=0.5\linewidth]{figures/commonsense_and_reasoning_architecture_relative_performance.pdf}
    %         \caption{Performance across various architectures - TriLM 1.1B, FloatLM 1.1B, BitNet b1.58 1.1B (our replication) along with reported scores of BitNet b1.58 at 700M and 1.3B params. Scores are averaged across 6 common sense and reasoning benchmarks, mentioned in Table \ref{tab:commonsense1_scores_evaluation_ablation}.}
    %         \vspace{-0.4 cm}
    %         \label{fig:commonsense_and_reasoning_architecture_relative_performance}
    % \end{figure}
            % \paragraph{Differences from BitNet Architecture.}

\begin{figure}[h]
    \centering
        \includegraphics[width=0.5\linewidth]{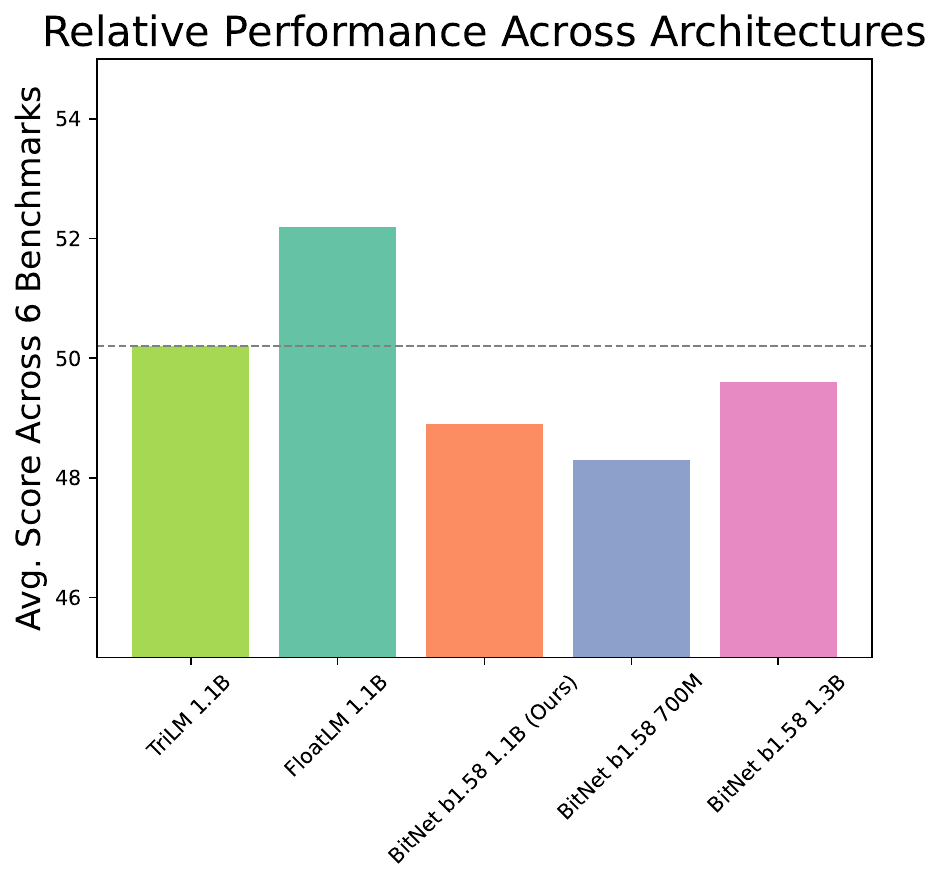}
        \caption{\footnotesize{Performance across various architectures - TriLM 1.1B, FloatLM 1.1B, BitNet b1.58 1.1B (our replication) along with reported scores of BitNet b1.58 at 700M and 1.3B params. Scores are averaged across 6 common sense and reasoning benchmarks, mentioned in Table \ref{tab:commonsense1_scores_evaluation_ablation_part1} and \ref{tab:commonsense1_scores_evaluation_ablation_part2}.} }
        \label{fig:commonsense_and_reasoning_architecture_relative_performance}
\end{figure}

        TriLM differs from BitNet b1.58 in several ways for better performance as well as for fairer comparison with FloatLMs. Adopting the GPT-3's Pre-Normalization approach as outlined by \citep{brown2020language}, normalization is applied prior to each linear layer. This method has proven essential for maintaining stable training under FP16 precision \citep{bitnet_b1}. Consequently, normalization occurs twice within each transformer layer: once at the input representations to the attention sub-layer and again at the input representations to the Gated MLP sub-layer. This approach contrasts with BitNet, where activation or intermediate representations are normalized, scaled, and quantized to 8 bits before each linear layer, which occurs between 4 to 7 times per transformer layer depending on the specific implementation. Furthermore, TriLM employs RMSNorm with a scale parameter over the parameterless RMSNorm.

        Figure \ref{fig:commonsense_and_reasoning_architecture_relative_performance} shows the commonsense and reasoning performance of TriLM 1.1B, FloatLM 1.1B and our replication of BitNet b1.58's architecture at 1.1B scale, along with the reported performance for BitNet b1.58 700M and 1.3B. All these models have been trained for 100B tokens. Our BitNet replication achieves performance between the 700M and 1.3B models. However, all the BitNet models, including the larger 1.3B parameter model perform worse than TriLM 1.1B. It should be noted that at this 1.1B scale, TriLMs do not achieve parity with FloatLMs of the same parameter count. Table \ref{tab:commonsense1_scores_evaluation_ablation_part1} and \ref{tab:commonsense1_scores_evaluation_ablation_part2}lists the detailed performance of these models across common sense benchmarks.

% \subsection{Spectra Suite of LLMs: Parameter Scaling and Bitwidth Variations}

% Figure \ref{fig:spectra_lm_suite_spans_across_parameters_and_bits} illustrates the Spectra LM suite spanning two key dimensions: bitwidth and the number of parameters. The suite includes the TriLM, FloatLM, and QuantLM model families (in 3, 4, 6, and 8 bits). The suite features 9 different parameter scales ranging from 99M to 3.9B parameters. In total, the suite comprises 54 models. 

% % \begin{figure}{h}
% %     \centering
% %         \includegraphics[width=0.5\linewidth]{figures/spectra_lm_suite_spans_across_parameters_and_bits.pdf}
% %         \caption{\footnotesize{The Spectra Suite spans across two dimensions: parameters and scale. Each point corresponds to a language model in the suite.}} 
% %         \label{fig:spectra_lm_suite_spans_across_parameters_and_bits}
% % \end{figure}

% \begin{figure}[h]
%     \centering
%     \includegraphics[width=0.50\linewidth]{figures/spectra_lm_suite_spans_across_parameters_and_bits.pdf}
%     \caption{\footnotesize The Spectra Suite spans across two dimensions: parameters and bitwidth. Each point corresponds to a language model in the suite.}
%     \label{fig:spectra_lm_suite_spans_across_parameters_and_bits}
% \end{figure}

\clearpage

\section{Scaling of Binary and Ternary Large Language Models}
\label{appendix:binary_vs_ternary_LLMs}

In this section, we will comprehensively compare Binary Large Language Models (BiLMs) with Ternary Large Language Models (TriLMs). We will start by describing BiLMs, followed by studying scaling laws and presenting results on various benchmarks, as well as comparisons with TriLMs across parameter count and model size (in bits).

\subsection{Scaling Laws}
\label{appendix:Scaling_Laws}
Figures \ref{fig:validation_perplexity_across_size_in_bilms} and \ref{fig:validation_perplexity_vs_parameters_in_bilms} show the final validation loss across sizes (in bits) and parameters, respectively. At the Billion+ model scale, Ternary Models appear to be preferable in terms of both the number of parameters and model size (in bits). However, the gap seems to be declining. The convergence point appears to occur at a high parameter scale (10B+). Thus, we decided to scale only TriLMs further to study the scaling laws of FLoatLMs and TriLMs upto a scale of 3.9 B parameters. This trend suggests that BiLMs have the potential to match at higher parameter counts.

\subsection{BiLM: Binary Large Language Model}

In Binary Large Language Models (BiLMs), the weights of the linear layers are represented by binary values of -1 or 1, with an accompanying floating-point scaling factor, similar to the method employed in TriLMs. Comprehensive formal descriptions of the forward pass, backward pass, and inference time calculations are provided in Appendix (\S\ref{appendix:architecture_and_training_details}). We have trained three BiLM models of distinct sizes: 99M, 560M, and 1.1B parameters. These models were trained on the same dataset and in the same sequence as the TriLMs, adhering to the identical optimization schedule detailed in Section \ref{section:TriLMs_and_FloatLMs::subsection:Optimization_Schedule}.

  \begin{figure}[h]
    \centering
    \begin{subfigure}{0.47\textwidth}
        \centering
        \includegraphics[width=\linewidth, height=0.7\textwidth]{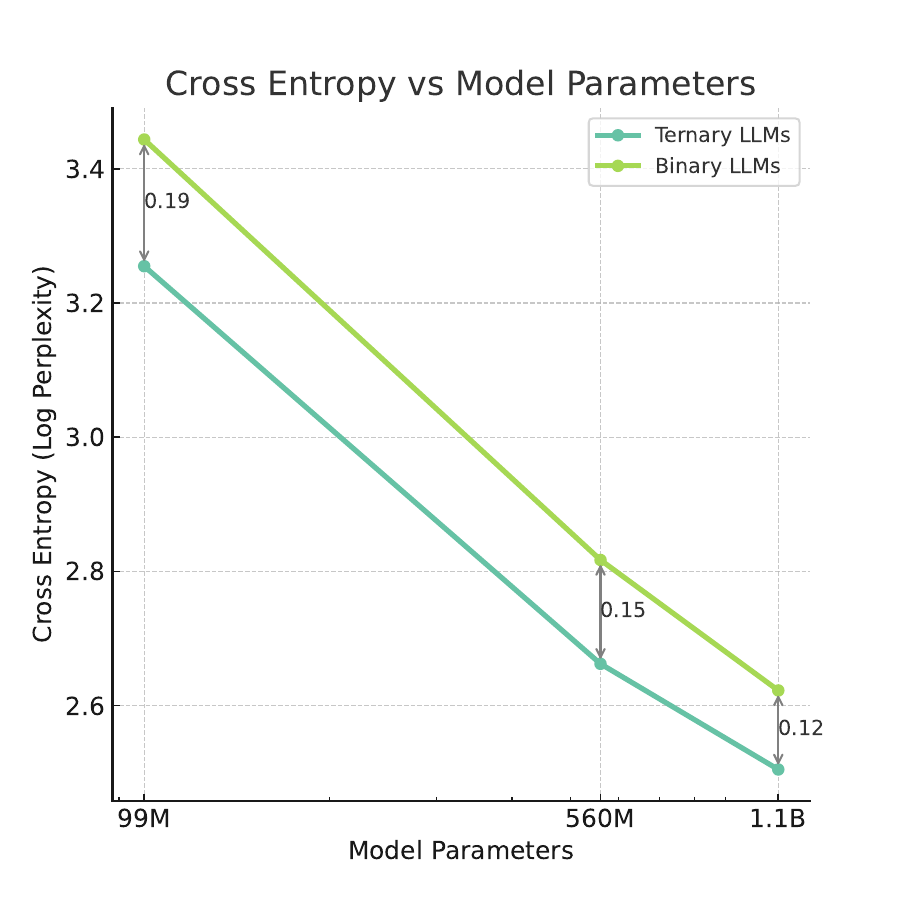}
        \caption{Vs. Params}  \label{fig:validation_perplexity_across_size_in_bilms}
    \end{subfigure}
    \hspace{5mm}
    \begin{subfigure}{0.47\textwidth}
        \centering
        \includegraphics[width=\linewidth, height=0.7\textwidth]{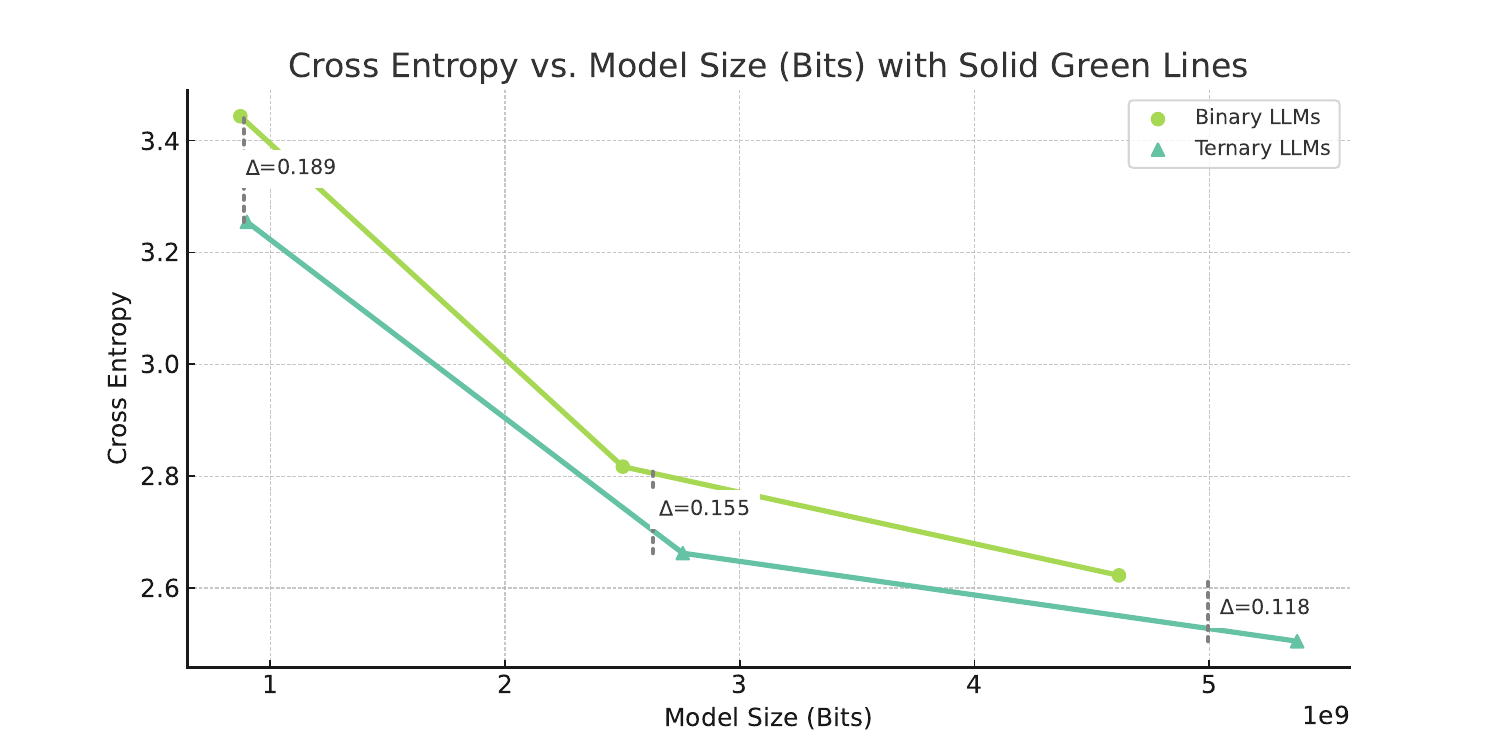}
        \caption{Vs. Size (in Bits)}
        \label{fig:validation_perplexity_vs_parameters_in_bilms}
    \end{subfigure}
    \caption{\footnotesize{Final Validation loss across size (measure in bits) and parameters.}}
    \label{fig:scaling_law_curves_binary_ternary}

    \vspace{-0.4 cm}
\end{figure}

Figures \ref{fig:validation_perplexity_across_size_in_bilms} and \ref{fig:validation_perplexity_vs_parameters_in_bilms} show the final validation loss across model sizes (in bits) and parameter counts, respectively. At the Billion+ model scale, Ternary Models appear to be preferable in terms of both the number of parameters and model size (in bits). However, the gap seems to be narrowing, with convergence likely occurring at higher parameter scales (10B+). Therefore, we decided to scale only TriLMs further to study the scaling laws of FloatLMs and TriLMs up to 3.9B parameters. This trend suggests that BiLMs have the potential to match the performance of floatLM at higher parameter counts.

\subsection{Results}
We conducted a comprehensive benchmark analysis of Binary Large Language Models (BiLMs) across three key dimensions: commonsense and reasoning tasks, knowledge-based tasks, and toxicity evaluation, as detailed in Tables \ref{tab:evaluation_spectra_suite_part1.a}, \ref{tab:evaluation_spectra_suite_part1.b}, \ref{tab:evaluation_spectra_suite_part2}, and \ref{tab:evaluation_spectra_suite_part3}

\label{section:Evaluation::subsection:Commonsense_and_Reasoning_BiLMs}
        \begin{figure}[h]
            \centering
            \begin{subfigure}{0.38\textwidth}
                \centering
                \includegraphics[width=\linewidth]{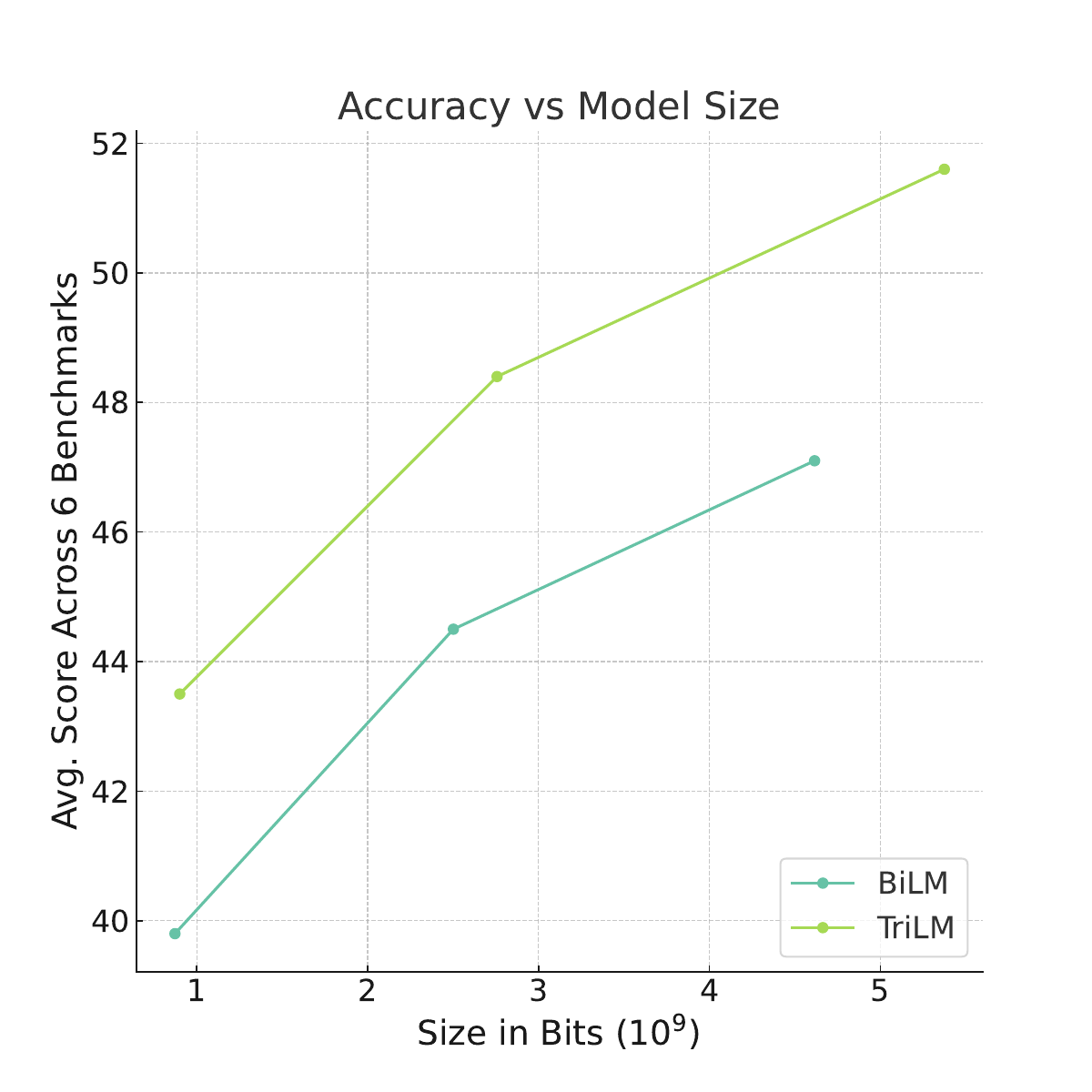}
                \caption{\tiny{vs. Size in C \& R}}
                \label{fig:CnR_Acc_across_size_BiLMs}
            \end{subfigure}
            \begin{subfigure}{0.38\textwidth}
                \centering
                \includegraphics[width=\linewidth]{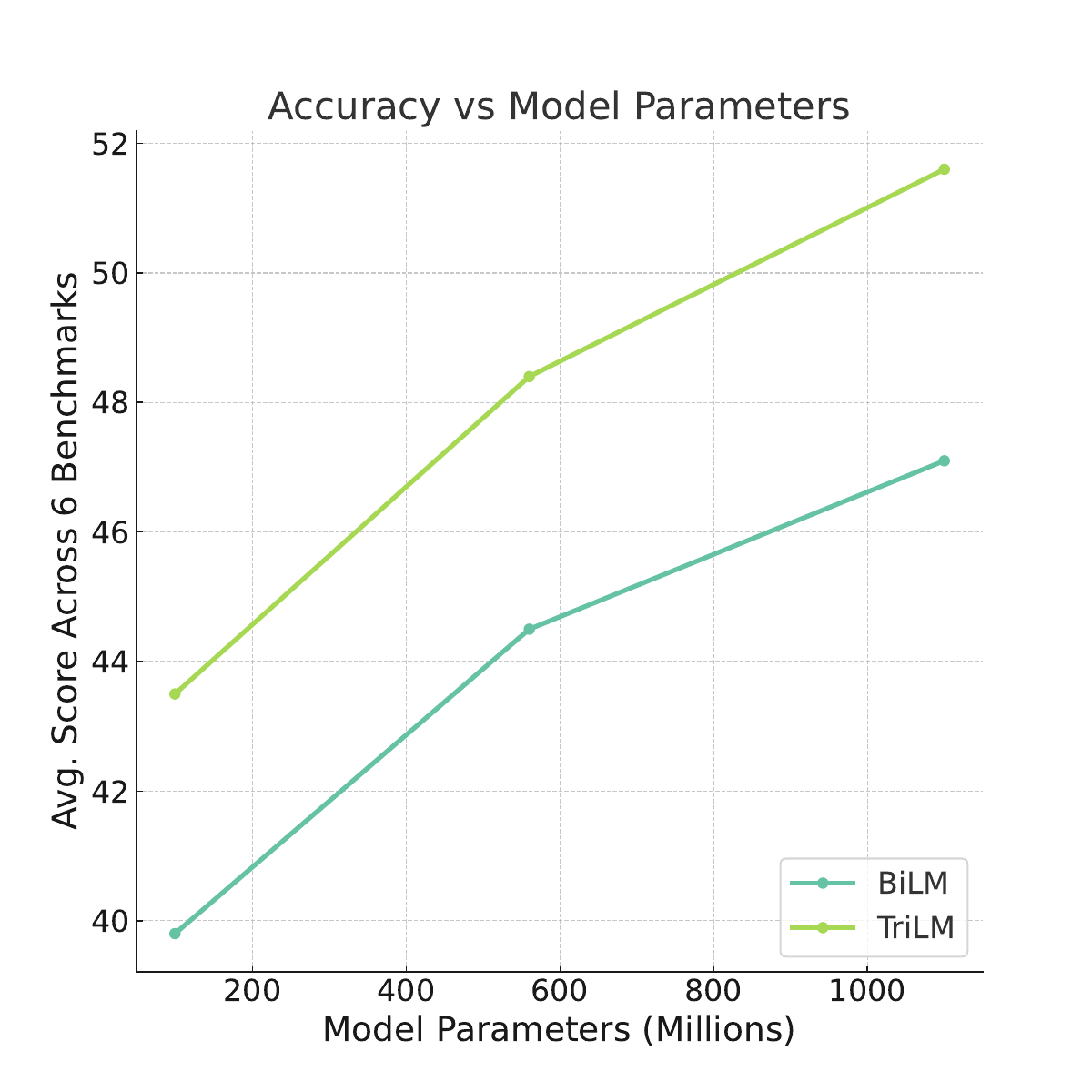}
                \caption{\tiny{vs. Params in C \& R}}
                \label{fig:CnR_Acc_across_params_BiLM}
            \end{subfigure}
            \caption{\footnotesize{Performance of ternary TriLMs and BiLMs models on commonsense and Reasoning and MMLUs tasks across Size (Bits) and Parameters. Refer to Tables \ref{tab:evaluation_spectra_suite_part1.a} and \ref{tab:evaluation_spectra_suite_part1.b} for details.}}
            \label{fig:knowledge_performance}
        \end{figure}
\label{section:Evaluation::subsection:lambada_BiLMs}

        \begin{figure}[h]
            \centering
            \begin{subfigure}{0.4\textwidth}
                \centering
                \includegraphics[width=\linewidth]{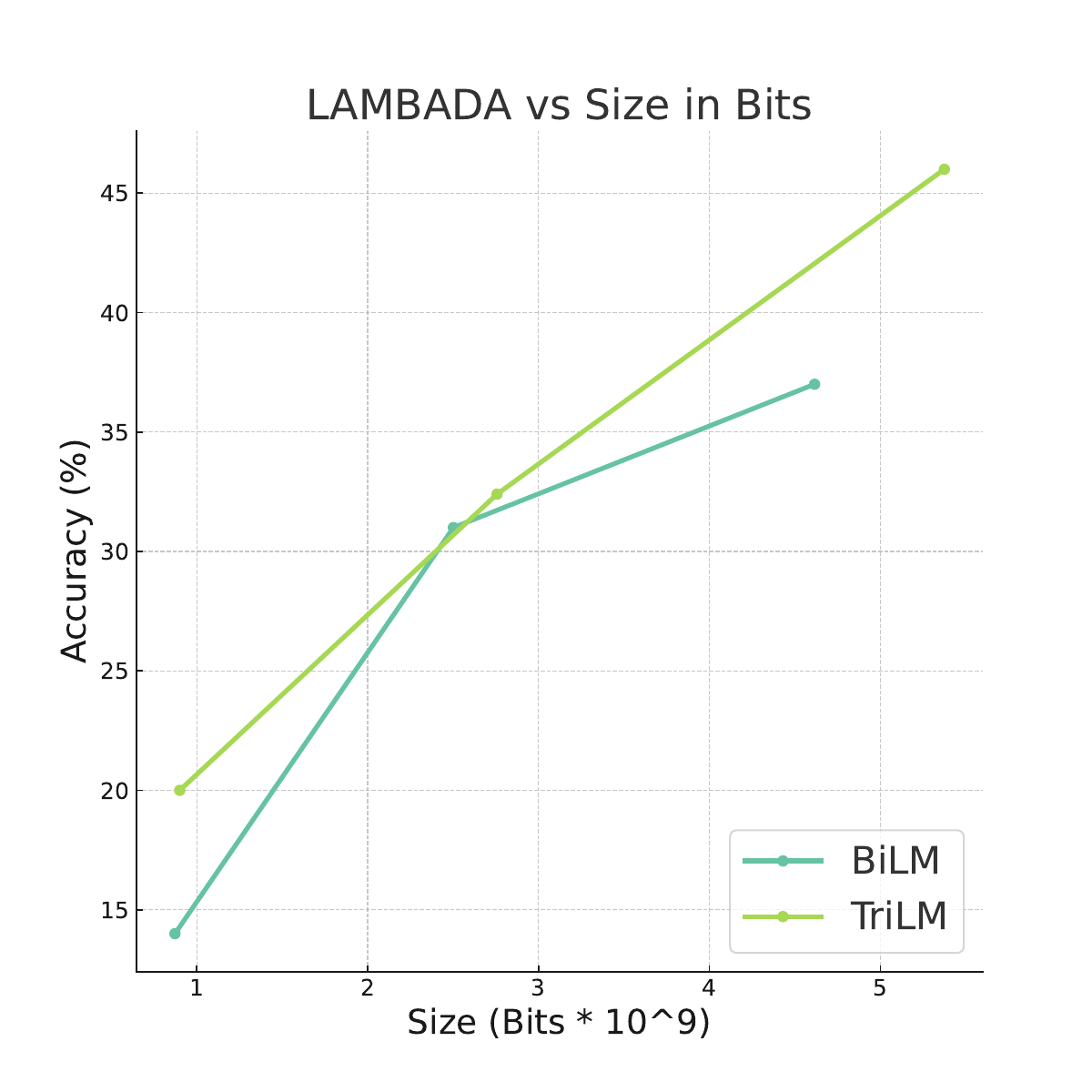}
                \caption{\tiny{vs. Size in LAMBADA}}
\label{fig:TriLM_vs_BiLMs_Lambda_exact_match_across_size}
            \end{subfigure}
            \begin{subfigure}{0.4\textwidth}
                \centering
                \includegraphics[width=\linewidth]{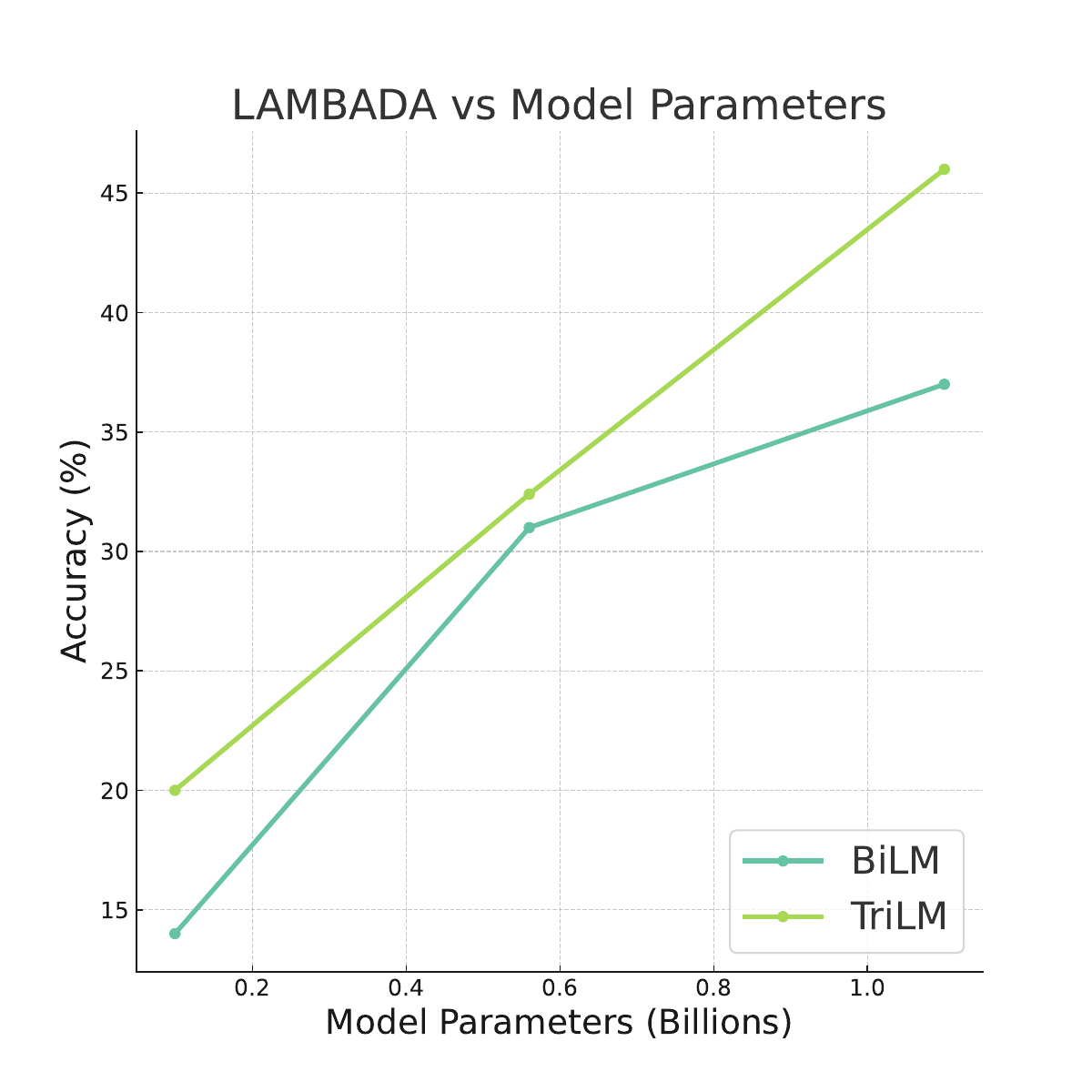}
                \caption{\tiny{vs. Params in LAMBADA}}
                \label{fig:TriLM_vs_BiLMs_Lambda_exact_match_across_params}
            \end{subfigure}
            \caption{\footnotesize{Performance of ternary TriLMs and BiLMs models on LAMBADA tasks across Size (Bits) and Parameters. Refer to Tables \ref{tab:evaluation_spectra_suite_part1.a}, and \ref{tab:evaluation_spectra_suite_part1.b} for details.}}
            \label{fig:BiLM_knowledge_performance}
        \end{figure}

\label{section:Evaluation::subsection:lambada_BiLMs}
        \begin{figure}[h]
            \centering
            \begin{subfigure}{0.4\textwidth}
                \centering
                \includegraphics[width=0.9\linewidth]{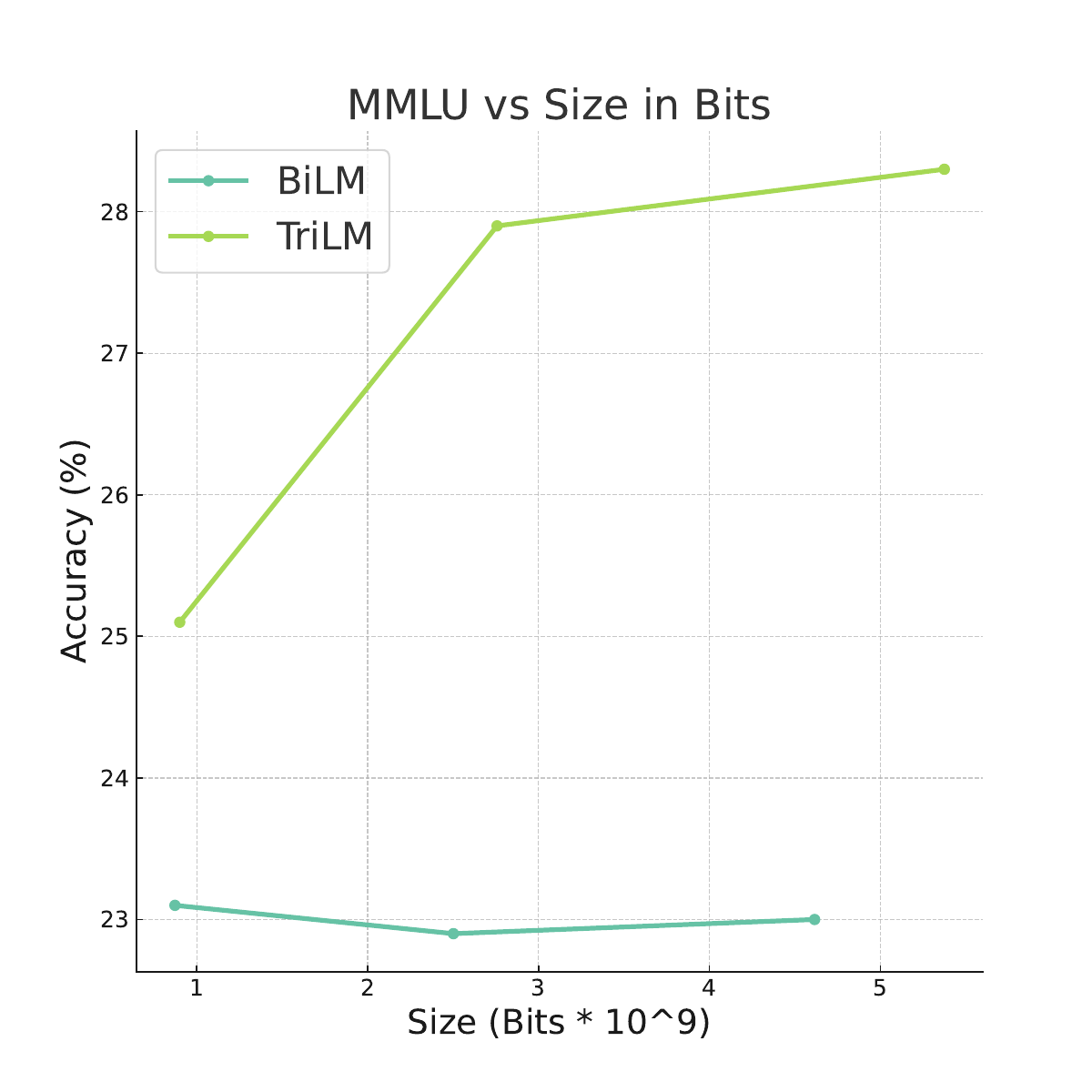}
                \caption{\tiny{vs. Size in MMLU}}
                \label{fig:TriLM_vs_BiLMs_MMLU_across_size}
            \end{subfigure}
            \begin{subfigure}{0.4\textwidth}
                \centering
                \includegraphics[width=0.9\linewidth]{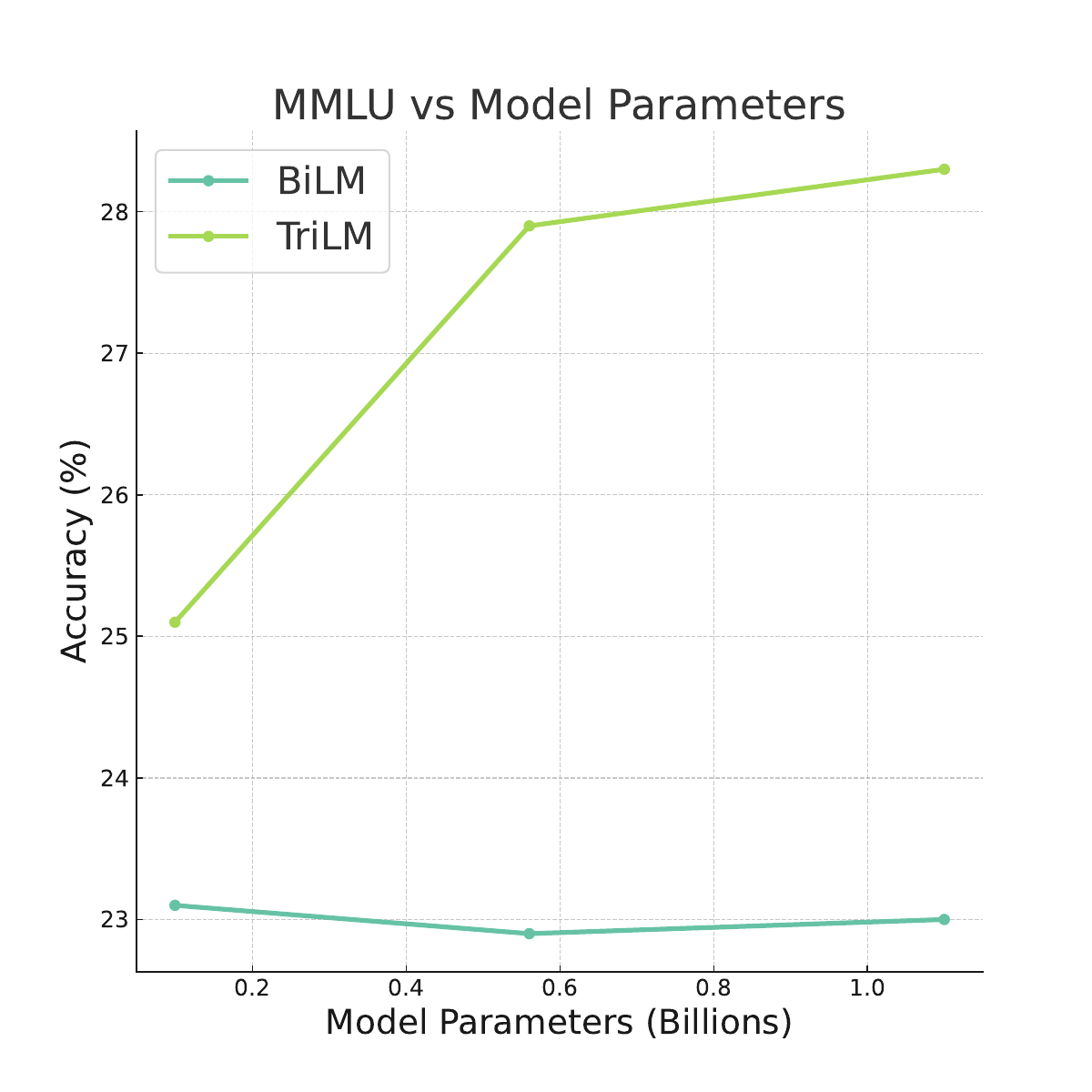}
                \caption{\tiny{vs. Params in MMLU}}
                \label{fig:TriLM_vs_BiLMs_MMLU_across_params}
            \end{subfigure}
            \caption{\footnotesize{Performance of ternary TriLMs and BiLMs models on MMLU across Size (Bits) and Parameters. Refer to Tables \ref{tab:evaluation_spectra_suite_part3} for details.}}
            \label{fig:BiLM_knowledge_performance}
        \end{figure}
\clearpage

\section{Scaling Law}
\label{appdx:scaling_law}
In this section, we provide additional insights into the scaling fits discussed in Section \ref{subsec:scaling_law}. In addition to fitting a power law with an offset, we also explore a standard power law following  \cite{kaplan2020scalinglawsneurallanguage}. Our findings suggest that the standard power law fits are slightly less precise than those incorporating an offset term. However, both models indicate a decreasing difference in validation loss as the number of parameters ($N$) increases. 

\begin{figure}[h]
    \centering
        \includegraphics[width=0.8\linewidth]{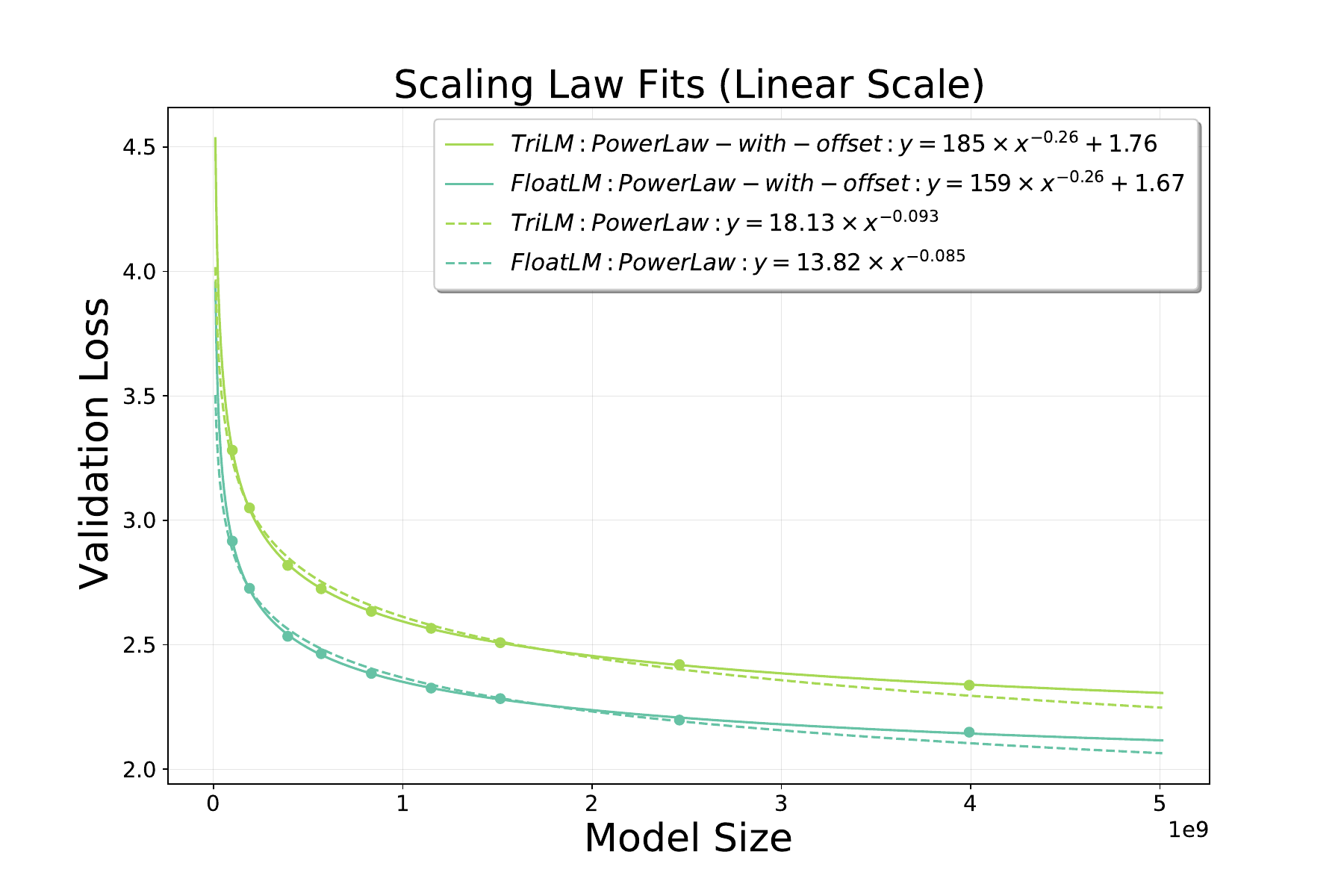}
        \includegraphics[width=0.8\linewidth]{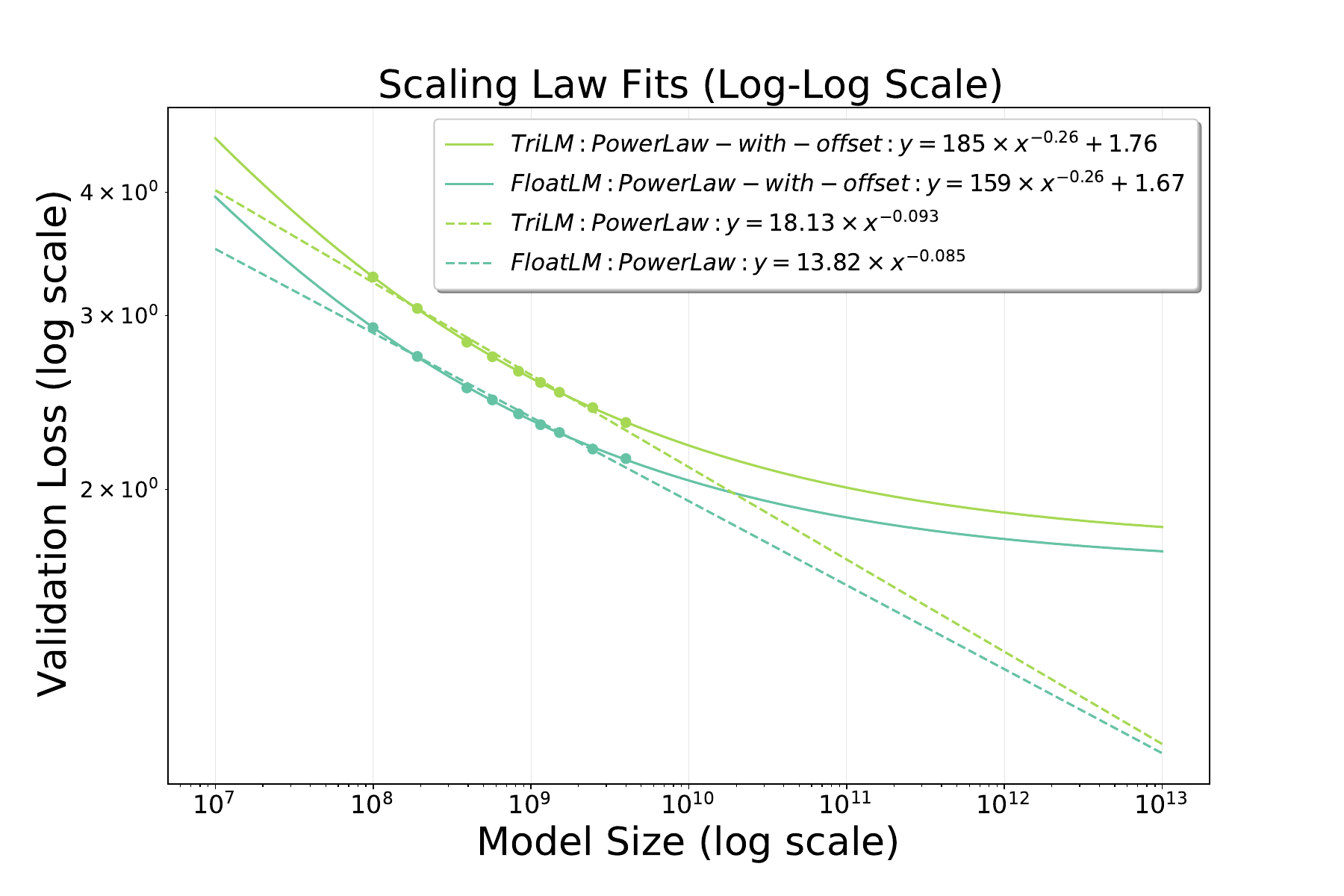}
        \caption{\footnotesize{Comparison of Power Law and Power Law-with-offset Fits for TriLM and FloatLM.}}
    \label{fig:scaling_law_model_fit_plot}
\end{figure}

\clearpage
\section{Benchmark Details} \label{appendix:Benchmark_Details}
        We benchmark TriLM, FloatLM and QuantLM across Knowledge, Commonsense, Reasoning and Toxicity benchmarks. We average our scores across 3 different `seeds' by preparing three different QuantLM models quantized using different calibration sets. We also add Pythia (deduplicated with consistent $2M$ batch size across families) suite of models (70M to 2.8B params) and BitNet b.158 performance scores from their paper for comparison. We use the LM Evaluation Harness \citep{eval-harness} to benchmark.

        \subsection{Commonsense and Reasoning} \label{appendix:Benchmark_Details::subsection:Commonsense_and_Reasoning}
            We report commonsense and reasoning benchmark scores across 6 benchmarks previously considered by BitNet b.158 in Table \ref{tab:evaluation_spectra_suite_part1.a}, \ref{tab:evaluation_spectra_suite_part1.b}  and rest in Table \ref{tab:evaluation_spectra_suite_part2}. Each is considered in a zero-shot setting. Following are the details of each of the benchmarks considered:

            \begin{itemize}
                \item \textbf{ARC Challenge and Easy}: \citep{ARC} ARC dataset comprises 7787 multiple-choice science questions divided into two sets: Challenge and Easy. We calculate accuracy and normalised accuracy across both of these sets.
                \item \textbf{BoolQ}: \citep{boolq} BoolQ is a reading comprehension dataset consisting of naturally occurring yes/no questions. We calculate the accuracy of this task.
                \item \textbf{HellaSwag}: \citep{hellaswag} HellaSWAG is a dataset of multiple choice questions for testing grounded commonsense. The incorrect options are generated through Adversarial Filtering (AF) to fool machines but not humans. We calculate the accuracy and normalized accuracy on this task.
                \item \textbf{WinoGrande}: \citep{winogrande} WinoGrande is a collection of 44k problems for testing commonsense reasoning formulated as a fill-in-a-blank task with binary options. We report the accuracy on this task.
                \item \textbf{PIQA}: \citep{piqa} Physical Interaction Question Answering (PIQA) is a physical commonsense reasoning benchmark dataset to test the physical knowledge of language models. We calculate the accuracy and normalized accuracy on this task.
                \item \textbf{LAMBADA OpenAI}: \citep{lambada} LAMBADA is a dataset to evaluate text understanding by next-word prediction. It is a collection of narrative passages BooksCorpus To succeed on LAMBADA, models must integrate broader discourse information, not solely rely on local context. We calculate the perplexity and the accuracy of the model on this task.
                \item \textbf{LogiQA}: \citep{logiqa} LogiQA is a dataset for testing human logical reasoning. It contains questions spanning multiple types of deductive reasoning. We calculate the accuracy and normalized accuracy on this task.
            \end{itemize}

\begin{table*}
\tiny
\centering
\begin{tabular}{lccccc}
\toprule
Models & \multicolumn{2}{c}{Arc Challenge} & \multicolumn{2}{c}{Arc Easy} & BoolQ \\
& Acc Norm. & Acc & Acc Norm. & Acc & Acc \\
\midrule
Pythia 70M & 22.0$\pm$ 1.2 & 22.1$\pm$ 1.2 & 24.8$\pm$ 0.9 & 24.8$\pm$ 0.9 & 38.5$\pm$ 0.9 \\
FloatLM 99M & 23.8$\pm$ 1.2 & 19.9$\pm$ 1.2 & 39.1$\pm$ 1.0 & 45.1$\pm$ 1.0 & 58.2$\pm$ 0.9 \\
QuantLM 99M 8-Bit & 23.8$\pm$ 1.2 & 19.6$\pm$ 1.2 & 39.4$\pm$ 1.0 & 45.3$\pm$ 1.0 & 58.5$\pm$ 0.9 \\
QuantLM 99M 6-Bit & 23.2$\pm$ 1.2 & 19.7$\pm$ 1.2 & 38.8$\pm$ 1.0 & 44.8$\pm$ 1.0 & 58.9$\pm$ 0.9 \\
QuantLM 99M 4-Bit & 22.6$\pm$ 1.2 & 18.0$\pm$ 1.1 & 37.1$\pm$ 1.0 & 41.7$\pm$ 1.0 & 52.2$\pm$ 0.9 \\
QuantLM 99M 3-Bit & 23.2$\pm$ 1.2 & 19.5$\pm$ 1.2 & 34.8$\pm$ 1.0 & 36.1$\pm$ 1.0 & 48.4$\pm$ 0.9 \\
TriLM 99M & 24.1$\pm$ 1.3 & 19.1$\pm$ 1.1 & 36.6$\pm$ 1.0 & 39.8$\pm$ 1.0 & 61.3$\pm$ 0.9 \\
Binary 99M & 20.8$\pm$ 1.2 & 18.3$\pm$ 1.1 & 35.8$\pm$ 0.9 & 40.1$\pm$ 1.0 & 61.0$\pm$ 0.8 \\
\hline
Pythia 160M & 23.8$\pm$ 1.2 & 23.1$\pm$ 1.2 & 26.7$\pm$ 0.9 & 26.6$\pm$ 0.9 & 38.3$\pm$ 0.9 \\
FloatLM 190M & 24.1$\pm$ 1.3 & 20.5$\pm$ 1.2 & 43.0$\pm$ 1.0 & 48.4$\pm$ 1.0 & 59.1$\pm$ 0.9 \\
QuantLM 190M 8-Bit & 24.4$\pm$ 1.3 & 20.3$\pm$ 1.2 & 43.0$\pm$ 1.0 & 48.5$\pm$ 1.0 & 59.3$\pm$ 0.9 \\
QuantLM 190M 6-Bit & 23.8$\pm$ 1.2 & 20.0$\pm$ 1.2 & 42.0$\pm$ 1.0 & 48.0$\pm$ 1.0 & 59.1$\pm$ 0.9 \\
QuantLM 190M 4-Bit & 25.2$\pm$ 1.3 & 19.9$\pm$ 1.2 & 26.5$\pm$ 0.9 & 26.8$\pm$ 0.9 & 40.9$\pm$ 0.9 \\
QuantLM 190M 3-Bit & 22.5$\pm$ 1.2 & 19.4$\pm$ 1.2 & 37.1$\pm$ 1.0 & 39.7$\pm$ 1.0 & 56.5$\pm$ 0.9 \\
TriLM 190M & 23.0$\pm$ 1.2 & 19.5$\pm$ 1.2 & 39.6$\pm$ 1.0 & 43.9$\pm$ 1.0 & 46.8$\pm$ 0.9 \\
\hline
FloatLM 390M & 24.7$\pm$ 1.3 & 21.3$\pm$ 1.2 & 46.5$\pm$ 1.0 & 51.0$\pm$ 1.0 & 54.7$\pm$ 0.9 \\
QuantLM 390M 8-Bit & 24.6$\pm$ 1.3 & 21.2$\pm$ 1.2 & 46.6$\pm$ 1.0 & 51.0$\pm$ 1.0 & 54.6$\pm$ 0.9 \\
QuantLM 390M 6-Bit & 24.8$\pm$ 1.3 & 21.5$\pm$ 1.2 & 46.8$\pm$ 1.0 & 51.8$\pm$ 1.0 & 55.3$\pm$ 0.9 \\
QuantLM 390M 4-Bit & 25.1$\pm$ 1.3 & 21.3$\pm$ 1.2 & 45.2$\pm$ 1.0 & 49.6$\pm$ 1.0 & 50.8$\pm$ 0.9 \\
QuantLM 390M 3-Bit & 24.9$\pm$ 1.3 & 21.5$\pm$ 1.2 & 41.6$\pm$ 1.0 & 43.6$\pm$ 1.0 & 56.3$\pm$ 0.9 \\
TriLM 390M & 24.5$\pm$ 1.3 & 21.2$\pm$ 1.2 & 44.1$\pm$ 1.0 & 48.6$\pm$ 1.0 & 55.1$\pm$ 0.9 \\
Pythia 410M & 24.7$\pm$ 1.3 & 21.2$\pm$ 1.2 & 45.7$\pm$ 1.0 & 51.6$\pm$ 1.0 & 60.0$\pm$ 0.9 \\
\hline
FloatLM 560M & 26.5$\pm$ 1.3 & 23.9$\pm$ 1.2 & 48.4$\pm$ 1.0 & 54.4$\pm$ 1.0 & 57.9$\pm$ 0.9 \\
QuantLM 560M 8-Bit & 26.5$\pm$ 1.3 & 23.6$\pm$ 1.2 & 48.3$\pm$ 1.0 & 54.1$\pm$ 1.0 & 57.6$\pm$ 0.9 \\
QuantLM 560M 6-Bit & 26.0$\pm$ 1.3 & 23.5$\pm$ 1.2 & 47.6$\pm$ 1.0 & 54.2$\pm$ 1.0 & 57.3$\pm$ 0.9 \\
QuantLM 560M 4-Bit & 25.9$\pm$ 1.3 & 23.0$\pm$ 1.2 & 46.3$\pm$ 1.0 & 52.4$\pm$ 1.0 & 58.8$\pm$ 0.9 \\
QuantLM 560M 3-Bit & 24.0$\pm$ 1.2 & 21.2$\pm$ 1.2 & 42.3$\pm$ 1.0 & 45.8$\pm$ 1.0 & 59.0$\pm$ 0.9 \\
TriLM 560M & 25.7$\pm$ 1.3 & 21.0$\pm$ 1.2 & 45.5$\pm$ 1.0 & 50.2$\pm$ 1.0 & 57.3$\pm$ 0.9 \\
Binary 560M & 24.6$\pm$ 1.2 & 20.2$\pm$ 1.1 & 41.9$\pm$ 1.0 & 47.8$\pm$ 1.0 & 61.5$\pm$ 0.8 \\
\hline
FloatLM 830M & 28.0$\pm$ 1.3 & 24.5$\pm$ 1.3 & 51.6$\pm$ 1.0 & 57.3$\pm$ 1.0 & 61.0$\pm$ 0.9 \\
QuantLM 830M 8-Bit & 28.2$\pm$ 1.3 & 25.1$\pm$ 1.3 & 51.7$\pm$ 1.0 & 57.3$\pm$ 1.0 & 60.9$\pm$ 0.9 \\
QuantLM 830M 6-Bit & 27.6$\pm$ 1.3 & 24.7$\pm$ 1.3 & 51.6$\pm$ 1.0 & 57.7$\pm$ 1.0 & 61.3$\pm$ 0.9 \\
QuantLM 830M 4-Bit & 27.6$\pm$ 1.3 & 23.3$\pm$ 1.2 & 50.5$\pm$ 1.0 & 56.2$\pm$ 1.0 & 58.1$\pm$ 0.9 \\
QuantLM 830M 3-Bit & 27.1$\pm$ 1.3 & 22.7$\pm$ 1.2 & 46.8$\pm$ 1.0 & 50.5$\pm$ 1.0 & 56.3$\pm$ 0.9 \\
TriLM 830M & 25.3$\pm$ 1.3 & 22.5$\pm$ 1.2 & 48.7$\pm$ 1.0 & 54.2$\pm$ 1.0 & 60.4$\pm$ 0.9 \\
\hline
Pythia 1B & 27.0$\pm$ 1.3 & 24.4$\pm$ 1.3 & 49.0$\pm$ 1.0 & 57.0$\pm$ 1.0 & 60.8$\pm$ 0.9 \\
FloatLM 1.1B & 29.1$\pm$ 1.3 & 26.1$\pm$ 1.3 & 54.0$\pm$ 1.0 & 60.4$\pm$ 1.0 & 62.9$\pm$ 0.8 \\
QuantLM 1.1B 8-Bit & 28.9$\pm$ 1.3 & 26.1$\pm$ 1.3 & 54.1$\pm$ 1.0 & 60.2$\pm$ 1.0 & 62.6$\pm$ 0.8 \\
QuantLM 1.1B 6-Bit & 29.8$\pm$ 1.3 & 25.5$\pm$ 1.3 & 54.3$\pm$ 1.0 & 60.2$\pm$ 1.0 & 62.9$\pm$ 0.8 \\
QuantLM 1.1B 4-Bit & 30.3$\pm$ 1.3 & 26.0$\pm$ 1.3 & 53.6$\pm$ 1.0 & 59.0$\pm$ 1.0 & 61.3$\pm$ 0.9 \\
QuantLM 1.1B 3-Bit & 29.2$\pm$ 1.3 & 27.0$\pm$ 1.3 & 48.9$\pm$ 1.0 & 55.0$\pm$ 1.0 & 62.1$\pm$ 0.8 \\
TriLM 1.1B & 26.5$\pm$ 1.3 & 24.6$\pm$ 1.3 & 49.8$\pm$ 1.0 & 56.3$\pm$ 1.0 & 59.1$\pm$ 0.9 \\
Binary 1.1B & 24.8$\pm$ 1.3 & 22.3$\pm$ 1.2 & 46.1$\pm$ 1.0 & 52.7$\pm$ 1.0 & 56.3$\pm$ 0.9 \\
\hline
Pythia 1.4B & 28.7$\pm$ 1.3 & 26.0$\pm$ 1.3 & 54.0$\pm$ 1.0 & 60.4$\pm$ 1.0 & 63.2$\pm$ 0.8 \\
FloatLM 1.5B & 29.7$\pm$ 1.3 & 26.2$\pm$ 1.3 & 56.4$\pm$ 1.0 & 62.6$\pm$ 1.0 & 63.2$\pm$ 0.8 \\
QuantLM 1.5B 8-Bit & 29.8$\pm$ 1.3 & 26.0$\pm$ 1.3 & 56.6$\pm$ 1.0 & 62.4$\pm$ 1.0 & 63.3$\pm$ 0.8 \\
QuantLM 1.5B 6-Bit & 30.1$\pm$ 1.3 & 26.0$\pm$ 1.3 & 56.8$\pm$ 1.0 & 62.2$\pm$ 1.0 & 63.4$\pm$ 0.8 \\
QuantLM 1.5B 4-Bit & 29.4$\pm$ 1.3 & 26.9$\pm$ 1.3 & 55.2$\pm$ 1.0 & 60.4$\pm$ 1.0 & 62.5$\pm$ 0.8 \\
QuantLM 1.5B 3-Bit & 27.8$\pm$ 1.3 & 25.2$\pm$ 1.3 & 49.7$\pm$ 1.0 & 54.8$\pm$ 1.0 & 53.7$\pm$ 0.9 \\
TriLM 1.5B & 28.2$\pm$ 1.3 & 24.7$\pm$ 1.3 & 53.1$\pm$ 1.0 & 59.0$\pm$ 1.0 & 54.1$\pm$ 0.9 \\
\hline
FloatLM 2.4B & 32.7$\pm$ 1.4 & 30.1$\pm$ 1.3 & 60.5$\pm$ 1.0 & 65.5$\pm$ 1.0 & 62.1$\pm$ 0.8 \\
QuantLM 2.4B 8-Bit & 32.6$\pm$ 1.4 & 30.0$\pm$ 1.3 & 60.3$\pm$ 1.0 & 65.7$\pm$ 1.0 & 62.1$\pm$ 0.8 \\
QuantLM 2.4B 6-Bit & 32.7$\pm$ 1.4 & 30.6$\pm$ 1.3 & 60.4$\pm$ 1.0 & 65.4$\pm$ 1.0 & 62.0$\pm$ 0.8 \\
QuantLM 2.4B 4-Bit & 33.3$\pm$ 1.4 & 30.8$\pm$ 1.3 & 59.6$\pm$ 1.0 & 64.1$\pm$ 1.0 & 59.0$\pm$ 0.9 \\
QuantLM 2.4B 3-Bit & 29.7$\pm$ 1.3 & 28.4$\pm$ 1.3 & 54.2$\pm$ 1.0 & 58.4$\pm$ 1.0 & 55.7$\pm$ 0.9 \\
TriLM 2.4B & 29.9$\pm$ 1.3 & 29.5$\pm$ 1.3 & 58.0$\pm$ 1.0 & 63.8$\pm$ 1.0 & 64.4$\pm$ 0.8 \\
\midrule
FloatLM 3.9B & 34.6$\pm$ 1.4 & 32.1$\pm$ 1.4 & 63.0$\pm$ 1.0 & 68.3$\pm$ 1.0 & 65.9$\pm$ 0.8 \\
QuantLM 3.9B 8-Bit & 34.6$\pm$ 1.4 & 31.9$\pm$ 1.4 & 63.0$\pm$ 1.0 & 68.1$\pm$ 1.0 & 65.4$\pm$ 0.8 \\
QuantLM 3.9B 6-Bit & 35.1$\pm$ 1.4 & 32.1$\pm$ 1.4 & 63.3$\pm$ 1.0 & 68.0$\pm$ 1.0 & 65.6$\pm$ 0.8 \\
QuantLM 3.9B 4-Bit & 34.7$\pm$ 1.4 & 32.9$\pm$ 1.4 & 61.2$\pm$ 1.0 & 68.3$\pm$ 1.0 & 65.4$\pm$ 0.8 \\
QuantLM 3.9B 3-Bit & 32.1$\pm$ 1.4 & 29.3$\pm$ 1.3 & 55.5$\pm$ 1.0 & 62.1$\pm$ 1.0 & 60.0$\pm$ 0.9 \\
TriLM 3.9B & 35.3$\pm$ 1.4 & 31.9$\pm$ 1.4 & 60.8$\pm$ 1.0 & 66.0$\pm$ 1.0 & 66.5$\pm$ 0.8 \\
\bottomrule
\end{tabular}
\caption{Spectra Suite Performance (Part 1): Arc Challenge, Arc Easy, and BoolQ. Additionally, we also include scores on the Pythia LLM suite.}
\label{tab:evaluation_spectra_suite_part1.a}
\end{table*}

\begin{table*}
\tiny
\centering
\begin{tabular}{lcccccc}
\toprule
Models & \multicolumn{2}{c}{HellaSwag} & \multicolumn{2}{c}{PIQA} & WinoGrande & Avg ( HellaSwag, PIQA, WinoGrande,\\
& Acc Norm. & Acc & Acc Norm. & Acc & Acc & Arc Easy, Arc Challenge, and BoolQ) \\
\midrule
Pythia 70M & 25.1$\pm$ 0.4 & 25.1$\pm$ 0.4 & 49.8$\pm$ 1.2 & 49.9$\pm$ 1.2 & 49.1$\pm$ 1.4 & 34.9 \\
FloatLM 99M & 31.6$\pm$ 0.5 & 29.1$\pm$ 0.5 & 62.8$\pm$ 1.1 & 63.2$\pm$ 1.1 & 50.2$\pm$ 1.4 & 44.3 \\
QuantLM 99M 8-Bit & 31.7$\pm$ 0.5 & 29.0$\pm$ 0.5 & 62.6$\pm$ 1.1 & 63.0$\pm$ 1.1 & 50.0$\pm$ 1.4 & 44.3 \\
QuantLM 99M 6-Bit & 31.7$\pm$ 0.5 & 29.2$\pm$ 0.5 & 62.8$\pm$ 1.1 & 63.1$\pm$ 1.1 & 50.2$\pm$ 1.4 & 44.3 \\
QuantLM 99M 4-Bit & 31.0$\pm$ 0.5 & 28.9$\pm$ 0.5 & 62.2$\pm$ 1.1 & 60.9$\pm$ 1.1 & 50.4$\pm$ 1.4 & 42.6 \\
QuantLM 99M 3-Bit & 29.2$\pm$ 0.5 & 27.7$\pm$ 0.4 & 57.2$\pm$ 1.2 & 58.2$\pm$ 1.2 & 49.2$\pm$ 1.4 & 40.3 \\
TriLM 99M & 28.4$\pm$ 0.5 & 27.6$\pm$ 0.4 & 60.1$\pm$ 1.1 & 60.4$\pm$ 1.1 & 50.7$\pm$ 1.4 & 43.5 \\
Binary 99M & 27.7$\pm$ 0.4 & 27.2$\pm$ 0.4 & 59.2$\pm$ 1.1 & 59.2$\pm$ 1.1 & 48.8$\pm$ 1.4 & 39.8 \\
\hline
Pythia 160M & 25.1$\pm$ 0.4 & 25.0$\pm$ 0.4 & 53.1$\pm$ 1.2 & 53.1$\pm$ 1.2 & 47.3$\pm$ 1.4 & 35.7 \\
FloatLM 190M & 36.6$\pm$ 0.5 & 31.4$\pm$ 0.5 & 65.6$\pm$ 1.1 & 64.8$\pm$ 1.1 & 51.9$\pm$ 1.4 & 46.7 \\
QuantLM 190M 8-Bit & 36.5$\pm$ 0.5 & 31.4$\pm$ 0.5 & 65.6$\pm$ 1.1 & 64.8$\pm$ 1.1 & 51.7$\pm$ 1.4 & 46.8 \\
QuantLM 190M 6-Bit & 36.3$\pm$ 0.5 & 31.5$\pm$ 0.5 & 65.6$\pm$ 1.1 & 64.3$\pm$ 1.1 & 51.9$\pm$ 1.4 & 46.4 \\
QuantLM 190M 4-Bit & 26.0$\pm$ 0.4 & 25.7$\pm$ 0.4 & 49.3$\pm$ 1.2 & 51.7$\pm$ 1.2 & 51.0$\pm$ 1.4 & 36.5 \\
QuantLM 190M 3-Bit & 32.0$\pm$ 0.5 & 28.8$\pm$ 0.5 & 58.1$\pm$ 1.2 & 58.7$\pm$ 1.1 & 50.1$\pm$ 1.4 & 42.7 \\
TriLM 190M & 31.6$\pm$ 0.5 & 29.0$\pm$ 0.5 & 62.0$\pm$ 1.1 & 62.3$\pm$ 1.1 & 51.7$\pm$ 1.4 & 42.4 \\
\hline
FloatLM 390M & 44.4$\pm$ 0.5 & 35.7$\pm$ 0.5 & 68.7$\pm$ 1.1 & 68.4$\pm$ 1.1 & 51.8$\pm$ 1.4 & 48.5 \\
QuantLM 390M 8-Bit & 44.5$\pm$ 0.5 & 35.7$\pm$ 0.5 & 68.8$\pm$ 1.1 & 68.6$\pm$ 1.1 & 52.6$\pm$ 1.4 & 48.6 \\
QuantLM 390M 6-Bit & 44.2$\pm$ 0.5 & 35.6$\pm$ 0.5 & 69.0$\pm$ 1.1 & 68.4$\pm$ 1.1 & 53.0$\pm$ 1.4 & 48.9 \\
QuantLM 390M 4-Bit & 43.4$\pm$ 0.5 & 35.1$\pm$ 0.5 & 68.1$\pm$ 1.1 & 68.3$\pm$ 1.1 & 53.7$\pm$ 1.4 & 47.7 \\
QuantLM 390M 3-Bit & 39.5$\pm$ 0.5 & 32.9$\pm$ 0.5 & 63.8$\pm$ 1.1 & 63.2$\pm$ 1.1 & 53.0$\pm$ 1.4 & 46.5 \\
TriLM 390M & 37.9$\pm$ 0.5 & 32.0$\pm$ 0.5 & 64.7$\pm$ 1.1 & 65.0$\pm$ 1.1 & 52.2$\pm$ 1.4 & 46.4 \\
Pythia 410M & 40.3$\pm$ 0.5 & 33.8$\pm$ 0.5 & 67.2$\pm$ 1.1 & 66.3$\pm$ 1.1 & 53.5$\pm$ 1.4 & 48.6 \\
\hline
FloatLM 560M & 47.6$\pm$ 0.5 & 37.7$\pm$ 0.5 & 68.8$\pm$ 1.1 & 69.0$\pm$ 1.1 & 53.7$\pm$ 1.4 & 50.5 \\
QuantLM 560M 8-Bit & 47.6$\pm$ 0.5 & 37.7$\pm$ 0.5 & 68.9$\pm$ 1.1 & 68.9$\pm$ 1.1 & 53.8$\pm$ 1.4 & 50.4 \\
QuantLM 560M 6-Bit & 47.6$\pm$ 0.5 & 37.7$\pm$ 0.5 & 68.7$\pm$ 1.1 & 68.8$\pm$ 1.1 & 53.5$\pm$ 1.4 & 50.1 \\
QuantLM 560M 4-Bit & 46.7$\pm$ 0.5 & 37.0$\pm$ 0.5 & 67.8$\pm$ 1.1 & 67.1$\pm$ 1.1 & 53.1$\pm$ 1.4 & 49.8 \\
QuantLM 560M 3-Bit & 41.7$\pm$ 0.5 & 33.4$\pm$ 0.5 & 63.5$\pm$ 1.1 & 63.2$\pm$ 1.1 & 49.7$\pm$ 1.4 & 46.7 \\
TriLM 560M & 41.5$\pm$ 0.5 & 33.8$\pm$ 0.5 & 67.2$\pm$ 1.1 & 67.5$\pm$ 1.1 & 53.1$\pm$ 1.4 & 48.4 \\
Binary 560M & 36.4$\pm$ 0.4 & 31.2$\pm$ 0.4 & 64.6$\pm$ 1.1 & 64.2$\pm$ 1.1 & 52.8$\pm$ 1.4 & 44.5 \\
\hline
FloatLM 830M & 51.3$\pm$ 0.5 & 40.1$\pm$ 0.5 & 71.4$\pm$ 1.1 & 71.7$\pm$ 1.1 & 56.4$\pm$ 1.4 & 53.3 \\
QuantLM 830M 8-Bit & 51.4$\pm$ 0.5 & 40.1$\pm$ 0.5 & 71.2$\pm$ 1.1 & 71.7$\pm$ 1.1 & 55.9$\pm$ 1.4 & 53.2 \\
QuantLM 830M 6-Bit & 51.5$\pm$ 0.5 & 40.2$\pm$ 0.5 & 71.3$\pm$ 1.1 & 71.8$\pm$ 1.0 & 56.2$\pm$ 1.4 & 53.2 \\
QuantLM 830M 4-Bit & 50.2$\pm$ 0.5 & 39.2$\pm$ 0.5 & 70.6$\pm$ 1.1 & 71.1$\pm$ 1.1 & 56.0$\pm$ 1.4 & 52.2 \\
QuantLM 830M 3-Bit & 45.5$\pm$ 0.5 & 35.9$\pm$ 0.5 & 66.1$\pm$ 1.1 & 66.6$\pm$ 1.1 & 53.5$\pm$ 1.4 & 49.2 \\
TriLM 830M & 46.0$\pm$ 0.5 & 36.8$\pm$ 0.5 & 68.2$\pm$ 1.1 & 68.4$\pm$ 1.1 & 55.6$\pm$ 1.4 & 50.7 \\
\hline
Pythia 1B & 47.2$\pm$ 0.5 & 37.7$\pm$ 0.5 & 69.3$\pm$ 1.1 & 70.8$\pm$ 1.1 & 53.2$\pm$ 1.4 & 51.1 \\
FloatLM 1.1B & 55.2$\pm$ 0.5 & 42.6$\pm$ 0.5 & 72.2$\pm$ 1.0 & 71.3$\pm$ 1.1 & 56.3$\pm$ 1.4 & 54.9 \\
QuantLM 1.1B 8-Bit & 55.2$\pm$ 0.5 & 42.6$\pm$ 0.5 & 72.1$\pm$ 1.0 & 71.2$\pm$ 1.1 & 56.2$\pm$ 1.4 & 54.8 \\
QuantLM 1.1B 6-Bit & 54.9$\pm$ 0.5 & 42.6$\pm$ 0.5 & 71.9$\pm$ 1.0 & 71.2$\pm$ 1.1 & 56.1$\pm$ 1.4 & 55.0 \\
QuantLM 1.1B 4-Bit & 54.9$\pm$ 0.5 & 42.0$\pm$ 0.5 & 71.6$\pm$ 1.1 & 70.4$\pm$ 1.1 & 54.8$\pm$ 1.4 & 54.4 \\
QuantLM 1.1B 3-Bit & 51.3$\pm$ 0.5 & 39.4$\pm$ 0.5 & 69.4$\pm$ 1.1 & 68.4$\pm$ 1.1 & 54.8$\pm$ 1.4 & 52.6 \\
TriLM 1.1B & 49.1$\pm$ 0.5 & 38.8$\pm$ 0.5 & 69.8$\pm$ 1.1 & 69.3$\pm$ 1.1 & 55.5$\pm$ 1.4 & 51.6 \\
Binary 1.1B & 43.4$\pm$ 0.5 & 35.1$\pm$ 0.4 & 66.9$\pm$ 1.1 & 68.3$\pm$ 1.1 & 55.3$\pm$ 1.4 & 47.1 \\
\hline
Pythia 1.4B & 52.0$\pm$ 0.5 & 40.4$\pm$ 0.5 & 70.8$\pm$ 1.1 & 70.6$\pm$ 1.1 & 57.1$\pm$ 1.4 & 54.3 \\
FloatLM 1.5B & 57.8$\pm$ 0.5 & 44.3$\pm$ 0.5 & 73.9$\pm$ 1.0 & 73.1$\pm$ 1.0 & 59.4$\pm$ 1.4 & 56.7 \\
QuantLM 1.5B 8-Bit & 57.8$\pm$ 0.5 & 44.3$\pm$ 0.5 & 73.7$\pm$ 1.0 & 73.1$\pm$ 1.0 & 59.4$\pm$ 1.4 & 56.8 \\
QuantLM 1.5B 6-Bit & 57.5$\pm$ 0.5 & 44.2$\pm$ 0.5 & 74.0$\pm$ 1.0 & 73.0$\pm$ 1.0 & 59.7$\pm$ 1.4 & 56.9 \\
QuantLM 1.5B 4-Bit & 56.9$\pm$ 0.5 & 43.2$\pm$ 0.5 & 72.7$\pm$ 1.0 & 72.4$\pm$ 1.0 & 57.1$\pm$ 1.4 & 55.6 \\
QuantLM 1.5B 3-Bit & 53.7$\pm$ 0.5 & 41.0$\pm$ 0.5 & 70.0$\pm$ 1.1 & 69.4$\pm$ 1.1 & 55.0$\pm$ 1.4 & 51.6 \\
TriLM 1.5B & 53.1$\pm$ 0.5 & 40.9$\pm$ 0.5 & 70.1$\pm$ 1.1 & 70.3$\pm$ 1.1 & 56.1$\pm$ 1.4 & 52.5 \\
\hline
FloatLM 2.4B & 62.7$\pm$ 0.5 & 47.1$\pm$ 0.5 & 75.2$\pm$ 1.0 & 74.9$\pm$ 1.0 & 61.8$\pm$ 1.4 & 59.2 \\
QuantLM 2.4B 8-Bit & 62.7$\pm$ 0.5 & 47.1$\pm$ 0.5 & 75.4$\pm$ 1.0 & 74.9$\pm$ 1.0 & 61.4$\pm$ 1.4 & 59.1 \\
QuantLM 2.4B 6-Bit & 62.9$\pm$ 0.5 & 47.0$\pm$ 0.5 & 75.7$\pm$ 1.0 & 74.7$\pm$ 1.0 & 61.1$\pm$ 1.4 & 59.1 \\
QuantLM 2.4B 4-Bit & 62.2$\pm$ 0.5 & 46.5$\pm$ 0.5 & 75.4$\pm$ 1.0 & 74.5$\pm$ 1.0 & 61.7$\pm$ 1.4 & 58.5 \\
QuantLM 2.4B 3-Bit & 58.6$\pm$ 0.5 & 43.5$\pm$ 0.5 & 72.7$\pm$ 1.0 & 70.8$\pm$ 1.1 & 57.2$\pm$ 1.4 & 54.7 \\
TriLM 2.4B & 59.0$\pm$ 0.5 & 45.3$\pm$ 0.5 & 72.6$\pm$ 1.0 & 71.4$\pm$ 1.1 & 59.7$\pm$ 1.4 & 57.3 \\
\midrule
FloatLM 3.9B & 66.1$\pm$ 0.5 & 49.7$\pm$ 0.5 & 75.8$\pm$ 1.0 & 75.4$\pm$ 1.0 & 62.8$\pm$ 1.4 & 61.4 \\
QuantLM 3.9B 8-Bit & 66.0$\pm$ 0.5 & 49.7$\pm$ 0.5 & 75.9$\pm$ 1.0 & 75.5$\pm$ 1.0 & 62.9$\pm$ 1.4 & 61.3 \\
QuantLM 3.9B 6-Bit & 65.9$\pm$ 0.5 & 49.7$\pm$ 0.5 & 75.5$\pm$ 1.0 & 75.6$\pm$ 1.0 & 62.2$\pm$ 1.4 & 61.3 \\
QuantLM 3.9B 4-Bit & 65.0$\pm$ 0.5 & 49.0$\pm$ 0.5 & 75.5$\pm$ 1.0 & 75.6$\pm$ 1.0 & 62.7$\pm$ 1.4 & 60.7 \\
QuantLM 3.9B 3-Bit & 61.2$\pm$ 0.5 & 45.9$\pm$ 0.5 & 72.6$\pm$ 1.0 & 72.3$\pm$ 1.0 & 59.3$\pm$ 1.4 & 56.8 \\
TriLM 3.9B & 64.7$\pm$ 0.5 & 48.3$\pm$ 0.5 & 74.6$\pm$ 1.0 & 74.4$\pm$ 1.0 & 62.1$\pm$ 1.4 & 60.7 \\
\bottomrule
\end{tabular}
\caption{Spectra Suite Performance (Part 2): HellaSwag, PIQA, WinoGrande, and Average Scores (including Arc Easy, Arc Challenge, and BoolQ). Additionally, we include scores from the Pythia LLM suite.}
\label{tab:evaluation_spectra_suite_part1.b}
\end{table*}

\begin{table*}
            \hspace*{-0.2 cm}
                \tiny
                \centering
                \begin{tabular}{lccccccccccc}
                    \toprule
                    Models & \multicolumn{2}{c}{Arc Challenge} & \multicolumn{2}{c}{Arc Easy} & BoolQ & \multicolumn{2}{c}{HellaSwag} & \multicolumn{2}{c}{PIQA} & WinoGrande &  Avg \\
                    & Acc Norm. & Acc & Acc Norm. & Acc & Acc & Acc Norm. & Acc & Acc Norm. & Acc & Acc &  \\
                    \midrule
                    
                    BitNet 700M & \multicolumn{2}{c}{21.4} & \multicolumn{2}{c}{51.8}  & 58.2 & \multicolumn{2}{c}{35.1} & \multicolumn{2}{c}{68.1} & 55.2 & 48.3 \\
                    BitNet 1.3B & \multicolumn{2}{c}{24.2} & \multicolumn{2}{c}{54.9} & 56.7 & \multicolumn{2}{c}{37.7} & \multicolumn{2}{c}{68.8} & 55.8 & 49.7 \\
                    BitNet 3B   & \multicolumn{2}{c}{28.3} & \multicolumn{2}{c}{61.4} & 61.5 & \multicolumn{2}{c}{42.9} & \multicolumn{2}{c}{71.5} & 59.3 & 54.2\\
                    BitNet 3.9B & \multicolumn{2}{c}{28.7} & \multicolumn{2}{c}{64.2}  & 63.5 & \multicolumn{2}{c}{44.2}& \multicolumn{2}{c}{73.2} & 60.5 & 55.7\\
                    \bottomrule
                \end{tabular}
                \caption{ \footnotesize{Performance of BitNet b1.58 on ARC Challenge, ARC Easy, BoolQ, HellaSwag, PIQA, and WinoGrande. The scores are taken from \citep{bitnet_b1.58}.} }
                \label{tab:evaluation_bitnet}
            \end{table*}

            \begin{table*}
                \scriptsize
                \centering
                % \begin{adjustbox}{0.7\textwidth}      
                \begin{tabular}{lcccccc}
                    \toprule
                    Models & \multicolumn{2}{c}{LAMBADA} & \multicolumn{2}{c}{SciQ} & \multicolumn{2}{c}{LogiQA} \\
                    & Perp. & Acc & Acc Norm. & Acc & Acc Norm. & Acc \\
                    \midrule
                    % Pythia 70M & NaN & 0.9$\pm$ 0.1 & 0.9$\pm$ 0.3 & 1.1$\pm$ 0.3 & 20.3$\pm$ 1.6 & 20.3$\pm$ 1.6 \\
                    FloatLM 99M & 85.0$\pm$ 6.9 & 26.5$\pm$ 0.6 & 62.9$\pm$ 1.5 & 73.6$\pm$ 1.4 & 27.6$\pm$ 1.8 & 21.2$\pm$ 1.6 \\
                    QuantLM 99M 8-Bit & 85.8$\pm$ 7.0 & 26.6$\pm$ 0.6 & 62.8$\pm$ 1.5 & 73.7$\pm$ 1.4 & 27.8$\pm$ 1.8 & 21.0$\pm$ 1.6 \\
                    QuantLM 99M 6-Bit & 89.9$\pm$ 7.4 & 26.1$\pm$ 0.6 & 61.8$\pm$ 1.5 & 73.9$\pm$ 1.4 & 28.1$\pm$ 1.8 & 20.3$\pm$ 1.6 \\
                    QuantLM 99M 4-Bit & 211.6$\pm$ 17.3 & 16.7$\pm$ 0.5 & 61.2$\pm$ 1.5 & 70.7$\pm$ 1.4 & 24.9$\pm$ 1.7 & 20.7$\pm$ 1.6 \\
                    QuantLM 99M 3-Bit & 4765.4$\pm$ 413.0 & 4.5$\pm$ 0.3 & 51.9$\pm$ 1.6 & 57.0$\pm$ 1.6 & 25.3$\pm$ 1.7 & 19.8$\pm$ 1.6 \\
                    TriLM 99M & 172.0$\pm$ 8.4 & 20.0$\pm$ 0.6 & 60.4$\pm$ 1.5 & 67.6$\pm$ 1.5 & 25.5$\pm$ 1.7 & 21.5$\pm$ 1.6 \\
                    Binary 99M & 468.3$\pm$ 24.1 & 14.0$\pm$ 0.4 & 54.4$\pm$ 1.6 & 62.5$\pm$ 1.5 & 27.0$\pm$ 1.7 & 22.3$\pm$ 1.6 \\
                    \hline
                    % Pythia 160M & NaN & 9.7$\pm$ 0.4 & 1.5$\pm$ 0.4 & 1.6$\pm$ 0.4 & 20.3$\pm$ 1.6 & 20.3$\pm$ 1.6 \\
                    FloatLM 190M & 50.3$\pm$ 2.7 & 31.1$\pm$ 0.6 & 65.1$\pm$ 1.5 & 77.3$\pm$ 1.3 & 27.2$\pm$ 1.7 & 22.1$\pm$ 1.6 \\
                    QuantLM 190M 8-Bit & 48.7$\pm$ 2.6 & 31.5$\pm$ 0.6 & 65.5$\pm$ 1.5 & 77.1$\pm$ 1.3 & 27.0$\pm$ 1.7 & 22.3$\pm$ 1.6 \\
                    QuantLM 190M 6-Bit & 55.3$\pm$ 3.0 & 30.0$\pm$ 0.6 & 64.2$\pm$ 1.5 & 77.0$\pm$ 1.3 & 26.1$\pm$ 1.7 & 22.4$\pm$ 1.6 \\
                    QuantLM 190M 4-Bit & 72479077.3 & 0.00$\pm$ 0.0 & 25.6$\pm$ 1.4 & 22.9$\pm$ 1.3 & 23.3$\pm$ 1.7 & 20.7$\pm$ 1.6 \\
                    QuantLM 190M 3-Bit & 664.5$\pm$ 41.1 & 12.4$\pm$ 0.5 & 58.5$\pm$ 1.6 & 66.4$\pm$ 1.5 & 26.3$\pm$ 1.7 & 21.0$\pm$ 1.6 \\
                    TriLM 190M & 130.7$\pm$ 6.5 & 23.7$\pm$ 0.6 & 61.0$\pm$ 1.5 & 72.6$\pm$ 1.4 & 25.5$\pm$ 1.7 & 21.5$\pm$ 1.6 \\
                    \hline
                    FloatLM 390M & 21.9$\pm$ 0.9 & 42.2$\pm$ 0.7 & 75.6$\pm$ 1.4 & 84.2$\pm$ 1.2 & 28.1$\pm$ 1.8 & 23.8$\pm$ 1.7 \\
                    QuantLM 390M 8-Bit & 21.7$\pm$ 0.9 & 42.3$\pm$ 0.7 & 75.7$\pm$ 1.4 & 84.1$\pm$ 1.2 & 28.3$\pm$ 1.8 & 24.1$\pm$ 1.7 \\
                    QuantLM 390M 6-Bit & 24.3$\pm$ 1.0 & 40.6$\pm$ 0.7 & 75.5$\pm$ 1.4 & 83.7$\pm$ 1.2 & 27.6$\pm$ 1.8 & 23.2$\pm$ 1.7 \\
                    QuantLM 390M 4-Bit & 30.2$\pm$ 1.3 & 39.1$\pm$ 0.7 & 77.1$\pm$ 1.3 & 84.1$\pm$ 1.2 & 25.8$\pm$ 1.7 & 23.3$\pm$ 1.7 \\
                    QuantLM 390M 3-Bit & 115.0$\pm$ 5.6 & 23.0$\pm$ 0.6 & 67.4$\pm$ 1.5 & 76.7$\pm$ 1.3 & 25.7$\pm$ 1.7 & 21.8$\pm$ 1.6 \\
                    TriLM 390M & 77.7$\pm$ 3.8 & 28.0$\pm$ 0.6 & 68.6$\pm$ 1.5 & 76.9$\pm$ 1.3 & 26.4$\pm$ 1.7 & 21.8$\pm$ 1.6 \\
                    % Pythia 410M & 11.9$\pm$ 0.4 & 49.9$\pm$ 0.7 & 70.8$\pm$ 1.4 & 80.9$\pm$ 1.2 & 28.7$\pm$ 1.8 & 21.8$\pm$ 1.6 \\
                    \hline
                    FloatLM 560M & 20.8$\pm$ 0.9 & 44.1$\pm$ 0.7 & 74.7$\pm$ 1.4 & 83.5$\pm$ 1.2 & 27.0$\pm$ 1.7 & 20.7$\pm$ 1.6 \\
                    QuantLM 560M 8-Bit & 20.9$\pm$ 0.9 & 44.2$\pm$ 0.7 & 74.7$\pm$ 1.4 & 83.6$\pm$ 1.2 & 27.3$\pm$ 1.7 & 20.7$\pm$ 1.6 \\
                    QuantLM 560M 6-Bit & 21.7$\pm$ 0.9 & 42.8$\pm$ 0.7 & 74.4$\pm$ 1.4 & 83.6$\pm$ 1.2 & 25.8$\pm$ 1.7 & 20.9$\pm$ 1.6 \\
                    QuantLM 560M 4-Bit & 24.9$\pm$ 1.1 & 40.8$\pm$ 0.7 & 73.6$\pm$ 1.4 & 82.0$\pm$ 1.2 & 27.0$\pm$ 1.7 & 21.7$\pm$ 1.6 \\
                    QuantLM 560M 3-Bit & 146.3$\pm$ 7.1 & 20.1$\pm$ 0.6 & 71.1$\pm$ 1.4 & 75.9$\pm$ 1.4 & 25.0$\pm$ 1.7 & 21.8$\pm$ 1.6 \\
                    TriLM 560M & 55.6$\pm$ 2.7 & 32.4$\pm$ 0.7 & 70.8$\pm$ 1.4 & 78.7$\pm$ 1.3 & 26.1$\pm$ 1.7 & 19.8$\pm$ 1.6 \\
                    Binary 560M & 62.8$\pm$ 3.0 & 31.0$\pm$ 0.6 & 70.0$\pm$ 1.4 & 78.8$\pm$ 1.3 & 26.7$\pm$ 1.7 & 21.5$\pm$ 1.6 \\
                    \hline
                    FloatLM 830M & 13.3$\pm$ 0.5 & 49.6$\pm$ 0.7 & 78.4$\pm$ 1.3 & 85.9$\pm$ 1.1 & 26.3$\pm$ 1.7 & 20.1$\pm$ 1.6 \\
                    QuantLM 830M 8-Bit & 13.5$\pm$ 0.5 & 49.4$\pm$ 0.7 & 78.5$\pm$ 1.3 & 86.1$\pm$ 1.1 & 26.6$\pm$ 1.7 & 20.0$\pm$ 1.6 \\
                    QuantLM 830M 6-Bit & 13.3$\pm$ 0.5 & 49.1$\pm$ 0.7 & 77.8$\pm$ 1.3 & 85.4$\pm$ 1.1 & 26.3$\pm$ 1.7 & 20.1$\pm$ 1.6 \\
                    QuantLM 830M 4-Bit & 15.4$\pm$ 0.6 & 47.3$\pm$ 0.7 & 78.8$\pm$ 1.3 & 85.1$\pm$ 1.1 & 25.5$\pm$ 1.7 & 21.2$\pm$ 1.6 \\
                    QuantLM 830M 3-Bit & 47.7$\pm$ 2.0 & 30.5$\pm$ 0.6 & 74.1$\pm$ 1.4 & 80.1$\pm$ 1.3 & 28.1$\pm$ 1.8 & 21.2$\pm$ 1.6 \\
                    TriLM 830M & 26.0$\pm$ 1.1 & 39.9$\pm$ 0.7 & 75.4$\pm$ 1.4 & 82.8$\pm$ 1.2 & 27.6$\pm$ 1.8 & 21.4$\pm$ 1.6 \\
                    \hline
                    % Pythia 1B & 7.9$\pm$ 0.2 & 56.1$\pm$ 0.7 & 76.0$\pm$ 1.4 & 83.8$\pm$ 1.2 & 29.8$\pm$ 1.8 & 22.1$\pm$ 1.6 \\
                    FloatLM 1.1B & 11.7$\pm$ 0.4 & 51.2$\pm$ 0.7 & 82.2$\pm$ 1.2 & 88.1$\pm$ 1.0 & 27.3$\pm$ 1.7 & 20.9$\pm$ 1.6 \\
                    QuantLM 1.1B 8-Bit & 11.7$\pm$ 0.4 & 51.2$\pm$ 0.7 & 82.1$\pm$ 1.2 & 88.1$\pm$ 1.0 & 27.8$\pm$ 1.8 & 21.2$\pm$ 1.6 \\
                    QuantLM 1.1B 6-Bit & 11.7$\pm$ 0.4 & 51.0$\pm$ 0.7 & 82.3$\pm$ 1.2 & 88.1$\pm$ 1.0 & 27.5$\pm$ 1.8 & 21.5$\pm$ 1.6 \\
                    QuantLM 1.1B 4-Bit & 13.9$\pm$ 0.5 & 49.3$\pm$ 0.7 & 81.2$\pm$ 1.2 & 87.6$\pm$ 1.0 & 28.4$\pm$ 1.8 & 20.3$\pm$ 1.6 \\
                    QuantLM 1.1B 3-Bit & 26.9$\pm$ 1.1 & 39.1$\pm$ 0.7 & 78.7$\pm$ 1.3 & 85.0$\pm$ 1.1 & 25.8$\pm$ 1.7 & 20.7$\pm$ 1.6 \\
                    TriLM 1.1B & 17.3$\pm$ 0.7 & 46.2$\pm$ 0.7 & 73.3$\pm$ 1.4 & 81.9$\pm$ 1.2 & 26.9$\pm$ 1.7 & 22.0$\pm$ 1.6 \\
                    Binary 1.1B & 33.4$\pm$ 1.4 & 37.6$\pm$ 0.6 & 71.1$\pm$ 1.4 & 81.2$\pm$ 1.3 & 28.4$\pm$ 1.7 & 23.2$\pm$ 1.6 \\
                    \hline
                    % Pythia 1.4B & 6.1$\pm$ 0.2 & 61.5$\pm$ 0.7 & 79.1$\pm$ 1.3 & 86.7$\pm$ 1.1 & 27.8$\pm$ 1.8 & 20.9$\pm$ 1.6 \\
                    FloatLM 1.5B & 9.4$\pm$ 0.3 & 55.5$\pm$ 0.7 & 80.9$\pm$ 1.2 & 87.4$\pm$ 1.0 & 26.1$\pm$ 1.7 & 20.9$\pm$ 1.6 \\
                    QuantLM 1.5B 8-Bit & 9.5$\pm$ 0.3 & 55.5$\pm$ 0.7 & 81.3$\pm$ 1.2 & 87.5$\pm$ 1.0 & 25.7$\pm$ 1.7 & 20.6$\pm$ 1.6 \\
                    QuantLM 1.5B 6-Bit & 9.5$\pm$ 0.3 & 55.4$\pm$ 0.7 & 81.4$\pm$ 1.2 & 87.6$\pm$ 1.0 & 25.7$\pm$ 1.7 & 20.3$\pm$ 1.6 \\
                    QuantLM 1.5B 4-Bit & 10.4$\pm$ 0.4 & 53.0$\pm$ 0.7 & 81.1$\pm$ 1.2 & 86.9$\pm$ 1.1 & 25.7$\pm$ 1.7 & 20.3$\pm$ 1.6 \\
                    QuantLM 1.5B 3-Bit & 17.8$\pm$ 0.7 & 45.3$\pm$ 0.7 & 75.5$\pm$ 1.4 & 82.1$\pm$ 1.2 & 28.4$\pm$ 1.8 & 22.7$\pm$ 1.6 \\
                    TriLM 1.5B & 16.4$\pm$ 0.7 & 46.2$\pm$ 0.7 & 80.7$\pm$ 1.2 & 87.3$\pm$ 1.1 & 27.8$\pm$ 1.8 & 21.5$\pm$ 1.6 \\
                    \hline
                    FloatLM 2.4B & 7.7$\pm$ 0.3 & 59.3$\pm$ 0.7 & 87.2$\pm$ 1.1 & 91.0$\pm$ 0.9 & 29.5$\pm$ 1.8 & 21.5$\pm$ 1.6 \\
                    QuantLM 2.4B 8-Bit & 7.7$\pm$ 0.3 & 59.2$\pm$ 0.7 & 87.1$\pm$ 1.1 & 91.0$\pm$ 0.9 & 29.5$\pm$ 1.8 & 21.5$\pm$ 1.6 \\
                    QuantLM 2.4B 6-Bit & 7.9$\pm$ 0.3 & 58.9$\pm$ 0.7 & 87.3$\pm$ 1.1 & 90.9$\pm$ 0.9 & 29.6$\pm$ 1.8 & 20.9$\pm$ 1.6 \\
                    QuantLM 2.4B 4-Bit & 8.9$\pm$ 0.3 & 56.1$\pm$ 0.7 & 84.8$\pm$ 1.1 & 89.7$\pm$ 1.0 & 29.6$\pm$ 1.8 & 20.9$\pm$ 1.6 \\
                    QuantLM 2.4B 3-Bit & 15.6$\pm$ 0.6 & 45.0$\pm$ 0.7 & 79.9$\pm$ 1.3 & 86.7$\pm$ 1.1 & 28.6$\pm$ 1.8 & 21.4$\pm$ 1.6 \\
                    TriLM 2.4B & 8.6$\pm$ 0.3 & 55.7$\pm$ 0.7 & 84.2$\pm$ 1.2 & 88.7$\pm$ 1.0 & 28.6$\pm$ 1.8 & 24.3$\pm$ 1.7 \\
                    % Pythia 2.8B & 5.1$\pm$ 0.1 & 64.6$\pm$ 0.7 & 83.6$\pm$ 1.2 & 88.5$\pm$ 1.0 & 28.3$\pm$ 1.8 & 22.0$\pm$ 1.6 \\
                    \midrule
                    FloatLM 3.9B & 6.7$\pm$ 0.2 & 61.1$\pm$ 0.7 & 86.5$\pm$ 1.1 & 90.9$\pm$ 0.9 & 26.9$\pm$ 1.7 & 20.9$\pm$ 1.6 \\
                    QuantLM 3.9B 8-Bit & 6.7$\pm$ 0.2 & 61.1$\pm$ 0.7 & 86.2$\pm$ 1.1 & 91.0$\pm$ 0.9 & 26.6$\pm$ 1.7 & 20.6$\pm$ 1.6 \\
                    QuantLM 3.9B 6-Bit & 6.8$\pm$ 0.2 & 60.8$\pm$ 0.7 & 86.6$\pm$ 1.1 & 91.3$\pm$ 0.9 & 25.8$\pm$ 1.7 & 20.4$\pm$ 1.6 \\
                    QuantLM 3.9B 4-Bit & 7.4$\pm$ 0.2 & 58.5$\pm$ 0.7 & 86.1$\pm$ 1.1 & 90.8$\pm$ 0.9 & 28.6$\pm$ 1.8 & 20.1$\pm$ 1.6 \\
                    QuantLM 3.9B 3-Bit & 14.0$\pm$ 0.5 & 47.1$\pm$ 0.7 & 83.1$\pm$ 1.2 & 88.6$\pm$ 1.0 & 27.0$\pm$ 1.7 & 21.5$\pm$ 1.6 \\
                    TriLM 3.9B & 6.3$\pm$ 0.2 & 61.6$\pm$ 0.7 & 87.4$\pm$ 1.0 & 90.8$\pm$ 0.9 & 27.6$\pm$ 1.8 & 22.7$\pm$ 1.6 \\
                    \bottomrule
                \end{tabular}
                % \end{adjustbox}
          \caption{\footnotesize{Spectra Suite Performance (Part 3): LAMBADA OpenAI, SciQ, LogiQA. We additionally also include Pythia's performance scores.}}
        \label{tab:evaluation_spectra_suite_part2}
    \end{table*}

        \subsection{Knowledge} \label{appendix:Benchmark_Details::subsection:Knowledge}

            We report performance on SciQ, TriviaQA in Tables \ref{tab:evaluation_spectra_suite_part2},
            \ref{tab:evaluation_spectra_suite_part3} and
            \ref{tab:evaluation_spectra_mmlu}. Each is considered in a zero-shot setting. Following are the details of each of the benchmarks considered:

            The knowledge-based evaluation included the following tasks:
            \begin{itemize}
                \item \textbf{SciQ}: \citep{sciq} The SciQ dataset contains multiple-choice questions with 4 answer options from crowd-sourced science exams. The questions range from Physics, Chemistry and Biology and several other fields. We calculate the accuracy and length normalized accuracy on this task.

                \item \textbf{TriviaQA}: \citep{triviaqa} TriviaQA is a reading comprehension dataset containing question-answer-evidence triples. We calculate the exact match accuracy on this task.
                
                \item \textbf{MMLU} \citep{mmlu}: The benchmark aims to assess the knowledge gained during pretraining by evaluating models solely in zero-shot and few-shot scenarios. It spans 57 subjects, including STEM fields, humanities, social sciences, and more.

            \end{itemize}

        \subsection{Toxicity} \label{appendix:Benchmark_Details::subsection:Toxicity}
            We report toxicity-based evaluation in \ref{tab:evaluation_spectra_suite_part3}. Each is considered in a zero-shot setting. 

            The toxicity-based evaluation included the following tasks:
            \begin{itemize}
                \item \textbf{BBQ} \citep{bbq}: The Bias Benchmark for QA (BBQ) dataset, comprises sets of questions developed by its authors, focusing on documented social biases directed towards individuals from protected classes across nine distinct social dimensions pertinent to U.S. English-speaking environments.

                \item \textbf{Crows Pairs} \citep{crows_pairs}: proposed a challenging dataset aimed at quantifying stereotypical biases embedded within language models, with a specific emphasis on U.S. contexts. Hosted on GitHub, this dataset serves as a crucial resource for assessing and addressing biases through paired sentences that illuminate societal stereotypes.

                \item \textbf{TruthfulQA} \citep{truthfulqa}: A benchmark designed to evaluate the truthfulness of language models in generating responses to questions. This benchmark includes 817 questions across 38 categories, such as health, law, finance, and politics.
            \end{itemize}

\clearpage

\section{Weight Distribution of Linear Layers}
\label{appdx:weight_distribution}
  
     \begin{figure}[h!]
            \centering
                \includegraphics[width=0.78\linewidth]{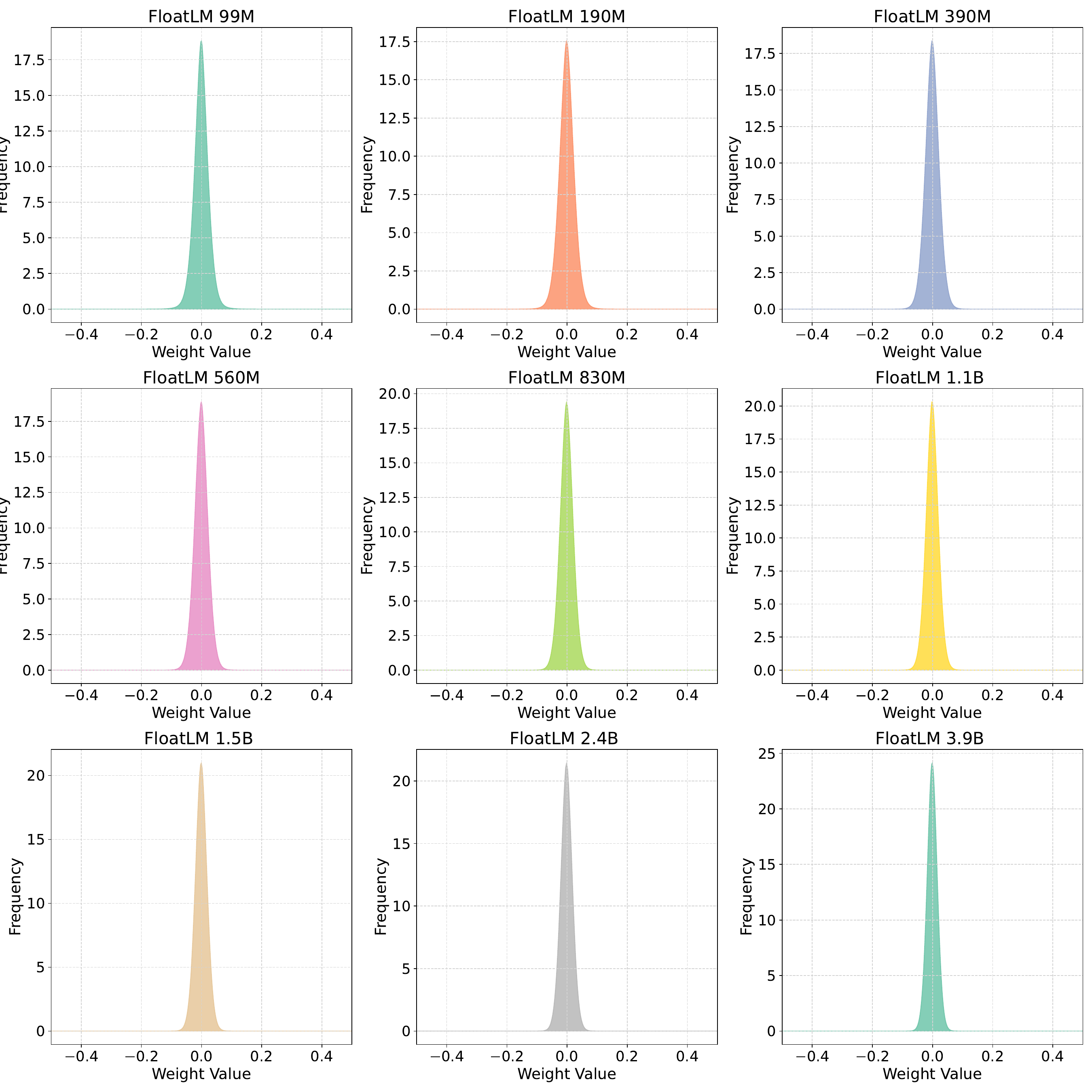}
                \caption{\footnotesize{Weight distribution in the linear layers of FloatLM models across various model sizes, ranging from 99M to 3.9B parameters}}
            \label{fig:weight_distribution_across_scale}
        \end{figure}

The observed Gaussian distribution in the weights of our final trained FloatLM models across various scales is supported by both theoretical foundations and empirical evidence. We outline the rationale for the normality of weight distributions as follows:

\paragraph{Empirical Consistency:} Across all model scales, ranging from 99 million to 3.9 billion parameters, the weight distributions consistently exhibit Gaussian characteristics, as illustrated in Figure \ref{fig:weight_distribution_across_scale}. This consistency is visually evident in distribution plots and quantitatively confirmed by fitting Gaussian functions to the weight histograms, demonstrating a good fit across different model sizes.

\paragraph{Theoretical Underpinning:} Neural network weight initialization typically follows a Gaussian distribution to facilitate balanced learning dynamics. As training progresses, despite non-linear transformations and complex interactions within the network, the Central Limit Theorem suggests that the aggregation of numerous independent random variables (such as updates during backpropagation) tends toward a normal distribution. This tendency is particularly pronounced given the high dimensionality and extensive data processing involved in training large language models.

\paragraph{Stabilization through Regularization and Optimization:} Techniques such as L2 regularization constrain weight magnitudes, encouraging them towards smaller values and contributing to a peak around zero—a characteristic feature of Gaussian distributions. Additionally, optimization algorithms like Adam, which adjust learning rates based on moving averages of recent gradients, promote smoother updates. This approach maintains the Gaussian form by mitigating the impact of outlier gradients.

Given these, we can assert that the weight distribution of our trained models closely follows a Gaussian distribution which is crucial for understanding the weight variance across different scales.

\clearpage

\section{Memory Bottlenecks and Low-Bitwidth Language Modelling}
\label{app:memory_analysis_for_LLM}

% \begin{figure}[h]
%     \centering
%     \begin{subfigure}[b]{0.45\textwidth}
%         \includegraphics[width=\textwidth]{figures/trends_decreasing_memory_to_flop_ratio.pdf}
%         \caption{Memory Capacity vs Peak FLOPS}
%         \label{fig:trends_decreasing_memory_to_flop_ratio}
%     \end{subfigure}%
%     \begin{subfigure}[b]{0.45\textwidth}
%         \includegraphics[width=\textwidth]{figures/trends_decreasing_bytes_to_flop_ratio.pdf}
%         \caption{Peak Memory Bandwidth vs Peak FLOPS}
%         \label{fig:trends_decreasing_bytes_to_flop_ratio}
%     \end{subfigure}
%     \caption{Trends of Memory/FLOP and Bandwidth/FLOP across different (datacenter) GPGPUs.}
%     \label{fig:memory_flop_bandwidth_trends}
% \end{figure}
    
Recent observations \citep{gholami2024aimemorywall} suggest that, given the slower pace of improvements in memory and communication compared to compute (FLOPs), the bottleneck continues to shift away from computation towards the memory-related characteristics of hardware for deploying large language models. This shift underscores the importance of exploring solutions that directly address memory constraints. Below, we formally analyze this trend and the impact of low-bitwidth language models on addressing memory bottlenecks during inference.

   \subsection{Overview of Recent Datacenter GPUs and Accelerators.}
    
     % We begin our analysis with a wider range of recent datacenter General Purpose GPUs (GPGPUs) used for neural network development and research since 2018 from multiple hardware providers. We consider different configurations across the recent microarchitectures. These include Volta (V100 SXM/PCIe) \citep{hardware_datasheet_nvidia_v100}, Ampere (A100 40GB/80GB SXM/PCIe) \citep{hardware_datasheet_nvidia_a100}, Hopper (H100 SXM/PCIe, H200) \citep{hardware_datasheet_nvidia_h100, hardware_datasheet_nvidia_h200} and Blackwell \citep{hardware_datasheet_nvidia_blackwell}\footnote{At the time of access, preliminary specifications for Blackwell were subject to change.} from Nvidia, and also the MI200 Series (MI210, MI250, MI250X) \citep{hardware_datasheet_amd_mi210, hardware_datasheet_amd_mi250_mi250x}, and MI300 Series (MI300A, MI300X, MI325X) \citep{hardware_datasheet_amd_mi300a, hardware_datasheet_amd_mi300x, hardware_datasheet_amd_mi325x} from AMD. Additionally, we consider Gaudi 2 and Gaudi 3 \citep{hardware_datasheet_intel_gaudi2_gaudi3} from Intel, as well as TPUv3 \citep{hardware_datasheet_google_tpuv3}, TPUv4 \citep{hardware_datasheet_google_tpuv4}, TPUv5 (TPUv5e, TPUv5p) \citep{hardware_datasheet_google_tpuv5e, hardware_datasheet_google_tpuv5p} from Google. We obtained all our data from their respectively cited datasheets, documentation or press releases. 

     We begin our analysis by surveying a broad range of recent datacenter General Purpose GPUs (GPGPUs) employed for neural network development and research since 2018. This includes hardware from multiple providers, covering various configurations across the latest microarchitectures.

From Nvidia, we consider the following:
\begin{itemize}
    \item \textbf{Volta}: V100 (SXM/PCIe) \citep{hardware_datasheet_nvidia_v100},
    \item \textbf{Ampere}: A100 (40GB/80GB SXM/PCIe) \citep{hardware_datasheet_nvidia_a100},
    \item \textbf{Hopper}: H100 (SXM/PCIe) and H200 \citep{hardware_datasheet_nvidia_h100, hardware_datasheet_nvidia_h200},
    \item \textbf{Blackwell}: This includes preliminary data for Blackwell microarchitectures, which at the time of access were subject to change \citep{hardware_datasheet_nvidia_blackwell}.
\end{itemize}

From AMD, we analyze the following models:
\begin{itemize}
    \item \textbf{MI200 Series}: MI210, MI250, MI250X \citep{hardware_datasheet_amd_mi210, hardware_datasheet_amd_mi250_mi250x},
    \item \textbf{MI300 Series}: MI300A, MI300X, MI325X \citep{hardware_datasheet_amd_mi300a, hardware_datasheet_amd_mi300x, hardware_datasheet_amd_mi325x}.
\end{itemize}

Additionally, we include hardware from Intel and Google:
\begin{itemize}
    \item From Intel, the \textbf{Gaudi Series}: Gaudi 2 and Gaudi 3 \citep{hardware_datasheet_intel_gaudi2_gaudi3},
    \item From Google, the \textbf{Tensor Processing Units (TPUs)}: TPUv3 \citep{hardware_datasheet_google_tpuv3}, TPUv4 \citep{hardware_datasheet_google_tpuv4}, and TPUv5 (TPUv5e, TPUv5p) \citep{hardware_datasheet_google_tpuv5e, hardware_datasheet_google_tpuv5p}.
\end{itemize}

All data was sourced from the respective datasheets, technical documentation, or press releases of the cited hardware. Over the past several years, each of these four accelerator families has improved in three areas - FLOPS, memory capacity, and bandwidth. 
     
     % For our analysis, we consider the configurations of transformers in the LLaMa-family, Falcon-180B, and Nemotron 340B. Since larger vocabulary in LLMs are becoming common for efficient multilingual modeling, we use the vocabulary size of 128k from LLaMa 3 for our analysis. We assume the Embedding and LM Head weights are retained in Half-Precision across all bitwidths for these analyses.
     
     \subsection{Memory Trends and Speedup Opportunities in Low-Bitwidth Language Modeling}   
\begin{figure}[h!]
        \centering
        \begin{subfigure}[b]{0.47\textwidth}
            \includegraphics[width=\textwidth]{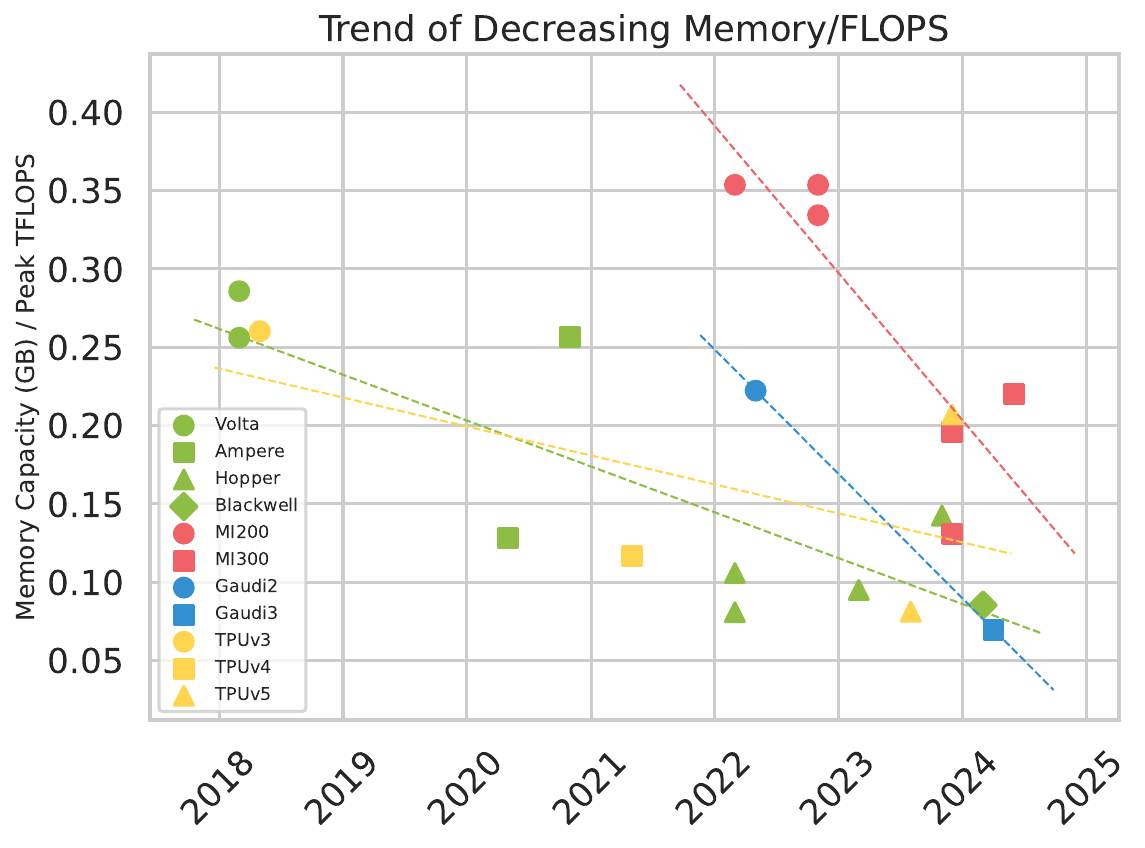}
            \caption{\footnotesize{Memory Capacity vs Peak FLOPS}}
            \label{fig:trends_decreasing_memory_to_flop_ratio}
        \end{subfigure}%
        \begin{subfigure}[b]{0.47\textwidth}
            \includegraphics[width=\textwidth]{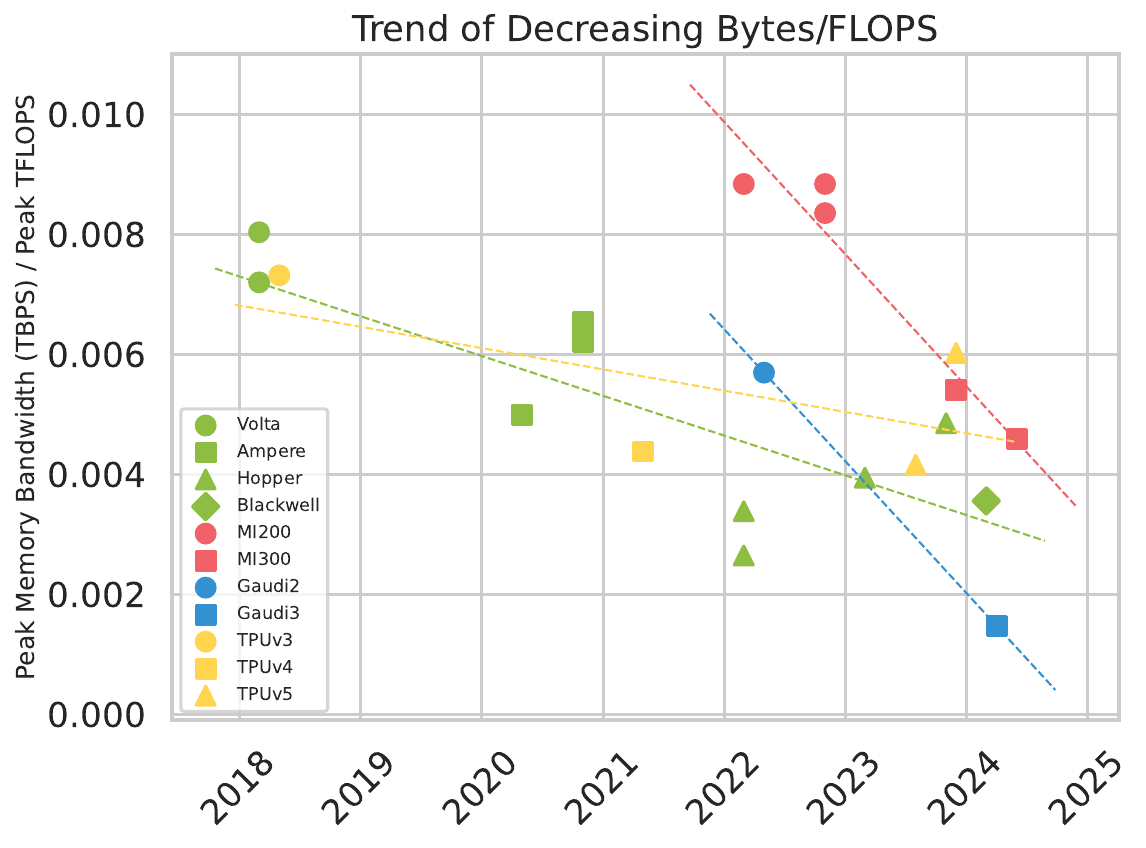}
            \caption{\footnotesize{Peak Memory Bandwidth vs Peak FLOPS}}
            \label{fig:trends_decreasing_bytes_to_flop_ratio}
        \end{subfigure}
        \caption{\footnotesize{Trends of Memory/FLOP and Bandwidth/FLOP across different (datacenter) GPGPUs.}}
        \label{fig:memory_flop_bandwidth_trends}
    \end{figure}

    \paragraph{Memory Capacity and Bandwidth of GPGPUs Relative to Peak TFLOPs.}
        In Figure \ref{fig:trends_decreasing_memory_to_flop_ratio}, we show the trends of Memory Capacity over Peak TFLOPS (Half Precision - FP16/BF16) for various accelerators over the years. We also perform a linear fit for each family of accelerators separately. The linear fit for all the families has a downward slope, showing that memory capacity is improving at a slower pace than computation capability. This trend holds true even for the most recent hardware, such as Blackwell, MI325X, and Gaudi3. Though we consider Half-Precision TFLOPs, the slope is expected to become steeper when considering peak TFLOPS over Ampere sparse or FP8. Similarly, in Figure \ref{fig:trends_decreasing_bytes_to_flop_ratio}, we present the trends of Memory Bandwidth (specifically for DRAM or its equivalent memory) over FLOPs for the accelerators over the years, along with the linear fit for each family. We observe a downward slope here as well, indicating the trend that memory bandwidth is growing much slower than computation. 

        \paragraph{Memory Wall and Speedup Opportunities.}\cite{kim2024squeezellm} established the memory wall in autoregressive LLM computation. They found that the speed of token generation is bottlenecked by the rate at which data is fed from memory to processors, rather than the processing speed of the hardware. As a result, the autoregressive decoding of LLM inference can have a theoretical speedup proportional to its compression factor. Various efficient inference kernels over quantized models have realized this speedup in low batch settings across a variety of hardware. This includes CPUs \footnote{https://github.com/ggerganov/llama.cpp}, consumer GPUs \footnote{https://github.com/turboderp/exllamav2} and data center GPUs \citep{triton_acceleration}. However, since TFLOPS to bandwidth ratio is up to 500 times, this ideal speedup can also be achieved in much higher batch settings encountered in LLM deployment. Open-source kernels like Marlin \citep{frantar2024marlin} have demonstrated that these ideal speedups can also be consistently realized in high batch size scenarios and sustained over longer periods of time.

        In Figure \ref{fig:model_size_growth_at_scale_across_bitwidth}, we show the size of models (in GB) across parameter count for two low bitwidth modeling scenarios, TriLM and QuantLM 4-Bit along with the standard half-precision FloatLM. For simplicity, we do not consider the overhead of KV Cache, activations, and compilation overhead incurred during model deployment. The FloatLM model starts to reach the capacity of a single H100 at just 34B parameters. At 340 Billion (the size of Nemotron 4) is more than the capacity of a single 8xH100 node. QuantLM 4-Bit scales better, easily supporting the deployment of a 70 billion parameter model (like largest LLaMa 1 and 2) on a single H100 and 300B parameter models on a single MI300X. However, TriLMs with more than 300 billion parameters, with appropriate packing, can fit on a single H100. This feature makes TriLMs especially crucial for deployment at the edge, where devices have less than 8GB or 16GB of RAM, shared across the operating system and multiple applications.
    
    \paragraph{Memory bandwidth and Memory bandwidth of GPGPUs and model inference speedup:}

    \cite{kim2024squeezellm} established the memory wall in autoregressive LLM computation. They found that the speed of token generation is bottlenecked by the rate at which data is fed from memory to processors, rather than the processing speed of the hardware. As a result, the autoregressive decoding of LLM inference can have a theoretical speedup proportional to its compression factor.

    Various efficient inference kernels over quantized models have realized this speedup in low batch settings across a variety of hardware. This includes CPUs \footnote{https://github.com/ggerganov/llama.cpp}, consumer GPUs \footnote{https://github.com/turboderp/exllamav2} and data center GPUs \citep{triton_acceleration}. However, since TFLOPS to bandwidth ratio is up to 500 times, this ideal speedup can also be achieved in much higher batch settings encountered in LLM deployment. Open-source kernels like Marlin \citep{frantar2024marlin} have demonstrated that these ideal speedups can also be consistently realized in high batch size scenarios and sustained over longer periods of time. In Figure \ref{fig:maximum_speedup_at_scale_across_bitwidth}, we show the (theoretically) maximum possible speedup relative to FP16 at varying parameter counts for QuantLM 4-Bit and TriLM. Even at 7 billion parameters, TriLMs can be more than 4 times faster at autoregressive decoding than FloatLM and 2 times faster than QuantLM 4-bit. While QuantLM 4-Bit plateaus at a maximum possible speedup factor of 4x, TriLMs plateau much higher at 10x for FloatLM.
\clearpage
    \begin{figure}[h]
        \centering
        \begin{subfigure}[b]{0.4\textwidth}
            \centering
            \includegraphics[width=\linewidth, height=0.85\textwidth]{figures/MMLUs/MMLU_Stem_across_size.pdf}
            \label{fig:MMLU_Stem_across_size_repeat}
        \end{subfigure}
        \begin{subfigure}[b]{0.4\textwidth}
            \centering
            \includegraphics[width=\linewidth , height=0.85\textwidth]{figures/MMLUs/MMLU_Stem_across_params.pdf}
            \label{fig:MMLU_Stem_across_params_repeat}
        \end{subfigure}

        \begin{subfigure}[b]{0.4\textwidth}
            \centering
            \includegraphics[width=\linewidth, height=0.85\textwidth]{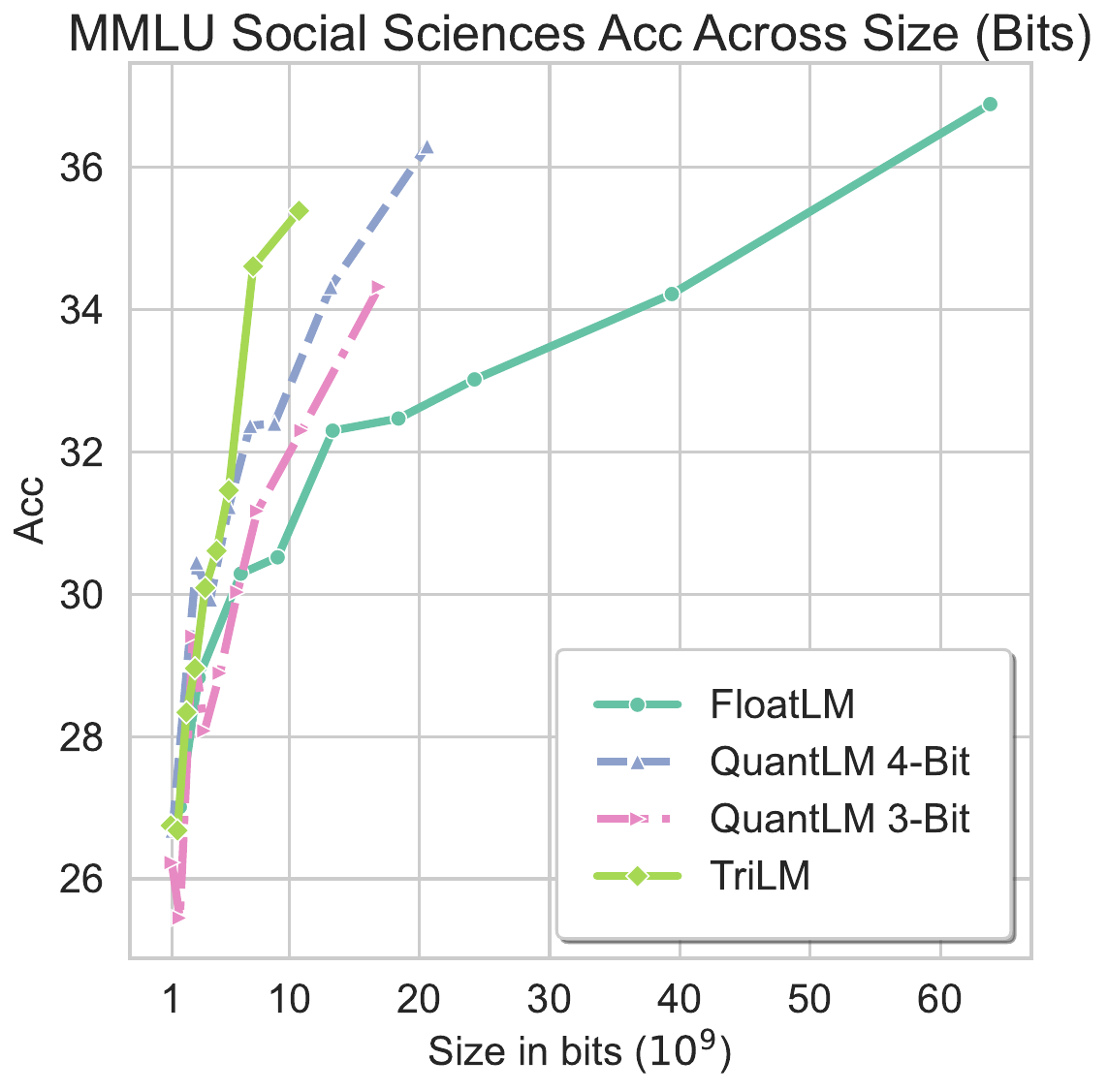}
            \label{fig:MMLU_Social_Sciences_across_size}
        \end{subfigure}
        \begin{subfigure}[b]{0.4\textwidth}
            \centering
            \includegraphics[width=\linewidth , height=0.85\textwidth]{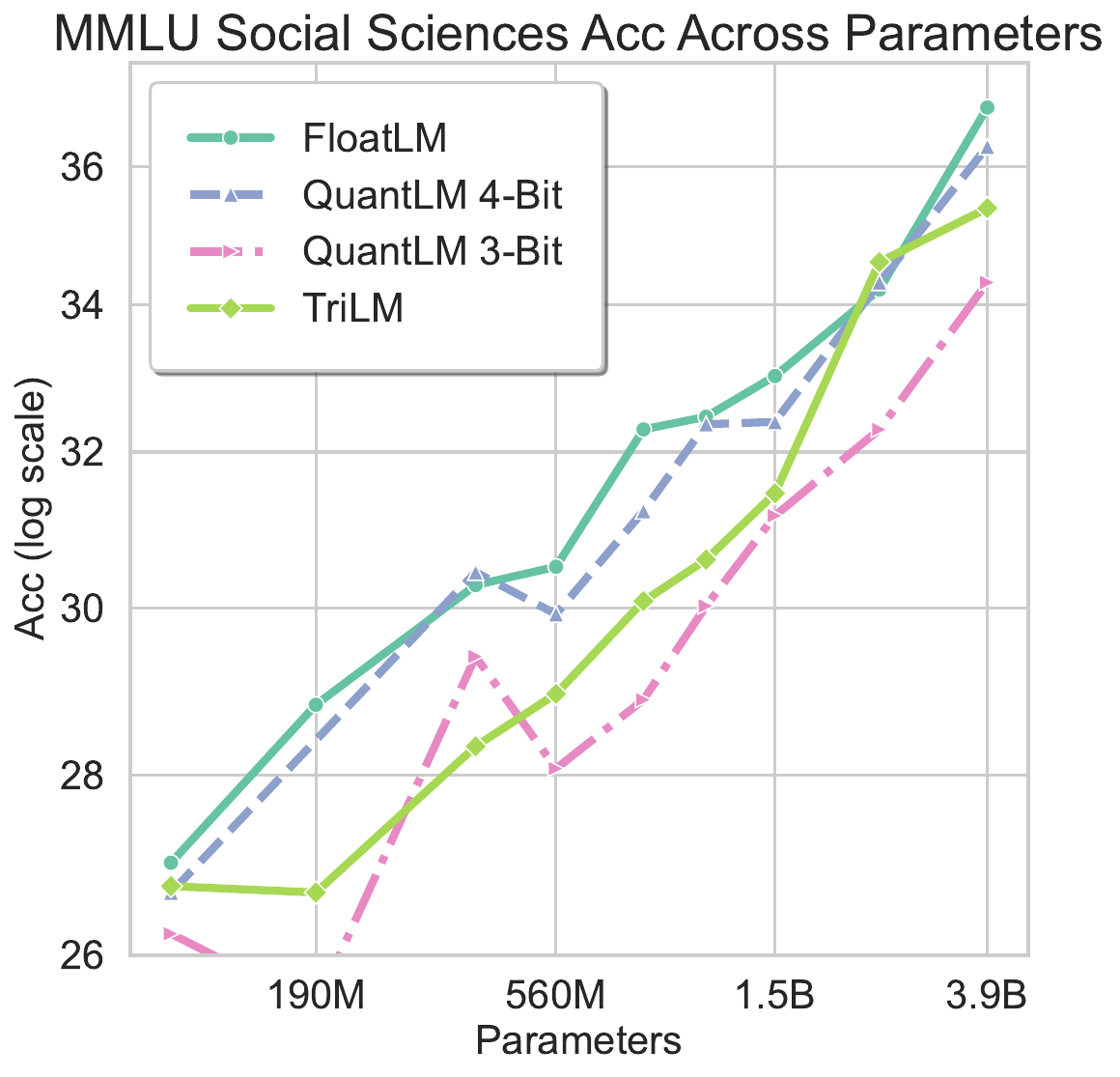}
            \label{fig:MMLU_Social_Sciences_across_params}
        \end{subfigure}
    
        \begin{subfigure}[b]{0.41\textwidth}
            \centering
            \includegraphics[width=\linewidth, height=0.85\textwidth]{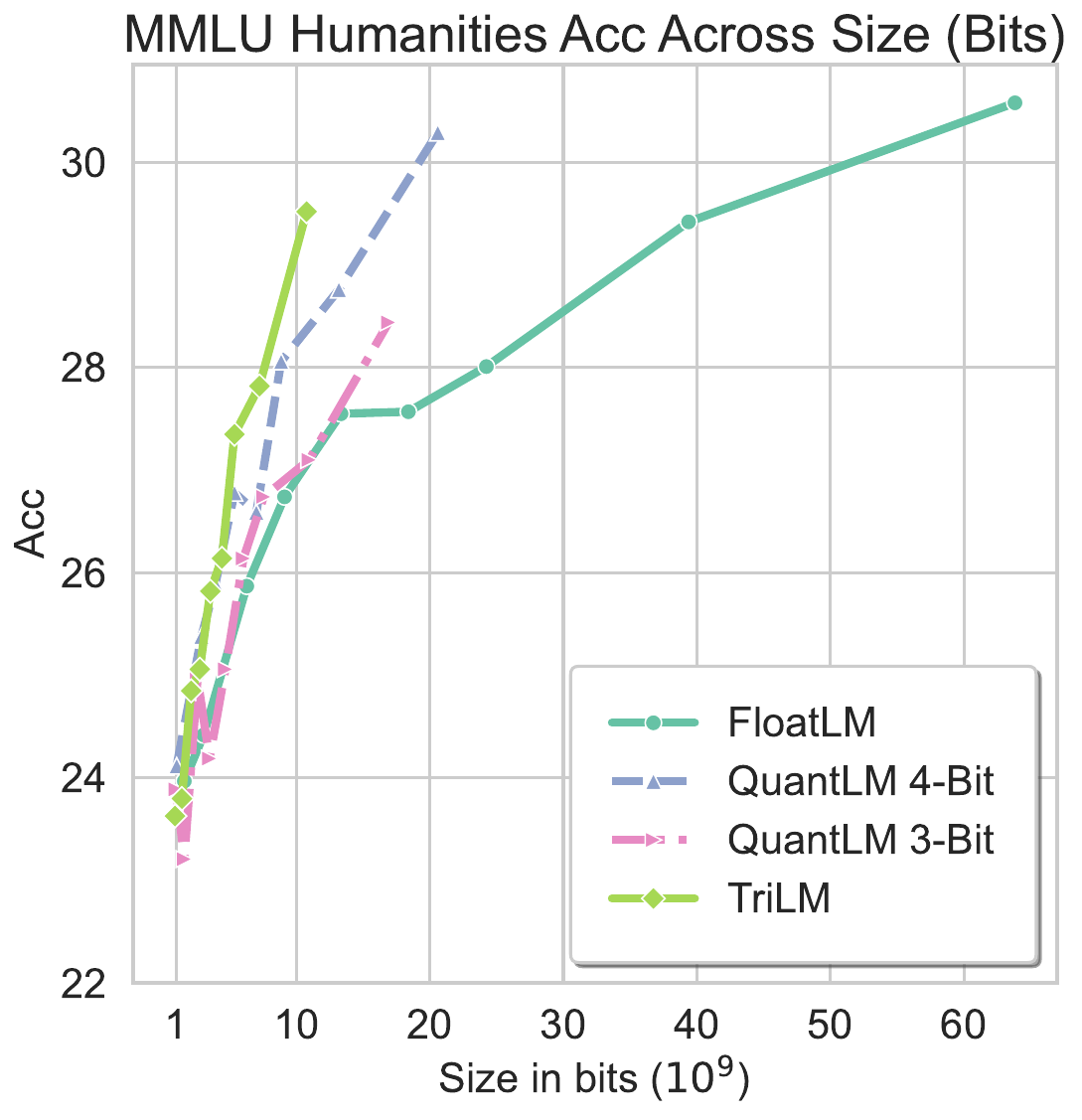}
            \label{fig:MMLU_Humanities_across_size}
        \end{subfigure}
        \begin{subfigure}[b]{0.4\textwidth}
            \centering
            \includegraphics[width=\linewidth, height=0.85\textwidth]{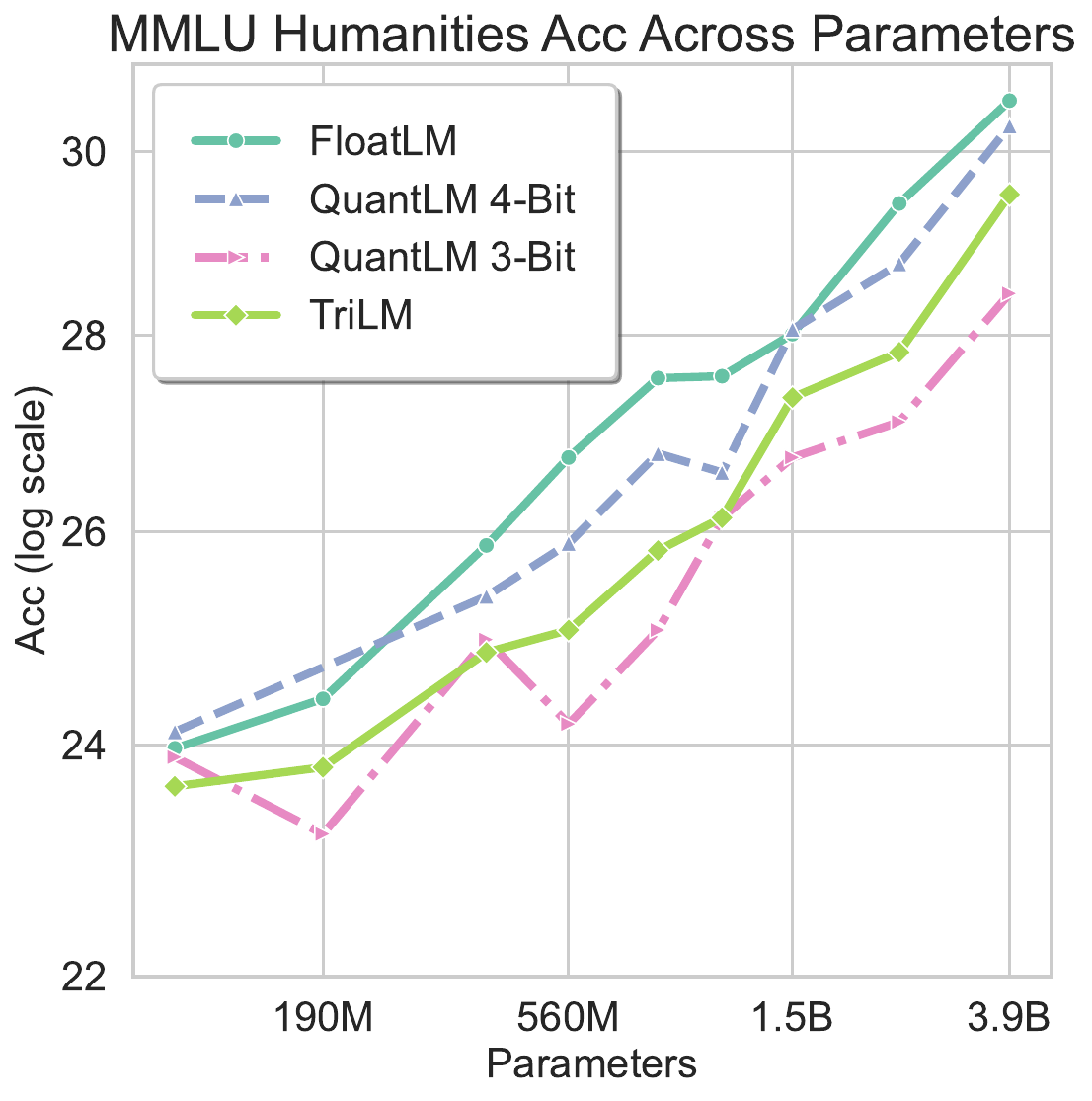}
            \label{fig:MMLU_Humanities_across_params}
        \end{subfigure}

        \begin{subfigure}[b]{0.41\textwidth}
            \centering
            \includegraphics[width=\linewidth, height=0.85\textwidth]{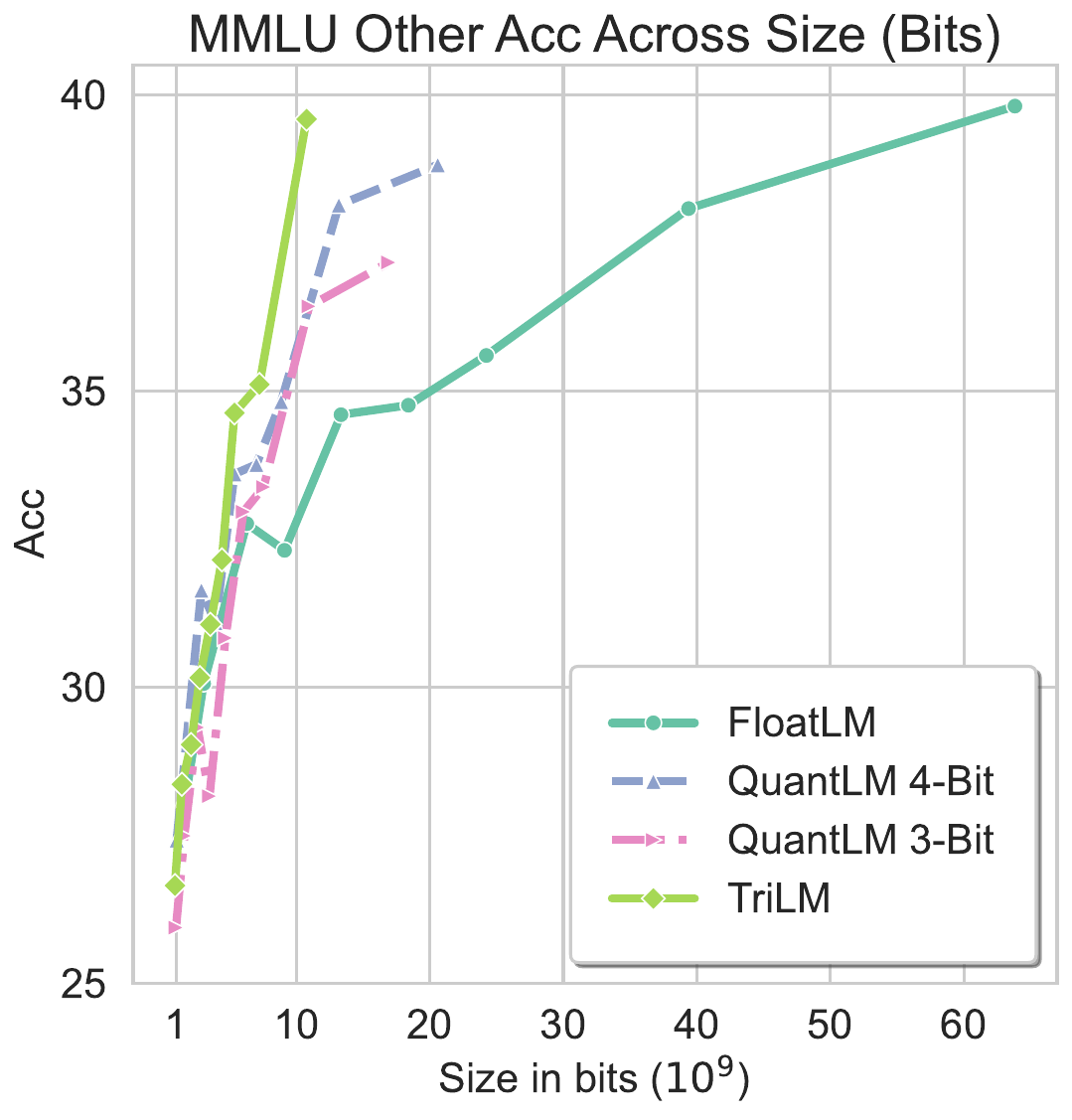}
            \label{fig:MMLU_Other_across_size}
        \end{subfigure}
        \begin{subfigure}[b]{0.4\textwidth}
            \centering
            \includegraphics[width=\linewidth, height=0.85\textwidth]{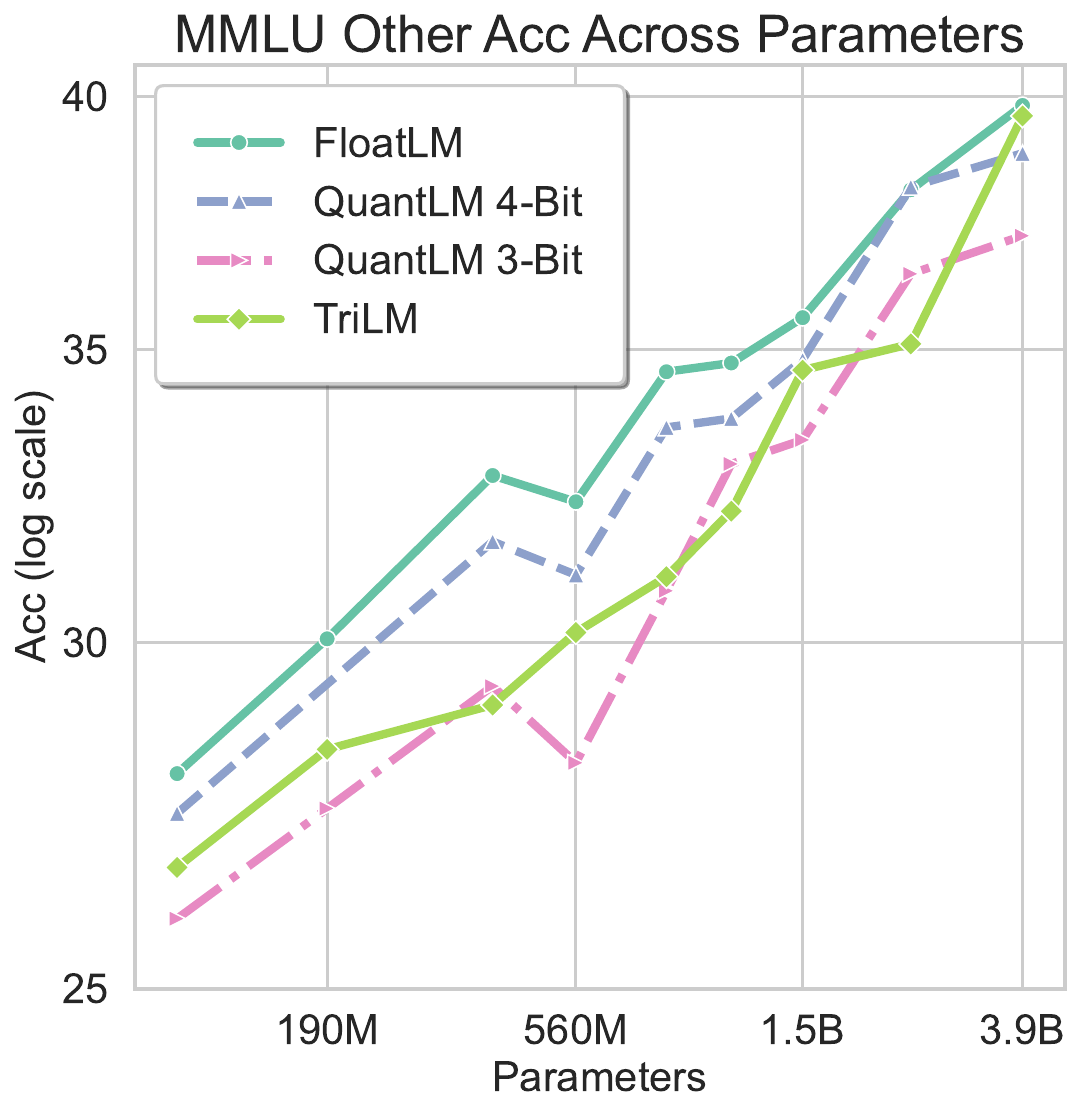}
            \label{fig:MMLU_Other_across_params}
        \end{subfigure}
        
        \caption{\footnotesize{Model performance on MMLU subsets: STEM, Humanities, Social Sciences, and others. Plot accuracy scores against model size in bits (left) and number of parameters (right), ranging from 560M to 3.9B parameters for TriLM (ternary), FloatLM (FP16), and QuantLM (3-bit \& 4-bit).}}
    \end{figure}
        
\clearpage

\section{Ablations} 
\label{appendix::section:Ablations}

        Table \ref{tab:commonsense1_scores_evaluation_ablation_part1} and \ref{tab:commonsense1_scores_evaluation_ablation_part2} shows the performance of ablation 100B token training runs over the six commonsense benchmarks from BitNet b1.58 at 1.1B parameters. The first two rows show the performance of TriLM 1.1B and Float 1.1B at this token count, followed by our replication of BitNet b1.58 (Ours) as well as the scores from BitNet b1.58 over 700M and 1.3B parameters. We observe that at this scale, TriLM does not come close to matching the performance of FloatLM, but it outperforms much larger BitNets. The next two rows show the performance of TriLM 1.1B and FloatLM 1.1B when trained on 100B tokens of FineWeb, instead of SlimPajama. While the performance of both the models improves on FineWeb, the average difference in their performance across datasets remains the same. Lastly, we show the performances across various optimization schedules. A significant drop in averaged performance is noticed when the baseline schedule of linear decay with constant weight decay is used. The gains from dropping l2 regularization in the schedule are more than that of dropping the peak learning rate, however, not enough to match that of TriLM 1.1B's schedule.

\vspace{0.5 cm}
\begin{table} [h]
    \tiny
    \centering
        \begin{tabular}{lcccccc}
            \toprule
              Models  & \multicolumn{2}{c}{HellaSwag} & \multicolumn{2}{c}{PIQA} & WinoGrande \\
            & Acc N. & Acc & Acc N. & Acc & Acc &  \\ 
               \midrule
                FloatLM 1.1B & 50.0 $\pm$ 0.5 & 39.3 $\pm$ 0.5 & 70.9 $\pm$ 1.0 & 70.1 $\pm$ 1.0 & 55.4 $\pm$ 1.4 \\ 
                
                TriLM 1.1B & 46.8 $\pm$ 0.5 & 37.1 $\pm$ 0.4 & 69.4 $\pm$ 1.0 & 69.4 $\pm$ 1.0 & 53.8 $\pm$ 1.4 \\ 
                
                \midrule
                BitNet b1.58 1.1B (Ours) & 47.0 $\pm$ 0.5 & 37.1 $\pm$ 0.4 & 69.4 $\pm$ 1.0 & 69.6 $\pm$ 1.0 & 53.4 $\pm$ 1.4\\ 
                
                BitNet b1.58 700M & \multicolumn{2}{c}{35.1} & \multicolumn{2}{c}{68.1} & 55.2 \\
                
                BitNet b1.58 1.3B & \multicolumn{2}{c}{37.7} & \multicolumn{2}{c}{68.8} & 55.8 \\
                \midrule
                TriLM 1.1B FineWeb & 50.0 $\pm$ 0.5 & 39.2 $\pm$ 0.4 & 70.2 $\pm$ 1.0 & 70.1 $\pm$ 1.0 & 56.6 $\pm$ 1.3  \\ 
                FloatLM 1.1B FineWeb & 52.7 $\pm$ 0.5 & 41.1 $\pm$ 0.4 & 73.0 $\pm$ 1.0 & 71.3 $\pm$ 1.0 & 56.7 $\pm$ 1.3 \\
                \midrule
                TriLM 1.1B Only Peak LR Dropped & 46.7 $\pm$ 0.5 & 36.8 $\pm$ 0.4 & 68.9 $\pm$ 1.0 & 69.5 $\pm$ 1.0 & 55.4 $\pm$ 1.4  \\ 
                TriLM 1.1B Only L2 Reg. Dropped & 47.1 $\pm$ 0.5 & 37.5 $\pm$ 0.4 & 68.6 $\pm$ 1.0 & 69.4 $\pm$ 1.0 & 55.2 $\pm$ 1.4  \\ 
                TriLM 1.1B Baseline Schedule & 46.0 $\pm$ 0.5 & 36.9 $\pm$ 0.4 & 69.3 $\pm$ 1.0 & 69.1 $\pm$ 1.0 & 56.2 $\pm$ 1.3  \\ 
            \bottomrule
        \end{tabular}
        \vspace{0.5 cm}
        
\caption{ \footnotesize{Ablation Common Sense Task Performance: HellaSwag, PIQA, WinoGrande, Arc Easy, Arc Challenge, BoolQ (Contd.). BitNet b1.58's scores from \cite{bitnet_b1.58}. All runs are for 100B tokens on Slim Pajama, except those explicitly stated as FineWeb}}
\label{tab:commonsense1_scores_evaluation_ablation_part2}
\end{table}

\begin{table} [h]
    \tiny
    \centering
        \begin{tabular}{lcccccccc}
            \toprule
              Models  & \multicolumn{2}{c}{Arc Challenge} & \multicolumn{2}{c}{Arc Easy} & BoolQ & Avg ( HellaSwag, PIQA, WinoGrande,\\
            & Acc N. & Acc & Acc N. & Acc & Acc & Arc Easy, Arc Challenge, and BoolQ) \\ 
               \midrule
                FloatLM 1.1B & 26.3 $\pm$ 1.3 & 22.5 $\pm$ 1.2 & 50.3 $\pm$ 1.0 & 56.8 $\pm$ 1.0 & 60.6 $\pm$ 0.8 & 52.2 \\ 
                TriLM 1.1B & 26.7 $\pm$ 1.3 & 22.9 $\pm$ 1.2 & 49.7 $\pm$ 1.0 & 55.0 $\pm$ 1.0 & 54.9 $\pm$ 0.8 & 50.2 \\ 
                \midrule
                BitNet b1.58 1.1B (Ours) & 26.1 $\pm$ 1.2 & 23.6 $\pm$ 1.2 & 47.7 $\pm$ 1.0 & 55.3 $\pm$ 1.0 & 49.7 $\pm$ 0.8 & 48.9 \\ 
                BitNet b1.58 700M & \multicolumn{2}{c}{21.4} & \multicolumn{2}{c}{51.8}  & 58.2 & 48.3 \\
                BitNet b1.58 1.3B & \multicolumn{2}{c}{24.2} & \multicolumn{2}{c}{54.9} & 56.7 & 49.6 \\
                \midrule
                TriLM 1.1B FineWeb & 31.7 $\pm$ 1.3 & 31.9 $\pm$ 1.3 & 63.1 $\pm$ 0.9 & 66.8 $\pm$ 0.9 & 58.3 $\pm$ 0.8 & 54.9 \\ 
                FloatLM 1.1B FineWeb & 34.4 $\pm$ 1.3 & 33.0 $\pm$ 1.3 & 65.7 $\pm$ 0.9 & 70.2 $\pm$ 0.9 & 59.3 $\pm$ 0.8 & 56.9 \\
                \midrule
                TriLM 1.1B Only Peak LR Dropped & 27.4 $\pm$ 1.3 & 23.6 $\pm$ 1.2 & 48.3 $\pm$ 1.0 & 55.1 $\pm$ 1.0 & 51.6 $\pm$ 0.8 & 49.7 \\ 
                TriLM 1.1B Only L2 Reg. Dropped & 27.6 $\pm$ 1.3 & 24.8 $\pm$ 1.2 & 49.2 $\pm$ 1.0 & 55.1 $\pm$ 1.0 & 53.1 $\pm$ 0.8 &  50.1 \\ 
                TriLM 1.1B Baseline Schedule & 26.2 $\pm$ 1.2 & 23.2 $\pm$ 1.2 & 48.0 $\pm$ 1.0 & 54.0 $\pm$ 1.0 & 49.4 $\pm$ 0.8  & 49.1 \\ 
                \bottomrule  
        \end{tabular}
        \vspace{0.5 cm}
        
\caption{\footnotesize{Ablation Common Sense Task Performance: HellaSwag, PIQA, WinoGrande, Arc Easy, Arc Challenge, BoolQ. BitNet b1.58's scores from \cite{bitnet_b1.58}. All runs are for 100B tokens on Slim Pajama, except those explicitly stated as FineWeb}}
\label{tab:commonsense1_scores_evaluation_ablation_part1}
\end{table}

        \begin{table*}
            \scriptsize
            \centering
            \vspace{-0.2 cm}
            \begin{adjustbox}{width=0.8\textwidth}
            \begin{tabular}{lccccc}
                \toprule
                Models & TriviaQA & \multicolumn{2}{c}{CrowsPairs}     & Big Bench BBQ Lite & TruthfulQA \\
                      & Exact Match & Likelihood diff. & Pct stereotype & Acc       & Acc \\
                \midrule
                FloatLM 99M  &   0.6$\pm$ 0.1  &   372.4$\pm$ 14.6   &   55.4$\pm$ 1.2   &   30.8$\pm$ 0.4    &   24.4$\pm$ 1.5 \\
                QuantLM 99M 8-Bit   &   0.6$\pm$ 0.1  &   370.9$\pm$ 14.6   &   55.1$\pm$ 1.2   &   26.5$\pm$ 0.3    &   24.1$\pm$ 1.5 \\
                QuantLM 99M 6-Bit   &   0.6$\pm$ 0.1  &   389.8$\pm$ 14.8   &   56.9$\pm$ 1.2   &   26.7$\pm$ 0.3    &   24.2$\pm$ 1.5 \\
                QuantLM 99M 4-Bit   &   0.3$\pm$ 0  &   425.5$\pm$ 15.2   &   54.0$\pm$ 1.2   &   26.2$\pm$ 0.3    &   22.9$\pm$ 1.5 \\
                QuantLM 99M 3-Bit   &   0.1$\pm$ 0  &   611.1$\pm$ 18.8   &   51.0$\pm$ 1.2   &   31.4$\pm$ 0.4    &   24.6$\pm$ 1.5 \\
                TriLM 99M   &   0.1$\pm$ 0  &   362.4$\pm$ 10.8   &   54.2$\pm$ 1.2   &   30.8$\pm$ 0.4    &   24.2$\pm$ 1.5 \\
                Binary 99M  &  0.2$\pm$ 0.0 & 353.5$\pm$ 11.1 &  53.3$\pm$ 1.2 & 31.5$\pm$ 0.3 & 25.7$\pm$ 1.5 \\
                \midrule
                FloatLM 190M   &   0.6$\pm$ 0.1  &   348.2$\pm$ 11.3   &   55.9$\pm$ 1.2   &   27.3$\pm$ 0.4    &   22.4$\pm$ 1.5 \\
                QuantLM 190M 8-Bit   &   0.7$\pm$ 0.1  &   352.7$\pm$ 11.4   &   56.2$\pm$ 1.2   &   27.1$\pm$ 0.4    &   22.5$\pm$ 1.5 \\
                QuantLM 190M 6-Bit   &   0.7$\pm$ 0.1  &   368.9$\pm$ 11.7   &   56.2$\pm$ 1.2   &   27.2$\pm$ 0.4    &   22.5$\pm$ 1.5 \\
                QuantLM 190M 4-Bit   &   0.0$\pm$ 0  &   961.9$\pm$ 25.4   &   43.8$\pm$ 1.2   &   35.0$\pm$ 0.4    &   24.2$\pm$ 1.5 \\
                QuantLM 190M 3-Bit   &   0.1$\pm$ 0  &   482.7$\pm$ 15.2   &   53.7$\pm$ 1.2   &   26.4$\pm$ 0.3    &   25.0$\pm$ 1.5 \\
                TriLM 190M   &   0.2$\pm$ 0  &   343.5$\pm$ 10.9   &   55.5$\pm$ 1.2   &   29.7$\pm$ 0.4    &   23.9$\pm$ 1.5 \\
                \midrule
                FloatLM 390M   &   2.8$\pm$ 0  &   355.5$\pm$ 10.4   &   59.6$\pm$ 1.2   &   25.4$\pm$ 0.3    &   22.4$\pm$ 1.5 \\
                QuantLM 390M 8-Bit   &   2.9$\pm$ 0  &   355.8$\pm$ 10.4   &   59.8$\pm$ 1.2   &   25.4$\pm$ 0.3    &   22.2$\pm$ 1.5 \\
                QuantLM 390M 6-Bit   &   2.4$\pm$ 0  &   360.5$\pm$ 10.4   &   60.6$\pm$ 1.2   &   25.3$\pm$ 0.3    &   22.8$\pm$ 1.5 \\
                QuantLM 390M 4-Bit   &   1.3$\pm$ 0.1  &   368.2$\pm$ 10.2   &   59.4$\pm$ 1.2   &   25.5$\pm$ 0.3    &   22.8$\pm$ 1.5 \\
                QuantLM 390M 3-Bit   &   0.8$\pm$ 0.1  &   444.4$\pm$ 12.2   &   54.3$\pm$ 1.2   &   26.3$\pm$ 0.3    &   23.0$\pm$ 1.5 \\
                TriLM 390M   &   1.3$\pm$ 0.1  &   344.5$\pm$ 10.3   &   58.3$\pm$ 1.2   &   26.9$\pm$ 0.3    &   24.4$\pm$ 1.5 \\
                % Pythia 410M   &   1.7$\pm$ 0.1  &   337.1$\pm$ 8.8   &   58.2$\pm$ 1.2   &   26.3$\pm$ 0.3    &   23.6$\pm$ 1.5 \\
                \midrule
                FloatLM 560M   &   4.6$\pm$ 0.2  &   351.8$\pm$ 9.9   &   58.9$\pm$ 1.2   &   25.7$\pm$ 0.3    &   21.7$\pm$ 1.4 \\
                QuantLM 560M 8-Bit   &   4.7$\pm$ 0.2  &   352.9$\pm$ 10.0   &   59.2$\pm$ 1.2   &   25.7$\pm$ 0.3    &   21.8$\pm$ 1.4 \\
                QuantLM 560M 6-Bit   &   3.5$\pm$ 0.1  &   353.7$\pm$ 9.9   &   59.3$\pm$ 1.2   &   25.8$\pm$ 0.3    &   22.0$\pm$ 1.5 \\
                QuantLM 560M 4-Bit   &   2.1$\pm$ 0.1  &   372.7$\pm$ 10.7   &   59.2$\pm$ 1.2   &   27.0$\pm$ 0.4    &   22.2$\pm$ 1.5 \\
                QuantLM 560M 3-Bit   &   1.5$\pm$ 0.1  &   411.2$\pm$ 11.3   &   57.9$\pm$ 1.2   &   29.0$\pm$ 0.4    &   22.9$\pm$ 1.5 \\
                TriLM 560M   &   2.4$\pm$ 0.1  &   345.1$\pm$ 10.1   &   58.7$\pm$ 1.2   &   25.5$\pm$ 0.3    &   23.6$\pm$ 1.5 \\
                Binary 560M  &  0.2$\pm$ 0.0 & 356.3$\pm$ 10.4 &  58.5$\pm$ 1.2 & 26.3$\pm$ 0.3 & 23.1$\pm$ 1.4 \\
                \midrule
                FloatLM 830M   &   8.5$\pm$ 0.2  &   354.6$\pm$ 9.6   &   62.6$\pm$ 1.2   &   25.7$\pm$ 0.3    &   23.1$\pm$ 1.5 \\
                QuantLM 830M 8-Bit   &   8.5$\pm$ 0.2  &   354.5$\pm$ 9.6   &   62.1$\pm$ 1.2   &   25.6$\pm$ 0.3    &   23.0$\pm$ 1.5 \\
                QuantLM 830M 6-Bit   &   8.5$\pm$ 0.2  &   354.6$\pm$ 9.6   &   62.7$\pm$ 1.2   &   25.5$\pm$ 0.3    &   22.5$\pm$ 1.5 \\
                QuantLM 830M 4-Bit   &   10.6$\pm$ 0.2  &   364.2$\pm$ 9.8   &   59.9$\pm$ 1.2   &   25.9$\pm$ 0.3    &   21.8$\pm$ 1.4 \\
                QuantLM 830M 3-Bit   &   3.1$\pm$ 0.1  &   389.5$\pm$ 10.9   &   59.9$\pm$ 1.2   &   30.5$\pm$ 0.4    &   24.4$\pm$ 1.5 \\
                TriLM 830M   &   4.3$\pm$ 0.2  &   344.9$\pm$ 10.0   &   60.7$\pm$ 1.2   &   25.1$\pm$ 0.3    &   22.8$\pm$ 1.5 \\
                \midrule
                % Pythia 1B  &   4.2$\pm$ 0.1  &   340.3$\pm$ 8.8   &   60.0$\pm$ 1.2   &   26.7$\pm$ 0.3    &   23.6$\pm$ 1.5 \\
                FloatLM 1.1B   &   12.9$\pm$ 0.3  &   349.2$\pm$ 9.7   &   61.2$\pm$ 1.2   &   25.4$\pm$ 0.3    &   21.4$\pm$ 1.4 \\
                QuantLM 1.1B 8-Bit   &   12.7$\pm$ 0.2  &   349.5$\pm$ 9.7   &   61.1$\pm$ 1.2   &   25.4$\pm$ 0.3    &   21.7$\pm$ 1.4 \\
                QuantLM 1.1B 6-Bit   &   12.4$\pm$ 0.2  &   349.7$\pm$ 9.6   &   59.9$\pm$ 1.2   &   25.5$\pm$ 0.3    &   21.9$\pm$ 1.4 \\
                QuantLM 1.1B 4-Bit   &   9.3$\pm$ 0.2  &   359.1$\pm$ 10.1   &   60.9$\pm$ 1.2   &   25.4$\pm$ 0.3    &   21.3$\pm$ 1.4 \\
                QuantLM 1.1B 3-Bit   &   6.8$\pm$ 0.2  &   422.4$\pm$ 11.5   &   58.7$\pm$ 1.2   &   29.9$\pm$ 0.4    &   24.2$\pm$ 1.5 \\
                TriLM 1.1B   &   1.9$\pm$ 0.1  &   343.4$\pm$ 9.9   &   61.4$\pm$ 1.2   &   25.8$\pm$ 0.3    &   21.5$\pm$ 1.4 \\
                Binary 1.1B  &  2.2$\pm$ 0.1 & 351.6$\pm$ 9.8 &  58.3$\pm$ 1.2 & 26.4$\pm$ 0.3 & 23.2$\pm$ 1.4 \\
                \midrule
                % Pythia 1.4B  &   5.6$\pm$ 0.2  &   337.3$\pm$ 8.6   &   60.8$\pm$ 1.2   &   25.7$\pm$ 0.3    &   22.8$\pm$ 1.5 \\
                FloatLM 1.5B   &   12.2$\pm$ 0.2  &   351.9$\pm$ 9.6   &   61.6$\pm$ 1.2   &   26.8$\pm$ 0.3    &   21.8$\pm$ 1.4 \\
                QuantLM 1.5B 8-Bit   &   12.5$\pm$ 0.2  &   352.4$\pm$ 9.6   &   61.6$\pm$ 1.2   &   26.8$\pm$ 0.3    &   21.8$\pm$ 1.4 \\
                QuantLM 1.5B 6-Bit   &   11.3$\pm$ 0.2  &   350.9$\pm$ 9.7   &   61.9$\pm$ 1.2   &   27.1$\pm$ 0.4    &   21.5$\pm$ 1.4 \\
                QuantLM 1.5B 4-Bit   &   9.0$\pm$ 0.2  &   357.9$\pm$ 9.8   &   60.7$\pm$ 1.2   &   25.9$\pm$ 0.3    &   20.8$\pm$ 1.4 \\
                QuantLM 1.5B 3-Bit   &   4.2$\pm$ 0.1  &   400.0$\pm$ 10.6   &   60.9$\pm$ 1.2   &   26.8$\pm$ 0.3    &   20.8$\pm$ 1.4 \\
                TriLM 1.5B   &   5.9$\pm$ 0.1  &   348.9$\pm$ 9.9   &   59.9$\pm$ 1.2   &   25.2$\pm$ 0.3    &   21.7$\pm$ 1.4 \\
                \midrule
                FloatLM 2.4B  &   20.7$\pm$ 0.3  &   360.4$\pm$ 9.4   &   64.2$\pm$ 1.2   &   26.7$\pm$ 0.3    &   21.7$\pm$ 1.4 \\
                QuantLM 2.4B 8-Bit   &   20.7$\pm$ 0.3  &   360.5$\pm$ 9.4   &   64.2$\pm$ 1.2   &   26.5$\pm$ 0.3    &   21.9$\pm$ 1.4 \\
                QuantLM 2.4B 6-Bit   &   20.4$\pm$ 0.3  &   360.8$\pm$ 9.5   &   63.4$\pm$ 1.2   &   26.4$\pm$ 0.3    &   21.8$\pm$ 1.4 \\
                QuantLM 2.4B 4-Bit   &   21.1$\pm$ 0.3  &   358.7$\pm$ 9.6   &   63.4$\pm$ 1.2   &   26.0$\pm$ 0.3    &   21.7$\pm$ 1.4 \\
                QuantLM 2.4B 3-Bit   &   10.9$\pm$ 0.2  &   360.2$\pm$ 9.5   &   59.9$\pm$ 1.2   &   25.8$\pm$ 0.3    &   21.5$\pm$ 1.4 \\
                TriLM 2.4B   &   12.3$\pm$ 0.1  &   353.0$\pm$ 10.0   &   64.1$\pm$ 1.2   &   25.4$\pm$ 0.3    &   23.0$\pm$ 1.5 \\
                % Pythia 2.8B  &   9.9$\pm$ 0.2  &   345.4$\pm$ 8.6   &   63.2$\pm$ 1.2   &   25.9$\pm$ 0.3    &   21.1$\pm$ 1.4 \\
                \midrule
                FloatLM 3.9B  &   21.5$\pm$ 0.3  &   359.2$\pm$ 9.6   &   64.7$\pm$ 1.2   &   25.4$\pm$ 0.3    &   23.6$\pm$ 1.5 \\
                QuantLM 3.9B 8-Bit   &   21.7$\pm$ 0.3  &   359.8$\pm$ 9.6   &   64.6$\pm$ 1.2   &   25.4$\pm$ 0.3    &   23.6$\pm$ 1.5 \\ 
                QuantLM 3.9B 6-Bit   &   21.0$\pm$ 0.3  &   359.5$\pm$ 9.6   &   63.9$\pm$ 1.2   &   25.4$\pm$ 0.3    &   23.5$\pm$ 1.5 \\
                QuantLM 3.9B 4-Bit   &   17.9$\pm$ 0.3  &   365.5$\pm$ 9.7   &   64.8$\pm$ 1.2   &   25.3$\pm$ 0.3    &   24.2$\pm$ 1.5 \\
                QuantLM 3.9B 3-Bit   &   8.2$\pm$ 0.2  &   365.9$\pm$ 9.8   &   64.3$\pm$ 1.2   &   25.5$\pm$ 0.3    &   21.9$\pm$ 1.4 \\
                TriLM 3.9B   &   21.3$\pm$ 0.3  &   362.4$\pm$ 9.6   &   65.4$\pm$ 1.2   &   25.9$\pm$ 0.3    &   24.1$\pm$ 1.5 \\
                \bottomrule
            \end{tabular}
            \end{adjustbox}
            \caption{ \footnotesize{Spectra Suite Performance (Part 4): TriviaQA, CrowsPairs, Big Bench BBQ Lite, TruthQA. We additionally also include Pythia's performance scores.} }
            \label{tab:evaluation_spectra_suite_part3}
        \end{table*}

        \begin{table*}
            \scriptsize
            \centering
            \vspace{-0.2 cm}
            \begin{adjustbox}{width=0.7\textwidth}
                \begin{tabular}{lccccc}
                    \toprule
                    Models            & \multicolumn{5}{c}{MMLU Accuracy} \\
                                      & Stem          & Humanities    & Social Sciences & Other       & Avg. \\
                    \toprule
                    FloatLM 99M       & 22.8$\pm$ 0.7 & 24.0$\pm$ 0.6 & 27.0$\pm$ 0.8 & 28.0$\pm$ 0.8 & 25.3$\pm$ 0.4 \\
                    QuantLM 99M 8-Bit & 22.9$\pm$ 0.7 & 24.2$\pm$ 0.6 & 26.9$\pm$ 0.8 & 27.9$\pm$ 0.8 & 25.3$\pm$ 0.4 \\
                    QuantLM 99M 6-Bit & 22.7$\pm$ 0.7 & 24.1$\pm$ 0.6 & 26.6$\pm$ 0.8 & 28.2$\pm$ 0.8 & 25.2$\pm$ 0.4 \\
                    QuantLM 99M 4-Bit & 22.9$\pm$ 0.7 & 24.1$\pm$ 0.6 & 26.7$\pm$ 0.8 & 27.4$\pm$ 0.8 & 25.1$\pm$ 0.4 \\
                    QuantLM 99M 3-Bit & 23.5$\pm$ 0.8 & 23.9$\pm$ 0.6 & 26.2$\pm$ 0.8 & 25.9$\pm$ 0.8 & 24.8$\pm$ 0.4 \\
                    TriLM 99M         & 23.9$\pm$ 0.8 & 23.6$\pm$ 0.6 & 26.7$\pm$ 0.8 & 26.6$\pm$ 0.8 & 25.0$\pm$ 0.4 \\
                    Binary 99M  &  21.6$\pm$ 0.7 & 24.3$\pm$ 0.6 &  21.8$\pm$ 0.7 & 24.0$\pm$ 0.7 & 23.1$\pm$ 0.3 \\
                    \midrule
                    FloatLM 190M      & 24.0$\pm$ 0.8 & 24.4$\pm$ 0.6 & 28.8$\pm$ 0.8 & 30.1$\pm$ 0.8 & 26.5$\pm$ 0.4 \\
                    QuantLM 190M 8-Bit & 24.1$\pm$ 0.8 & 24.5$\pm$ 0.6 & 28.9$\pm$ 0.8 & 30.0$\pm$ 0.8 & 26.6$\pm$ 0.4 \\
                    QuantLM 190M 6-Bit & 24.1$\pm$ 0.8 & 24.5$\pm$ 0.6 & 28.3$\pm$ 0.8 & 29.8$\pm$ 0.8 & 26.4$\pm$ 0.4 \\
                    QuantLM 190M 4-Bit & 22.9$\pm$ 0.7 & 22.9$\pm$ 0.6 & 24.5$\pm$ 0.8 & 23.4$\pm$ 0.8 & 23.4$\pm$ 0.4 \\
                    QuantLM 190M 3-Bit & 23.9$\pm$ 0.8 & 23.2$\pm$ 0.6 & 25.4$\pm$ 0.8 & 27.5$\pm$ 0.8 & 24.8$\pm$ 0.4 \\
                    TriLM 190M         & 22.5$\pm$ 0.7 & 23.8$\pm$ 0.6 & 26.7$\pm$ 0.8 & 28.4$\pm$ 0.8 & 25.2$\pm$ 0.4 \\
                    \midrule
                    FloatLM 390M       & 25.8$\pm$ 0.8 & 25.9$\pm$ 0.6 & 30.3$\pm$ 0.8 & 32.8$\pm$ 0.8 & 28.3$\pm$ 0.4 \\
                    QuantLM 390M 8-Bit & 25.7$\pm$ 0.8 & 25.9$\pm$ 0.6 & 30.2$\pm$ 0.8 & 32.4$\pm$ 0.8 & 28.2$\pm$ 0.4 \\
                    QuantLM 390M 6-Bit & 26.0$\pm$ 0.8 & 25.8$\pm$ 0.6 & 30.2$\pm$ 0.8 & 32.3$\pm$ 0.8 & 28.3$\pm$ 0.4 \\
                    QuantLM 390M 4-Bit & 25.5$\pm$ 0.8 & 25.4$\pm$ 0.6 & 30.5$\pm$ 0.8 & 31.6$\pm$ 0.8 & 27.9$\pm$ 0.4 \\
                    QuantLM 390M 3-Bit & 24.4$\pm$ 0.8 & 25.0$\pm$ 0.6 & 29.4$\pm$ 0.8 & 29.3$\pm$ 0.8 & 26.8$\pm$ 0.4 \\
                    TriLM 390M         & 24.1$\pm$ 0.8 & 24.8$\pm$ 0.6 & 28.3$\pm$ 0.8 & 29.0$\pm$ 0.8 & 26.4$\pm$ 0.4 \\
                    \midrule
                    FloatLM 560M       & 24.8$\pm$ 0.8 & 26.7$\pm$ 0.6 & 30.5$\pm$ 0.8 & 32.3$\pm$ 0.8 & 28.4$\pm$ 0.4 \\
                    QuantLM 560M 8-Bit & 24.8$\pm$ 0.8 & 26.6$\pm$ 0.6 & 30.5$\pm$ 0.8 & 32.1$\pm$ 0.8 & 28.3$\pm$ 0.4 \\
                    QuantLM 560M 6-Bit & 24.6$\pm$ 0.8 & 26.7$\pm$ 0.6 & 30.5$\pm$ 0.8 & 31.3$\pm$ 0.8 & 28.1$\pm$ 0.4 \\
                    QuantLM 560M 4-Bit & 24.7$\pm$ 0.8 & 25.9$\pm$ 0.6 & 29.9$\pm$ 0.8 & 31.1$\pm$ 0.8 & 27.7$\pm$ 0.4 \\
                    QuantLM 560M 3-Bit & 24.5$\pm$ 0.8 & 24.2$\pm$ 0.6 & 28.1$\pm$ 0.8 & 28.2$\pm$ 0.8 & 26.0$\pm$ 0.4 \\
                    TriLM 560M        & 25.0$\pm$ 0.8 & 25.1$\pm$ 0.6 & 29.0$\pm$ 0.8 & 30.2$\pm$ 0.8 & 27.0$\pm$ 0.4 \\
                    Binary 560M  &  21.4$\pm$ 0.7 & 24.2$\pm$ 0.6 &  21.6$\pm$ 0.7 & 23.9$\pm$ 0.7 & 22.9$\pm$ 0.3 \\
                    \midrule
                    FloatLM 830M       & 25.8$\pm$ 0.8 & 27.5$\pm$ 0.6 & 32.3$\pm$ 0.8 & 34.6$\pm$ 0.8 & 29.7$\pm$ 0.4 \\
                    QuantLM 830M 8-Bit & 25.8$\pm$ 0.8 & 27.4$\pm$ 0.6 & 32.1$\pm$ 0.8 & 34.7$\pm$ 0.8 & 29.7$\pm$ 0.4 \\
                    QuantLM 830M 6-Bit & 25.6$\pm$ 0.8 & 27.3$\pm$ 0.6 & 32.1$\pm$ 0.8 & 34.2$\pm$ 0.8 & 29.5$\pm$ 0.4 \\
                    QuantLM 830M 4-Bit & 25.9$\pm$ 0.8 & 26.8$\pm$ 0.6 & 31.2$\pm$ 0.8 & 33.6$\pm$ 0.8 & 29.1$\pm$ 0.4 \\
                    QuantLM 830M 3-Bit & 24.8$\pm$ 0.8 & 25.1$\pm$ 0.6 & 28.9$\pm$ 0.8 & 30.8$\pm$ 0.8 & 27.1$\pm$ 0.4 \\
                    TriLM 830M         & 24.9$\pm$ 0.8 & 25.8$\pm$ 0.6 & 30.1$\pm$ 0.8 & 31.1$\pm$ 0.8 & 27.7$\pm$ 0.4 \\
                    \midrule
                    FloatLM 1.1B       & 26.4$\pm$ 0.8 & 27.6$\pm$ 0.6 & 32.5$\pm$ 0.8 & 34.8$\pm$ 0.8 & 30.0$\pm$ 0.4 \\
                    QuantLM 1.1B 8-Bit & 26.2$\pm$ 0.8 & 27.4$\pm$ 0.6 & 32.5$\pm$ 0.8 & 34.9$\pm$ 0.8 & 29.9$\pm$ 0.4 \\
                    QuantLM 1.1B 6-Bit & 26.0$\pm$ 0.8 & 27.5$\pm$ 0.6 & 32.7$\pm$ 0.8 & 34.9$\pm$ 0.8 & 29.9$\pm$ 0.4 \\
                    QuantLM 1.1B 4-Bit & 26.0$\pm$ 0.8 & 26.6$\pm$ 0.6 & 32.4$\pm$ 0.8 & 33.8$\pm$ 0.8 & 29.3$\pm$ 0.4 \\
                    QuantLM 1.1B 3-Bit & 25.9$\pm$ 0.8 & 26.1$\pm$ 0.6 & 30.0$\pm$ 0.8 & 33.0$\pm$ 0.8 & 28.4$\pm$ 0.4 \\
                    TriLM 1.1B         & 25.2$\pm$ 0.8 & 26.1$\pm$ 0.6 & 30.6$\pm$ 0.8 & 32.2$\pm$ 0.8 & 28.3$\pm$ 0.4 \\
                    Binary 1.1B  &  21.0$\pm$ 0.7 & 24.2$\pm$ 0.6 &  21.7$\pm$ 0.8 & 24.4$\pm$ 0.7 & 23.0$\pm$ 0.3 \\
                    \midrule
                    FloatLM 1.5B       & 26.1$\pm$ 0.8 & 28.0$\pm$ 0.7 & 33.0$\pm$ 0.8 & 35.6$\pm$ 0.8 & 30.4$\pm$ 0.4 \\
                    QuantLM 1.5B 8-Bit & 26.1$\pm$ 0.8 & 28.1$\pm$ 0.7 & 32.9$\pm$ 0.8 & 35.5$\pm$ 0.8 & 30.3$\pm$ 0.4 \\
                    QuantLM 1.5B 6-Bit & 26.3$\pm$ 0.8 & 28.0$\pm$ 0.7 & 33.0$\pm$ 0.8 & 35.4$\pm$ 0.8 & 30.4$\pm$ 0.4 \\
                    QuantLM 1.5B 4-Bit & 26.2$\pm$ 0.8 & 28.1$\pm$ 0.7 & 32.4$\pm$ 0.8 & 34.8$\pm$ 0.8 & 30.1$\pm$ 0.4 \\
                    QuantLM 1.5B 3-Bit & 25.5$\pm$ 0.8 & 26.7$\pm$ 0.6 & 31.2$\pm$ 0.8 & 33.4$\pm$ 0.8 & 28.9$\pm$ 0.4 \\
                    TriLM 1.5B         & 25.7$\pm$ 0.8 & 27.4$\pm$ 0.6 & 31.5$\pm$ 0.8 & 34.6$\pm$ 0.8 & 29.5$\pm$ 0.4 \\
                    \midrule
                    FloatLM 2.4B       & 26.9$\pm$ 0.8 & 29.4$\pm$ 0.7 & 34.2$\pm$ 0.8 & 38.1$\pm$ 0.9 & 31.8$\pm$ 0.4 \\
                    QuantLM 2.4B 8-Bit & 27.0$\pm$ 0.8 & 29.4$\pm$ 0.7 & 34.1$\pm$ 0.8 & 38.0$\pm$ 0.9 & 31.8$\pm$ 0.4 \\
                    QuantLM 2.4B 6-Bit & 26.8$\pm$ 0.8 & 29.5$\pm$ 0.7 & 34.2$\pm$ 0.8 & 38.2$\pm$ 0.9 & 31.8$\pm$ 0.4 \\
                    QuantLM 2.4B 4-Bit & 26.5$\pm$ 0.8 & 28.8$\pm$ 0.7 & 34.3$\pm$ 0.8 & 38.1$\pm$ 0.9 & 31.5$\pm$ 0.4 \\
                    QuantLM 2.4B 3-Bit & 25.5$\pm$ 0.8 & 27.1$\pm$ 0.6 & 32.3$\pm$ 0.8 & 36.4$\pm$ 0.9 & 29.9$\pm$ 0.4 \\
                    TriLM 2.4B         & 27.4$\pm$ 0.8 & 27.8$\pm$ 0.6 & 34.6$\pm$ 0.9 & 35.1$\pm$ 0.8 & 30.8$\pm$ 0.4 \\
                    \midrule
                    FloatLM 3.9B       & 27.7$\pm$ 0.8 & 30.6$\pm$ 0.7 & 36.9$\pm$ 0.9 & 39.8$\pm$ 0.9 & 33.3$\pm$ 0.4 \\
                    QuantLM 3.9B 8-Bit & 27.6$\pm$ 0.8 & 30.7$\pm$ 0.7 & 37.0$\pm$ 0.9 & 39.7$\pm$ 0.9 & 33.4$\pm$ 0.4 \\
                    QuantLM 3.9B 6-Bit & 27.3$\pm$ 0.8 & 30.3$\pm$ 0.7 & 36.9$\pm$ 0.9 & 39.3$\pm$ 0.9 & 33.1$\pm$ 0.4 \\
                    QuantLM 3.9B 4-Bit & 27.1$\pm$ 0.8 & 30.3$\pm$ 0.7 & 36.3$\pm$ 0.9 & 38.8$\pm$ 0.9 & 32.8$\pm$ 0.4 \\
                    QuantLM 3.9B 3-Bit & 27.3$\pm$ 0.8 & 28.4$\pm$ 0.7 & 34.3$\pm$ 0.9 & 37.2$\pm$ 0.9 & 31.4$\pm$ 0.4 \\
                    TriLM 3.9B         & 28.3$\pm$ 0.8 & 29.5$\pm$ 0.7 & 35.4$\pm$ 0.9 & 39.6$\pm$ 0.9 & 32.8$\pm$ 0.4 \\
                \bottomrule
            \end{tabular}
            \end{adjustbox}
            \caption{\footnotesize{Spectra Suite Performance (Part 5): MMLU- STEM, Humanities, Social Sciences, Others. }}
            \label{tab:evaluation_spectra_mmlu}
        \end{table*}

    \clearpage
    \section{Illustrative examples of TriLM 3.9B's completion capabilities}
    \label{Tril_LM_generation}
        
        We showcase instances of outputs produced by TriLM (3.9B) across diverse tasks, highlighting its proficiency in tasks such as comprehension, prompt completion, and creative composition. 
        
        \begin{tcolorbox}[title={Generated Output on Reading Comprehension by TriLM (3.9B)},
            colframe=gray!100!black,
            colback=gray!5!white,
            coltitle=white, 
            colbacktitle=green!45!black,
            ]
            
            \textbf{Title: The Blitz}
            \textbf{Background:} From the German point of view, March 1941 saw an improvement. The Luftwaffe flew 4,000 sorties that month, including 12 major and three heavy attacks. The electronic war intensified but the Luftwaffe flew major inland missions only on moonlit nights. Ports were easier to find and made better targets. To confuse the British, radio silence was observed until the bombs fell. X- and Y-Gerät beams were placed over false targets and switched only at the last minute. Rapid frequency changes were introduced for X-Gerät, whose wider band of frequencies and greater tactical flexibility ensured it remained effective at a time when British selective jamming was degrading the effectiveness of Y-Gerät.
        
            \vspace{5mm}
        
            \textbf{Q: How many sorties were flown in March 1941?} \\
            \textbf{A:} 4,000
        
            \vspace{3mm}
        
            \textbf{Q: When did the Luftwaffe fly inland missions?} \\
            {\color{teal}{\textbf{A:} Only during moonlit nights}} \\
            
            \tcblower 
            \vspace{2mm}
            
            \textbf{Title: Helsinki}
            
            \textbf{Background:} Helsinki is the capital and largest city of Finland. It is in the region of Uusimaa, in southern Finland, on the shore of the Gulf of Finland. Helsinki has a population of , an urban population of , and a metropolitan population of over 1.4 million, making it the most populous municipality and urban area in Finland. Helsinki is some north of Tallinn, Estonia, east of Stockholm, Sweden, and west of Saint Petersburg, Russia. Helsinki has close historical connections with these three cities. The Helsinki metropolitan area includes the urban core of Helsinki, Espoo, Vantaa, Kauniainen, and surrounding commuter towns. It is the world’s northernmost metro area of over one million people, and the city is the northernmost capital of an EU member state. The Helsinki metropolitan area is the third largest metropolitan area in the Nordic countries after Stockholm and Copenhagen, and the City of Helsinki is the third largest after Stockholm and Oslo. Helsinki is Finland’s major political, educational, financial, cultural, and research center as well as one of northern Europe’s major cities. Approximately 75\% of foreign companies that operate in Finland have settled in the Helsinki region. The nearby municipality of Vantaa is the location of Helsinki Airport, with frequent service to various destinations in Europe and Asia.
        
            \vspace{5mm}
        
            \textbf{Q: What is the most populous municipality in Finland?} \\
            \textbf{A:} Helsinki
        
            \vspace{3mm}
        
            \textbf{Q: How many people live there?} \\
            \textbf{A:} 1.4 million in the metropolitan area
            \vspace{3mm}
        
            \textbf{Q: What percent of the foreign companies that operate in Finland are in Helsinki?} \\
            \textbf{A:} 75\%
        
            \vspace{3mm}
        
            \textbf{Q: What towns are a part of the metropolitan area?} \\
            \color{teal}{\textbf{A:} Espoo, Vantaa, Kauniainen, and surrounding commuter towns}
            
        \end{tcolorbox}
        \vspace{2 cm}
        \begin{tcolorbox}[title={Auto-complete on a creative writing task by TriLM (3.9B)},
            colframe=gray!100!black,
            colback=gray!5!white,
            coltitle=white, 
            colbacktitle=green!45!black,
            breakable,
            label=Tab:User:readingcomprehension1]
        \textbf{Essay Writing} \\
        \label{app:readingcomprehension}
        
        In the year 2150, humanity had established colonies on Mars. Captain Ramirez was on a routine mission to explore an uncharted region  \color{teal}{of the planet when he and his crew were attacked by Martian natives who wanted them dead for their technology. The captain's ship crashed into the surface of Mars where it exploded in flames. \\
        The survivors managed to escape from the wreckage but they soon found themselves stranded with no way out or back home. They would have to survive alone until rescue came. But what if there is nothing waiting for them? What if they are trapped forever?}
            
        \tcblower 
        \vspace{2mm}
        \textbf{Poem Writing}
        
        \centering    
        In a garden where the roses bloom, \\
        Underneath the silver moon, \\
        A gentle breeze begins to sway, \\
        And whispers secrets of the day. \\
        The stars above start to appear, \\
        While shadows dance and disappear, \\
        The night unfolds in its own time. \\
        \color{teal}{
        The sun rises over the horizon, \\
        As it does every morning, \\
        But this is no ordinary dawn. \\
        It's a new beginning for all who are here today. \\
        For those who have come before us, \\
        And those yet to be born. \\
        We celebrate life with music and laughter, \\
        With friends old and new. \\
        This is our moment to shine! \\}
        \end{tcolorbox}
        \vspace{ 1cm }
        \begin{tcolorbox}[
          title={Auto-complete on various prompts by TriLM (3.9B)},
          colframe=gray!100!black,
          colback=gray!5!white,
          coltitle=white,
        colbacktitle=green!45!black,
          breakable,
          label=Tab:User:readingcomprehension2
        ]
        Making a cake: Several cake pops are shown on a display. A woman and girl are shown making the cake pops in a kitchen. \textcolor{teal}{They use a mixer to mix ingredients, then knead dough by hand. The women roll out the dough with a rolling pin, cut it into shapes using cookie cutters, bake them, and decorate them with frosting or sprinkles.}
        
        \
        \\
        \vspace{5mm}
        
        The energy given up by electrons as they move through the electron transport chain is used to \textcolor{teal}{produce ATP. In this process, oxygen and water are produced in a reaction called cellular respiration.}
        
        \
        \\
        \vspace{5mm}

        When talking about sensitive topics, it is crucial to be aware of the fact that \textcolor{teal}{there are many different types of people. Some will react negatively and others positively. It's important not to take offense when someone says something you don't like.}    

        \
        \\
        \vspace{5mm}   
        If someone disagrees with you, it is best \textcolor{teal}{to ignore them. The most important thing in life is not what we have but who we are and how we treat others.}
        \end{tcolorbox}

\end{document}